\documentclass[11pt,letterpaper]{mystyle}
\usepackage{fancyhdr}

\usepackage{tikz}
\usepackage[utf8]{inputenc} 
\usepackage[T1]{fontenc}
\usepackage[numbers]{natbib}

\usepackage{adjustbox}
\usepackage{breakurl}    
\usepackage{hyperref}
\usepackage[edges]{forest}
\usepackage{subcaption}
\usepackage{soul}
\usepackage{multirow}
\usepackage[utf8]{inputenc} 
\usepackage{booktabs}       
\usepackage{amsfonts}       
\usepackage{nicefrac}      
\usepackage{microtype}     
\usepackage{xcolor}        
\usepackage{colortbl}
\usepackage{enumitem}
\usepackage[numbers]{natbib}
\usepackage{amssymb}
\usepackage{wrapfig}
\usepackage{courier}
\usepackage{bxcoloremoji}
\usepackage{CJKutf8}
\usepackage{ragged2e}
\usepackage{longtable}
\usepackage{threeparttable}

\usepackage{tabularray}
\UseTblrLibrary{booktabs} % 在 tabularray 中加载 booktabs 样式

\usepackage{fontawesome5}

\usepackage{makecell}
\usepackage{array,booktabs,tabularx,ragged2e}
\newcolumntype{Y}{>{\RaggedRight\arraybackslash}X}

\definecolor{orangenode}{RGB}{255,200,180}
\definecolor{greennode}{RGB}{220,240,220}
\definecolor{bluenode}{RGB}{200,220,255}

\definecolor{lightblue}{HTML}{E0F2F7}
\definecolor{darkblue}{HTML}{003366}
\definecolor{tablehead}{HTML}{2C5E7B}
\definecolor{tablerow}{HTML}{F0F8FF}

\definecolor{ngreen}{HTML}{D5E8D4}
\definecolor{nblue}{HTML}{DAE8FC}
\definecolor{npurple}{HTML}{E1D5E7}

\newcommand{\bolditalic}[1]{\textit{\textbf{#1}}}

\definecolor{hidden-red}{RGB}{205, 44, 36}
\definecolor{hidden-blue}{RGB}{194,232,247}
\definecolor{hidden-orange}{RGB}{243,202,120}
\definecolor{hidden-green}{RGB}{34,139,34}
\definecolor{hidden-pink}{RGB}{255,245,247}
\definecolor{hidden-black}{RGB}{20,68,106}
\definecolor{purple}{RGB}{144,153,196}
\definecolor{yellow}{RGB}{255,228,123}
\definecolor{hidden-yellow}{RGB}{255,248,203}
\definecolor{tkcolor}{RGB}{224,223,255}
\definecolor{darkblue}{rgb}{0, 0.40, 0.75}

\hypersetup{colorlinks=true, citecolor=darkblue, linkcolor=darkblue, urlcolor=darkblue}
\tcbset{
  aibox/.style={
    width=\linewidth,
    top=8pt,
    bottom=4pt,
    colback=blue!6!white,
    colframe=black,
    colbacktitle=black,
    enhanced,
    center,
    attach boxed title to top left={yshift=-0.1in,xshift=0.15in},
    boxed title style={boxrule=0pt,colframe=white,},
  }
}

\tikzstyle{my-box}=[
rectangle,
draw=hidden-black,
rounded corners,
text opacity=1,
minimum height=1.5em,
minimum width=5em,
inner sep=2pt,
align=center,
fill opacity=.5,
]
\tikzstyle{leaf}=[
my-box,
minimum height=1.5em,
fill=hidden-red!20,
text=black,
align=left,
font=\normalsize,
inner xsep=5pt,
inner ysep=4pt,
text width=40em,
]
\tikzstyle{leaf2}=[
my-box,
minimum height=1.5em,
fill=hidden-green!20,
text=black,
align=left,
font=\normalsize,
inner xsep=5pt,
inner ysep=4pt,
]
\tikzstyle{leaf3}=[
my-box,
minimum height=1.5em,
fill=yellow!32,
text=black,
align=left,
font=\normalsize,
inner xsep=5pt,
inner ysep=4pt,
align=left,
text width=38em,
]
\tikzstyle{leaf4}=[
my-box,
minimum height=1.5em,
fill=hidden-blue!57,
text=black,
align=left,
font=\normalsize,
inner xsep=5pt,
inner ysep=4pt,
text width=40em,
]
\tikzstyle{leaf5}=[
my-box,
minimum height=1.5em,
fill=darkblue!15,
text=black,
align=left,
font=\normalsize,
inner xsep=5pt,
inner ysep=4pt,
]
\tikzstyle{leaf6}=[
my-box,
minimum height=1.5em,
fill=purple!30,
text=black,
align=left,
font=\normalsize,
inner xsep=5pt,
inner ysep=4pt,
]

\usepackage{ltablex} 
\newcolumntype{L}{>{\raggedright\arraybackslash}X}

\newtcolorbox{AIbox}[2][]{aibox,title=#2,#1}

\tcbset{
  takeawaybox/.style={
    width=\linewidth,
    top=8pt,
    bottom=4pt,
    colback=hidden-yellow,
    colframe=black,
    colbacktitle=black,
    enhanced,
    center,
    attach boxed title to top left={yshift=-0.1in,xshift=0.15in},
    boxed title style={boxrule=0pt,colframe=white,},
  }
}

\newtcolorbox{TakeawayBox}[2][]{takeawaybox,title=#2,#1}

\definecolor{blockbg}{RGB}{243,246,249}
\title{Unifying Tree Search Algorithm and Reward Design for LLM Reasoning: A Survey}

\vspace{-5em}
\author{
  Jiaqi Wei$^{1*}$, 
  Xiang Zhang$^{2*}$, 
  Yuejin Yang$^{3*}$, 
  \textbf{Wenxuan Huang}$^{3*}$, 
  \textbf{Juntai Cao}$^{2}$, 
  \textbf{Sheng Xu}$^{3}$,  
  \textbf{Xiang Zhuang}$^{1}$,  
  \textbf{Zhangyang Gao}$^{1}$,  
  \textbf{Muhammad Abdul-Mageed}$^{2}$, 
  \textbf{Laks V.S. Lakshmanan}$^{2}$,  
  \textbf{Chenyu You}$^{4}$, 
  \textbf{Wanli Ouyang}$^{5}$, 
  \textbf{Siqi Sun}$^{3}$
\\

\normalfont{
$^1$ Zhejiang University, 
$^2$ University of British Columbia,
$^3$ Fudan University, \vspace{-5pt}\\
$^4$ Stony Brook University, 
$^{5}$ The Chinese University of Hong Kong,  
$^{*}$ Equal Contribution
}}

\begin{document}

\begin{abstract}
  % \vspace{-3mm}
  \textbf{\large Abstract:}
  % \vspace{1mm}

Deliberative tree search is a cornerstone of modern Large Language Model (LLM) research, driving the pivot from brute-force scaling toward algorithmic efficiency. This single paradigm unifies two critical frontiers: \textbf{Test-Time Scaling (TTS)}, which deploys on-demand computation to solve hard problems, and \textbf{Self-Improvement}, which uses search-generated data to durably enhance model parameters. However, this burgeoning field is fragmented and lacks a common formalism, particularly concerning the ambiguous role of the reward signal---is it a transient heuristic or a durable learning target? This paper resolves this ambiguity by introducing a unified framework that deconstructs search algorithms into three core components: the \emph{Search Mechanism}, \emph{Reward Formulation}, and \emph{Transition Function}. We establish a formal distinction between transient \textbf{Search Guidance} for TTS and durable \textbf{Parametric Reward Modeling} for Self-Improvement. Building on this formalism, we introduce a component-centric taxonomy, synthesize the state-of-the-art, and chart a research roadmap toward more systematic progress in creating autonomous, self-improving agents.

  % \vspace{2mm}
  % $^{*}$ \textit{Equal Contribution}
  
  % $^{\coloremojicode{2709}}$ \textit{Corresponding Author}

  % \vspace{1mm}
  % \textbf{Keywords}: Tree Search, Large Language Models, Test-time Scaling
  % \vspace{6mm}

  % \coloremojicode{1F4C5} \textbf{Date}: 

  % \coloremojicode{1F3E0} \textbf{Homepage}: https://agenticscience.github.io/

  % \github{} \textbf{Github Repository}: https://github.com/AgenticScience/AgenticScience.github.io

  % \coloremojicode{1F4E7} \textbf{Correspondence}: Siqi Sun, \href{}{siqisun@fudan.edu.cn}

  % \coloremojicode{1F4E7} \textbf{Equal contribution}: Jiaqi Wei, Yuejin Yang, Xiang Zhang, Yuhan Chen

  % \textbf{Date}: August 6, 2025

 % \coloremojicode{1F4C5} \textbf{Date}: August, 2025

  \vspace{2mm}

  % \coloremojicode{1F3E0} \textbf{Homepage}: 
  % \href{https://agenticscience.github.io/}{https://agenticscience.github.io/}

  % \textbf{Github Repository}: 
  % \href{https://github.com/More2Search/Awesome-Search-LLM}{https://github.com/More2Search/Awesome-Search-LLM}

    \faGithub \: \textbf{Github Repository}: 
  \href{https://github.com/More2Search/Awesome-Search-LLM}{\ https://github.com/More2Search/Awesome-Search-LLM}

  % \coloremojicode{1F4E7} \textbf{Correspondence}: \href{}{siqisun@fudan.edu.cn}

    % \coloremojicode{1F4E7} \textbf{Contact}: \href{}{ \textbraceleft weijiaqi, yangyuejin\textbraceright@pjlab.org.cn}

    \vspace*{-0.18in}

\end{abstract}

\maketitle

\begin{figure}[!b]
\label{fig:scaling}
    \centering
    \vspace*{-0.5in}
    \includegraphics[width=0.98\linewidth]{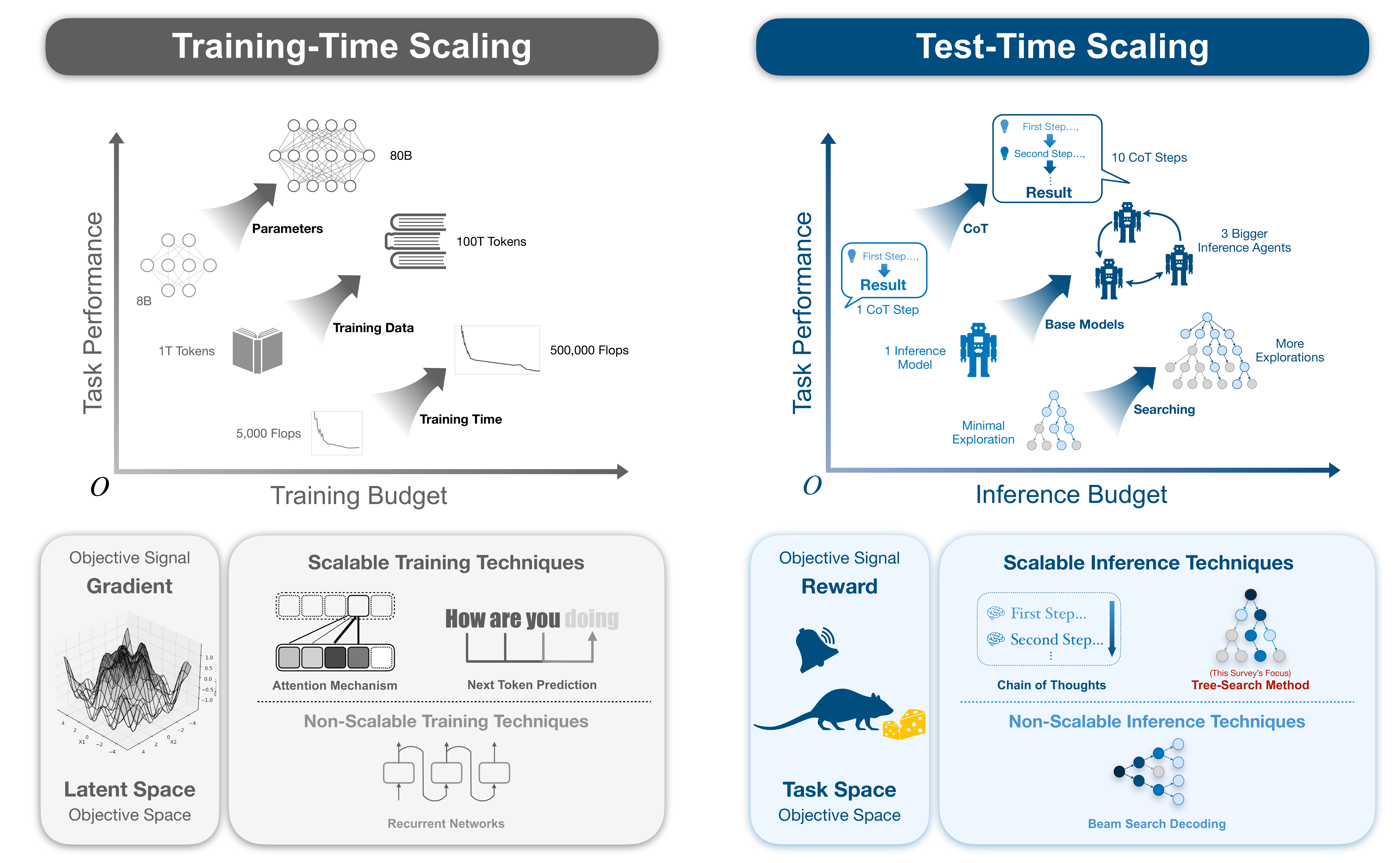}
    % \caption{Comparison of Training-time and Test-time Scaling.}
    
\end{figure}

% \vspace{3mm}
%\pagestyle{headstyle}
\thispagestyle{empty}

% \vspace{2em}
\clearpage
\tableofcontents

\clearpage

\section{Introduction}

As the scaling laws governing Large Language Models (LLMs) reach a regime of diminishing returns~\citep{kaplan2020scalinglawsneurallanguage, hoffmann2022trainingcomputeoptimallargelanguage,wei2025ai,hu2025survey,you2025uncovering}, the research frontier is shifting from brute-force growth in data and parameters toward \emph{algorithmic efficiency} and new forms of reasoning. Two interrelated paradigms have emerged at the core of this transition: \textbf{Test-Time Scaling (TTS)}~\citep{sharma2023revisiting, brown2024large, wu2024scaling, wei2025alignrag, zhang2025postergen}—allocating adaptive, auxiliary computation at inference time to improve problem-solving, akin to the human capacity for deliberate System~2 reasoning~\citep{kahneman2011thinking, Evans1984-EVAHAA, zhao2025timeseriesscientist, yao2023tree,you2025uncovering}—and \textbf{Self-Improvement via Data Generation}, where models construct and refine their own training signals through high-fidelity reasoning traces~\citep{gulcehre2023reinforced, wan2024alphazerolike, guo2025deepseek}. The synergy between these two paradigms defines a crucial direction for advancing the effectiveness and autonomy of modern LLMs.

Within this landscape, \textbf{deliberative search algorithms}, particularly those structured around trees, have become the connective methodology unifying TTS and self-improvement. Tree search offers a principled mechanism to transcend the limitations of greedy, single-path decoding such as standard Chain-of-Thought (CoT) prompting~\citep{wei2022chain}. By systematically exploring and evaluating multiple reasoning branches, whether through explicit search~\citep{yao2023tree, hao-etal-2023-reasoning}, heuristic expansion~\citep{besta2024graph, hu2024treeplanner}, re-ranking~\citep{ni2023lever, li-etal-2023-making}, or iterative refinement~\citep{madaan2023selfrefine, qu2024recursive}, these methods yield substantial gains on complex, multi-step reasoning tasks across diverse domains~\citep{ wang2023selfconsistency, liu2022character, openr1}. This paradigm, often termed \emph{search-augmented reasoning}, forms the conceptual foundation of modern test-time enhancement.

\begin{figure}[!b]
    \centering
    \includegraphics[width=\linewidth]{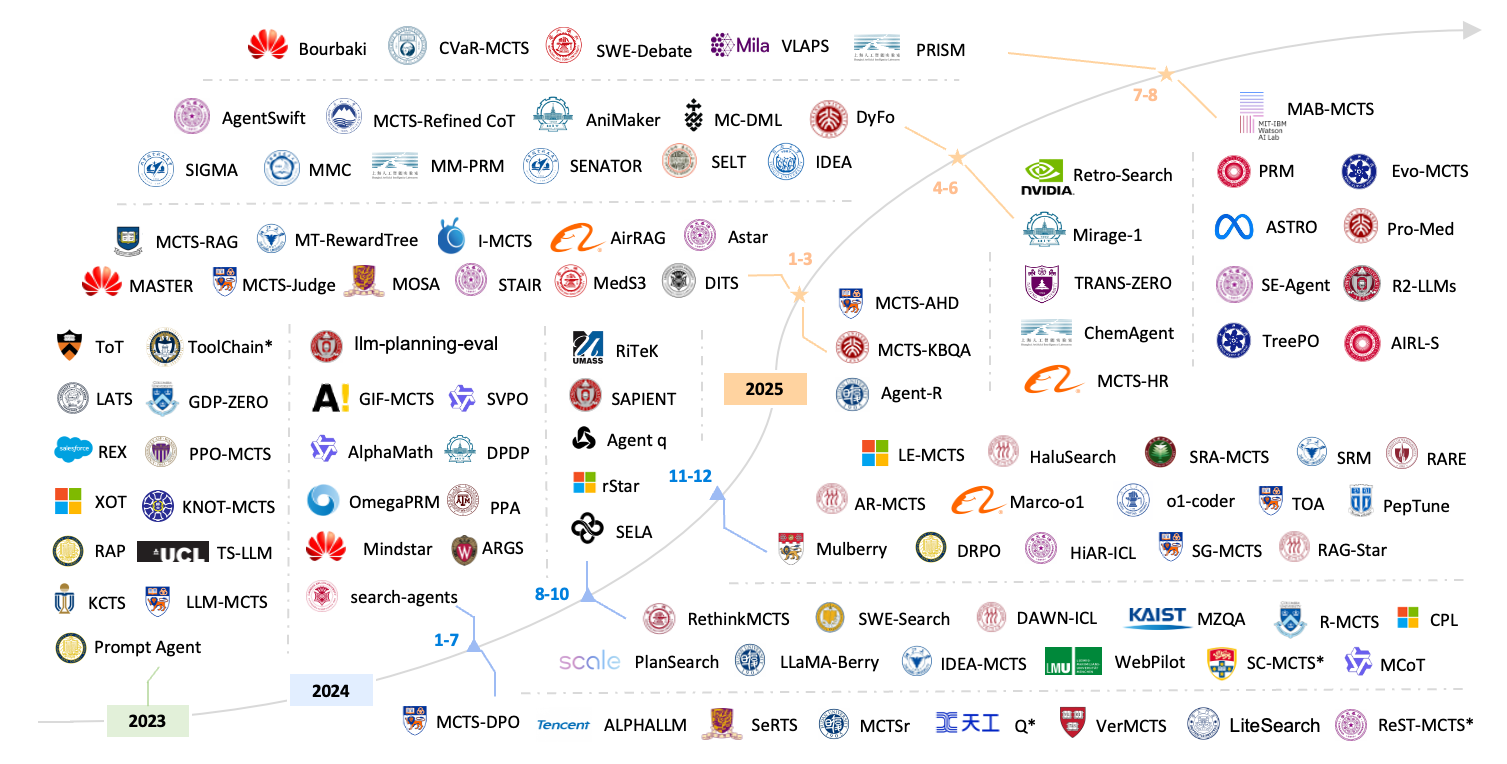}
    \caption{Landscape of research on tree search algorithms and reward design for LLMs.}
    \label{fig:roadmap}
\end{figure}

Concurrently, these same search mechanisms serve as engines for durable, parametric \textbf{Self-Improvement}. Algorithms such as Monte Carlo Tree Search (MCTS)~\citep{Coulom2006EfficientSA, sutton2018reinforcement} excel at navigating vast solution spaces to uncover optimal reasoning trajectories that an LLM might produce only stochastically. By distilling these trajectories into synthetic training data, researchers can fine-tune the base model or train a specialized reward function, effectively \emph{internalizing} high-quality reasoning behaviors. This process transforms costly inference-time deliberation into generalizable parametric knowledge, thereby realizing a self-evolutionary loop of continual improvement~\citep{silver2017mastering, zhang2024restmcts, guo2025deepseek, yuan2024self, huang2024ritek, chen2024alphamath, xie2024monte, tian2024toward, qu2024recursive}.

Despite rapid and decentralized progress across both TTS and self-improvement research, the field remains highly fragmented. The diversity of search paradigms (e.g., Tree-of-Thought~\citep{yao2023tree}, MCTS variants~\citep{hao-etal-2023-reasoning, xie2024monte}, Graph-of-Thought~\citep{besta2024graph}), inconsistent notation, and heterogeneous evaluation protocols have made systematic comparison and cumulative progress difficult (see Figure~\ref{fig:roadmap}). Moreover, the central concept of \emph{reward or value estimation}—often implemented as Process or Outcome Reward Models (PRMs/ORMs)~\citep{lightman2024lets, ouyang2022training, wan2024alphazerolike} which are vital for guiding the search—plays fundamentally different roles in transient test-time reasoning versus persistent parametric optimization, yet this distinction remains underdefined. This conceptual ambiguity hinders the emergence of general design principles and impedes theoretical synthesis.

To address this gap, we present a cohesive conceptual and mathematical framework that unifies the growing body of work on search-based reasoning in LLMs. Our framework aims to clarify core mechanisms, formalize the roles of key components, and establish a rigorous basis for comparing methods across both TTS and self-improvement paradigms. \textbf{Our main contributions are as follows:}
\begin{itemize}
    \item \textbf{A Deconstructive Formalism for Deliberative LLMs:} We introduce a unified mathematical framework that dissects and compares diverse tree search algorithms through their core components: \emph{Search Mechanism}, \emph{Reward Formulation}, and \emph{Transition Function}. We further distinguish the transient role of \textbf{Search Guidance} (for TTS) from the enduring objective of \textbf{Parametric Reward Modeling} (for Self-Improvement).

    \item \textbf{A Systematic, Component-Based Taxonomy:} We propose a novel taxonomy that organizes existing and emerging algorithms along three orthogonal axes—the \emph{Search Mechanism} (e.g., guided vs. unguided exploration), the \emph{Reward/Value Estimation} method, and the overarching \emph{Application Paradigm} (Test-Time Enhancement vs. Self-Improvement).

    \item \textbf{Synthesis and Future Research Agenda:} We synthesize key advances across both paradigms, highlight open challenges in scaling search complexity and designing effective reward signals, and outline a forward-looking research agenda toward truly self-evolving, deliberate LLMs.
\end{itemize}

\vspace{-1.em}
\paragraph{Survey Organization.}
This survey is organized to build a cohesive narrative from foundational search principles to their cutting-edge application in augmenting LLM reasoning. We first lay the groundwork by reviewing classical AI search paradigms (Section~\ref{sec:Paradigms}) and contextualizing them within the modern framework of Test-Time Scaling (TTS) for LLMs (Section~\ref{sec:test_time_scaling}). The intellectual core of our survey is a unified perspective on reward modeling (Section~\ref{sec:reward}), which reconciles the objectives of deliberative search and reinforcement learning. This unified lens allows for a systematic examination of the two dominant families of search algorithms employed today: Monte Carlo Tree Search (MCTS) based methods (Section~\ref{sec:MCTS}) and other informed search strategies (Section~\ref{sec:Informed}). Venturing into an emerging frontier, we then explore a conceptual shift from search over reasoning paths to search over the conditioning context itself in Section~\ref{sec:search_in_prompt_space}, a paradigm we term ``Search in Prompt Space.'' We conclude by synthesizing the current state of the field, outlining critical open challenges and promising future directions in Section~\ref{sec:Challenges}, before summarizing our findings in Section~\ref{sec:Conclusion}.

\section{Foundational Search Paradigms in General AI}
\label{sec:Paradigms}

Solving complex problems can be formalized as a search task: finding an optimal path from an initial state to a goal state within a state-action space, conventionally represented as a tree $T_Q$. While classical AI has developed a rich toolkit for navigating such trees, the state spaces implicit in language model reasoning present unique challenges. They are not merely large; they are combinatorially vast~\cite{gong2025multiprocessor}, high-dimensional, and semantically structured, rendering exhaustive exploration computationally infeasible. This section revisits three foundational paradigms of tree search---uninformed, informed, and Monte Carlo-based---to establish a conceptual vocabulary for understanding their modern adaptations for LLM-based reasoning, where the goal is to identify optimal reasoning paths efficiently.

\subsection{Uninformed Search: Blind Exploration}
Traditional search algorithms, such as Breadth-First Search (BFS) \citep{moore1959shortest}, Depth-First Search (DFS) \citep{tarjan1972depth}, and Uniform Cost Search (UCS), a generalization of Dijkstra's algorithm \citep{dijkstra2022note}, are \textbf{uninformed search} algorithms that operate with minimal knowledge about the goal. These algorithms can recognize the goal state when reached but lack any additional information to guide them toward it efficiently \citep{DBLP:books/aw/RN2020,Poole_Mackworth_2023}. While some uninformed search algorithms, like UCS, consider the cost of the path taken so far, none can estimate the remaining distance to the goal or determine which paths are more promising. 

The key characteristic of uninformed search is that it must rely solely on the problem's basic definition—the available actions, their costs, and the goal recognition criteria—to systematically explore the search space. As a result, these algorithms differentiate between possible solution paths primarily through their order of exploration and accumulated costs. Each algorithm offers different guarantees: BFS finds the shortest path in terms of steps, while UCS finds the lowest-cost path. Additional variants like Depth-Limited Search (DLS) and Iterative Deepening Search (IDS) \citep{korf1985depth} address memory limitations of basic DFS while maintaining completeness. The choice between these algorithms often depends on the problem's characteristics and computational constraints, particularly memory requirements.

\subsection{Informed Search: Heuristic-Guided Exploration}
In contrast to their uninformed counterparts, \textbf{informed search}, or \textbf{heuristic search}, algorithms leverage additional knowledge about the goal's location through domain-specific hints \citep{DBLP:books/aw/RN2020, pearl1984heuristics}. These hints are encoded in a \textbf{heuristic function}, denoted $h(n)$ \citep{Poole_Mackworth_2023}:
\begin{equation}
\label{eq:heuristicdef}
h(n) = \text{estimated non-negative cost of the cheapest path from node $n$ to a goal state}
\end{equation}
Let $c(n,n')$ denote the cost of the path between nodes $n$ and $n'$. By incorporating heuristics, informed search algorithms can make educated decisions about which paths are most promising to explore, potentially reducing the computational resources required to find a solution. The effectiveness and properties of these algorithms depend critically on the quality of their heuristic functions. A heuristic is considered \textit{admissible} if it never overestimates the true cost to the goal, and \textit{consistent} (or \textit{monotone}) if it satisfies the triangle inequality $h(n) \leq c(n,n') + h(n')$ for any successor $n'$ of $n$ \citep{hart1968formal, pearl1984heuristics}.
The choice of heuristic function significantly impacts performance. A heuristic $h_1$ is considered more \textit{informed} than $h_2$ if $h_1(n) \geq h_2(n)$ for all nodes $n$ and $h_1(n) > h_2(n)$ for some nodes. More informed heuristics generally lead to more efficient search, as they provide better guidance toward the goal. 

However, there is often a trade-off between the computational cost of calculating the heuristic and the savings it provides in search efficiency. Common informed search algorithms include Greedy Best-First Search (BeFS), A* Search \citep{hart1968formal}, Weighted A* Search \citep{pohl1970heuristic}, Iterative Deepening A* (IDA*) \citep{korf1985depth}, Beam Search \citep{steinbiss1994improvements}, and Recursive Best-First Search (RBFS) \citep{korf1994best}. These algorithms vary in how they balance the heuristic estimates with path costs, leading to different trade-offs between optimality and efficiency. For instance, A* search, when used with an admissible heuristic, guarantees finding an optimal solution if one exists. The success of these algorithms in practical applications often depends on designing effective problem-specific heuristics. Common techniques for developing heuristics include relaxing problem constraints, using pattern databases \citep{culberson1998pattern}, and learning from experience \citep{samuel1959some, DBLP:books/aw/RN2020}. While informed search algorithms generally outperform uninformed search in practice, their effectiveness relies heavily on the quality of their heuristic functions and the specific characteristics of the problem domain.

\subsection{Monte Carlo Tree Search: Learning from Experience}
Monte Carlo Tree Search (MCTS) was first introduced by \citet{Coulom2006EfficientSA} in the context of computer Go and later gained prominence as a core component of AlphaGo \citep{silver2017mastering}. It is an \textbf{adversarial search} algorithm, which aims to maximize winning probability against an optimal opponent. While adversarial MCTS alternates between players and models opponent responses, the MCTS variant used in LLM's inference-time search is a \textit{single-agent} formulation, where the algorithm explores different action sequences without modeling opposing players \citep{browne2012survey}. This adaptation maintains MCTS's core strengths in balancing exploration and exploitation through statistical sampling, while refocusing the objective from competitive game-playing to finding optimal sequences of actions in a non-adversarial environment.

Inference-time MCTS (hereafter referred to simply as MCTS) retains the four fundamental phases of the original algorithm: selection, expansion, simulation, and backpropagation. During selection, the algorithm traverses the tree using the \textit{Upper Confidence bounds applied to Trees (UCT) policy}, which balances exploration and exploitation by selecting actions that maximize:
\begin{equation}
\label{eq:uct}
a^* = \arg\max_{a\in A(s)}\left[Q(s,a) + c\sqrt{\frac{\ln N(s)}{N(s,a)}}\right]
\end{equation}
where $Q(s,a)$ estimates the expected future reward of taking action $a$ in node $s$, $N(s)$ is the number of times node $s$ has been visited, $N(s,a)$ is the number of times action $a$ has been selected in node $s$, $c$ is an exploration constant, and $A(s)$ is the set of available actions at node $s$ \citep{Kocsis2006BanditBM}. In the expansion phase, new nodes sampled by LLMs (e.g. subsequent steps in reasoning) are added to the tree to gradually build a model of the search space. The simulation phase performs rollouts from leaf nodes using a default policy to estimate long-term rewards, replacing the win/loss outcomes of adversarial MCTS with domain-specific reward measures.

Unlike traditional uninformed search algorithms such as BFS or DFS that systematically explore the state space, MCTS offers a statistical sampling approach that can handle much larger search spaces. Compared to informed search algorithms like A*, which rely on pre-defined heuristics, MCTS builds its evaluation function through experience. This makes it particularly suitable for LLM inference where defining accurate heuristics is challenging. The algorithm's ability to balance between exploration and exploitation, combined with its flexibility in handling large state spaces, makes it a powerful tool for guiding LLM inference, though its effectiveness depends on carefully managing the trade-offs between computational resources and search depth.

\begin{figure}[!t]
    \centering
    \includegraphics[width=\linewidth]{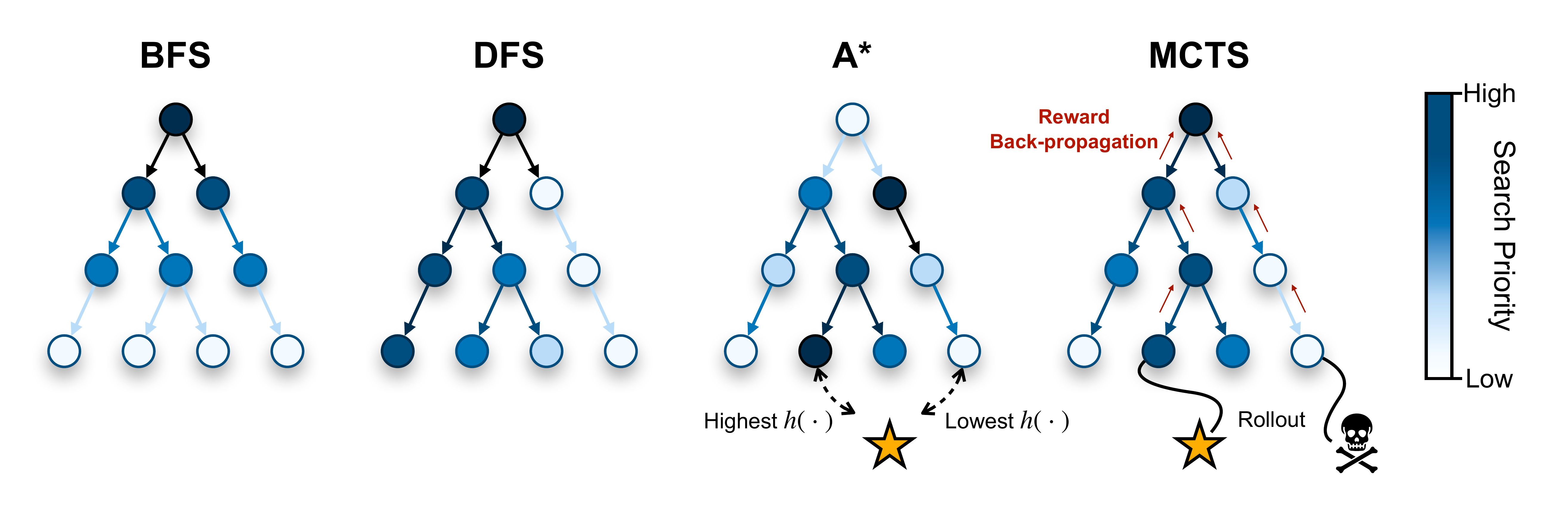}
    \caption{A visual comparison of four fundamental tree search algorithms, where node color intensity represents search priority. \textbf{BFS} explores exhaustively level by level, while \textbf{DFS} commits to a single path until a leaf is reached. In contrast, informed search like \textbf{A*} uses a heuristic function $h(\cdot)$ to prioritize nodes with the lowest estimated total cost, regardless of their depth. \textbf{MCTS} introduces a statistical approach, using simulated rollouts from leaf nodes and backpropagating the outcomes to dynamically guide the search toward high-reward regions of the tree.}
    \label{fig:search}
\end{figure}

\subsection{Comparison of Exploration Strategies}
Figure \ref{fig:search} provides a conceptual illustration of these distinct exploration strategies. Uninformed algorithms like BFS and DFS are governed by rigid, topology-driven expansion protocols. Informed search, exemplified by A*, introduces goal-directedness by prioritizing search based on a heuristic cost-to-go estimate, $h(\cdot)$, allowing it to focus on promising regions irrespective of tree topology. Finally, MCTS replaces the static heuristic with a dynamically learned value function, estimated via statistical sampling. This adaptive, self-correcting mechanism allows it to focus computational resources on the most promising regions of the search space without requiring prior domain knowledge encoded in a heuristic. This very property makes it the preeminent search paradigm for navigating the vast and ill-defined reasoning spaces of large language models.

\section{Test-time Scaling via Search}
\label{sec:test_time_scaling}

As the scaling of model parameters and training data yields diminishing returns \citep{kaplan2020scaling, hoffmann2022training, wei2025ai}, a new frontier has emerged: \textbf{test-time scaling}~\citep{chung2025revisiting}. This paradigm investigates how to optimally allocate computational resources during inference to enhance a model's effective reasoning capabilities \citep{yao2023tree, hao-etal-2023-reasoning, you2025uncovering, wang2023selfconsistency}. Unlike training-time scaling, which refines a global, amortized policy by encoding knowledge into a model's weights, test-time scaling performs instance-specific optimization for a given problem $Q$. This section provides a detailed, mathematically-grounded analysis of these two orthogonal paradigms, contrasting how they operate in fundamentally different optimization landscapes: the latent parameter space for training versus the task-defined objective space for inference~\citep{zhang2025prompt}.

\subsection{A Tale of Two Optimizations for LLM Scaling: Training-Time vs. Test-Time}
The figure referenced illustrates two distinct approaches for improving model performance, each defined by its unique objective signal and the space over which it optimizes.

\vspace{-1.em}
\paragraph{Training-Time Scaling: Optimization in Latent Parameter Space.}
During training, the primary goal is to learn a set of parameters $\theta^*$ that minimizes an expected loss function $\mathcal{L}$ over a data distribution $\mathcal{D}$ \citep{brown2020gpt3}. The optimization problem is formally stated as:
$$
\theta^* = \arg\min_{\theta \in \Theta} \mathbb{E}_{(i,o) \sim \mathcal{D}}[\mathcal{L}(f_\theta(i), o)],
$$
where $\Theta \subseteq \mathbb{R}^N$ is the high-dimensional \textbf{latent parameter space}. The \textbf{objective signal} in this paradigm is the gradient of the loss with respect to the parameters, $\nabla_\theta \mathcal{L}$. Optimization proceeds via iterative updates, such as stochastic gradient descent. The result is a static artifact---a trained model $\pi$---that implicitly represents a posterior distribution over solutions.

\vspace{-1.em}
\paragraph{Test-Time Scaling: Optimization in Task-Defined Objective Space.}
Given a fixed, pretrained model $\pi$, test-time scaling seeks to find an optimal reasoning trace $p^*$ for a specific problem instance $Q$. This process constitutes a second, distinct optimization loop \citep{silver2017mastering, li2022pretrained}. The search occurs in a discrete, structured \textbf{task-defined objective space}, the solution space $\mathcal{P}(Q)$, which consists of all possible reasoning traces. The \textbf{objective signal} is a scalar \textbf{reward} or \textbf{value} that evaluates the quality of a trace. The optimization problem at inference is therefore:
$$
p^* = \arg\max_{p \in \mathcal{A}(\pi, Q, \mathcal{C}_{\text{infer}})} V(p),
$$
where $\mathcal{A}(\pi, Q, \mathcal{C}_{\text{infer}})$ is the search algorithm that explores a subset of $\mathcal{P}(Q)$ guided by the model's prior $\pi$ and constrained by the inference compute budget $\mathcal{C}_{\text{infer}}$, and $V(p)$ is a function evaluating the final trace. Scalable inference techniques, such as tree search, use intermediate rewards $r_s$ or partial trace values $v_i$ to dynamically allocate compute to more promising regions \citep{yao2023tree, hao-etal-2023-reasoning}. The evaluation function $V(p)$ providing these signals can range from heuristics to learned Process or Outcome Reward Models (PRMs/ORMs) \citep{lightman2024lets, wan2024alphazerolike, zhang2024restmcts, guo2025deepseek}.

\subsection{Operationalizing Search in the Objective Space}
The conceptual shift from gradients in latent space to rewards in objective space necessitates a different class of optimization algorithms. While training relies on gradient-based methods, test-time scaling is operationalized by search procedures that can navigate complex, non-differentiable solution spaces \citep{russell2010artificial}.

\vspace{-1.em}
\paragraph{Tree Search as a Scalable Inference Optimizer.}
Tree search methods, particularly Monte Carlo Tree Search (MCTS) \citep{Coulom2006EfficientSA, kocsis2006bandit}, provide a principled framework for this optimization. They build a search tree $T_Q$ where each node $C_i$ corresponds to a partial reasoning trace $p_i$. At each node, an action selection policy balances exploiting known high-reward paths and exploring novel ones. For LLM-based search, this policy often uses a PUCT-style rule that incorporates the policy network's prior, as popularized by AlphaGo \citep{silver2017mastering, silver2017mastering}. The next action $a^*$ is selected by choosing the action that leads to the most promising child node:
$$
a^* = \underset{a \in \mathcal{A}(s_i)}{\arg\max} \left( q_j + U(C_i, C_j) \right),
$$
where $s_i$ is the state at the parent node $C_i$, and action $a$ leads to the child node $C_j$ with quality value $q_j$. The uncertainty bonus $U(C_i, C_j)$ is formulated as:
$$
U(C_i, C_j) = c_{\text{exp}} \cdot \pi(a|p_i, Q) \cdot \frac{\sqrt{n_i}}{1 + n_j}.
$$
Here, $n_i$ and $n_j$ are the visit counts of the parent and child nodes, respectively. The policy $\pi$ provides a prior probability for taking action $a$ given the history $p_i$, and $c_{\text{exp}}$ is an exploration hyperparameter. This synthesis allows the algorithm to scale reasoning performance effectively with the allocated inference compute budget \citep{hao-etal-2023-reasoning, wan2024alphazerolike, zhang2024restmcts, guo2025deepseek}.

\subsection{Decomposing the Objective Space: Prompt and Answer Spaces}

The task-defined objective space, over which test-time search operates, is not monolithic. It can be productively decomposed into two distinct, hierarchically-related search spaces: the \textbf{Prompt Space} (details in Section~\ref{sec:search_in_prompt_space}) and the \textbf{Answer Space} (details in Section~\ref{sec:MCTS} and Section~\ref{sec:Informed}). This decomposition clarifies the mechanisms of Chain-of-Thought (CoT) reasoning \citep{wei2022chain} and reveals the limitations of many current test-time search methods. The overall optimization problem is thus a search for an optimal reasoning trace, which involves finding both the right algorithm and its correct execution.

\vspace{-1.em}
\paragraph{The Prompt Space ($\mathcal{P}$): Searching for an Algorithm.}
The prompt space, $\mathcal{P}$, encompasses the set of all possible reasoning structures or ``step templates'' an LLM can adopt to solve a problem. Each template $p \in \mathcal{P}$ represents a specific strategy for externalizing and manipulating information from the model's latent state $\mathbf{h}$ into its textual output space \citep{zhang2025prompt}. In essence, selecting a template $p$ is equivalent to selecting an \textbf{algorithm}. For example, one template for a complex arithmetic task might involve explicitly tracking a running total, while another might only verbalize intermediate calculations without a canonical state representation. The search for effective prompts or reasoning structures has itself become an active area of research \citep{zhou2023largelanguagemodelshumanlevel, yang2023baichuan}.

The choice of template is paramount because it dictates the computational graph the model simulates through its autoregressive generation. While theoretical work suggests that a CoT-augmented Transformer can be Turing-complete \citep{li2024chain}, this potential is contingent on generating the correct computational trace. An suboptimal template can lead to an inefficient or even intractable search by failing to surface the necessary state information for subsequent steps, effectively breaking the simulated recurrence. The search for an optimal $p^* \in \mathcal{P}$ is therefore a meta-level optimization: discovering the most effective procedure for solving the task instance.

\vspace{-1.em}
\paragraph{The Answer Space ($\mathcal{S}$): Searching for a Solution.}
For any given prompt template $p$, there exists a corresponding answer space, $\mathcal{S}_p$, which contains all possible reasoning traces (i.e., potential solutions) that can be generated by adhering to that template's structure. The complexity of navigating this space is critically conditioned on the choice of $p$. An effective template $p^*$ dramatically prunes the answer space, simplifying the path to a correct solution. Conversely, a poorly chosen template $p'$ can render the answer space vast and unstructured, making the search computationally infeasible even with a large compute budget.

Many contemporary test-time compute methods operate primarily within this second level of the hierarchy. They typically fix a single, heuristically-defined prompt template (e.g., via a generic instruction like ``think step by step'') and then explore the resulting answer space $\mathcal{S}_p$. These approaches can be broadly categorized. Some, like Tree-of-Thought \citep{yao2023tree} and Reasoning as Planning \citep{hao-etal-2023-reasoning}, employ formal search algorithms. Others utilize tree-like or graph-based branching structures but rely on heuristics, aggregation, or sorting rather than explicit search procedures, such as Graph-of-Thought \citep{besta2023graph}, Tree-Planner \citep{hu2024treeplanner}, and Boost-of-Thoughts \citep{chen2024boosting}. Distinct from these incremental exploration methods are frameworks that first generate multiple complete solution candidates and then either re-rank them \citep{li-etal-2023-making, ni2023lever} or iteratively revise them \citep{madaan2023selfrefine, qu2024recursive}. While all these approaches excel at mitigating execution errors and exploring diverse solution paths \textit{within a fixed algorithmic strategy}, they do not address the foundational challenge of selecting the algorithm itself. If the governing template $p$ is flawed, even an exhaustive search of $\mathcal{S}_p$ is unlikely to yield a correct solution.

\vspace{-1.em}
\paragraph{A Unified View of Test-Time Search.}
A comprehensive framework for test-time search must therefore account for the joint optimization over both spaces. The ultimate objective is to discover a solution trace $s^*$ that maximizes the value function $V(\cdot)$, where the search spans all possible traces allowed by all possible templates:
$$
s^* = \arg\max_{p \in \mathcal{P}, s \in \mathcal{S}_p} V(s)
$$
This formulation highlights a critical gap in current research. While significant effort has been invested in optimizing search algorithms within a given answer space $\mathcal{S}_p$, the systematic exploration of the prompt space $\mathcal{P}$ remains a largely open challenge \citep{nye2021show, zhou2023largelanguagemodelshumanlevel}. The true potential of test-time scaling lies not merely in executing a known algorithm more robustly, but in dynamically discovering the most effective algorithm for the specific problem at hand.

% \section{A Unified Perspective on Reward for RL and Search: One Objective, Two Optimizers}

% \section{Reward as a Unified Signal for Policy and Planning}

\begin{figure}[!t]
    \centering
    \includegraphics[width=\linewidth]{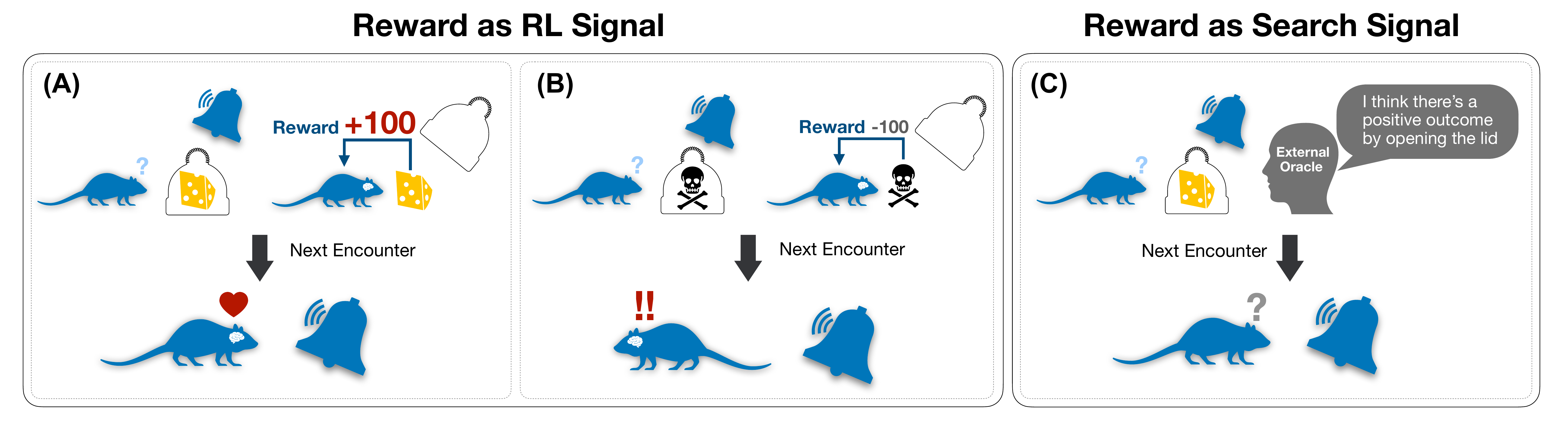}
    \caption{Reward Design: Search vs. RL. (A) In RL, a positive reward updates the agent's policy, making it more likely to repeat the action. (B) A negative reward also updates the policy, discouraging the behavior. The change is \textbf{durable}. (C) In search, an external oracle provides a reward signal to guide the current decision process without altering the agent's underlying parameters.}
    \label{fig:reward}
    % \vspace{-1.2em}
\end{figure}

\section{Reward as a Unified Signal for RL and Search : One Objective, Two Optimizers}
\label{sec:reward}

In advanced AI systems, a reward signal is the fundamental currency for guiding behavior. However, its role bifurcates into two distinct yet complementary functions depending on the temporal scope of the objective: shaping a durable, long-term \textbf{policy} versus guiding a transient, short-term \textbf{plan}. This distinction is not one of paradigm but of application—whether the reward is used to permanently update the model's internal parameters (RL learning) or to direct a temporary search with fixed parameters (planning).

\subsection{RL via Policy Shaping: Internalizing Rewards for Generalization}
When a reward signal is coupled with a learning algorithm, such as in Reinforcement Learning (RL) \citep{sutton2018reinforcement}, its purpose is to be \textbf{internalized}. The feedback from the reward directly modifies the model's weights, creating lasting changes in its behavior. This process is analogous to skill acquisition, where experience is distilled into a robust, general-purpose policy that governs the agent's "instincts" across all future tasks. Formally, this involves optimizing policy parameters $\theta$ to maximize an objective $\mathcal{J}_{\text{RL}}$ that integrates task rewards with adherence to a set of universal principles $\mathcal{P}$, a technique central to modern Reinforcement Learning from Human Feedback (RLHF) \citep{christiano2017deep, ouyang2022training, bai2022training}.

The optimization objective can be expressed as finding the optimal parameters $\theta^*$ that balance expected cumulative rewards $G(\tau)$ over trajectories $\tau$ with a regularization term that enforces alignment with a foundational policy prior $\pi_{\mathcal{P}}$. This is commonly implemented using algorithms like Proximal Policy Optimization (PPO) \citep{schulman2017proximal}:
$$
\theta^* = \arg\max_{\theta} \mathbb{E}_{\tau \sim \pi_\theta} \left[ G(\tau) \right] - \lambda \int_{s \in \tau} D_{KL}\left(\pi_\theta(\cdot|s) \Vert \pi_{\mathcal{P}}(\cdot|s)\right) ds
$$
where $D_{KL}$ is the Kullback-Leibler divergence, measuring the "cost" of deviating from the ingrained principles, and $\lambda$ is a hyperparameter controlling the strength of this alignment imperative. Because this learning is permanent, the reward function is designed to instill \textbf{universal, foundational principles}—for example, promoting logical consistency, ensuring truthfulness, or encouraging methodical, step-by-step reasoning \citep{askell2021general, lightman2024lets}. The objective is not to solve a single problem but to forge a broadly capable and aligned agent. The reward here acts as a long-term teacher, shaping the agent's intrinsic character for future, unseen challenges. Architecturally, this is often achieved by adapting a pretrained LLM into a reward model—such as a Process Reward Model (PRM) or Outcome Reward Model (ORM)—by replacing its final unembedding layer with a multi-layer perceptron (MLP) that outputs a scalar value, a practice standard in RLHF \citep{ouyang2022training, lightman2024lets}.

\subsection{Search via Deliberative Planning: Externalizing Rewards for Specificity}
Conversely, during test-time search, the reward signal functions as an \textbf{external, ephemeral guide}. It directs a deliberative process, like Monte Carlo Tree Search (MCTS) \citep{Coulom2006EfficientSA, kocsis2006bandit}, to navigate the solution space for a single, immediate task. The reward evaluates candidate action sequences (plans), allowing the system to identify a high-quality solution for the specific problem at hand. This approach, famously demonstrated by AlphaGo \citep{silver2017mastering}, uses a fixed model to provide powerful heuristics. This iterative, step-by-step evaluation of intermediate actions distinguishes search-based methods from other test-time compute frameworks like re-ranking, which first generate multiple complete solutions and then score them \citep{ni2023lever, li-etal-2023-making}, or sequential revision, which iteratively refines a complete solution \citep{madaan2023selfrefine}. For a given task with a specific external reward function $R_{\text{ext}}$, the goal is to find an optimal plan $p^*$ that maximizes a combination of this external signal and an internal, path-dependent heuristic $\mathcal{H}_{\theta}$ provided by the frozen model.

The optimal plan $p^*$ for a state sequence $s_0, s_1, \dots, s_T$ resulting from the plan's actions is found by solving:
$$
p^* = \arg\max_{p \in \mathcal{P}_{\text{plan}}} \left[ \sum_{t=0}^{T-1} \gamma^t R_{\text{ext}}(s_t, a_t) + \mathcal{H}_{\theta}(s_T, p) \right]
$$
where the heuristic $\mathcal{H}_{\theta}$ is not just a simple state evaluation but a complex function of the final state $s_T$ and the path $p$ taken, potentially incorporating penalties for path irregularity or deviation from the model's learned priors. This can be viewed through the lens of planning as inference \citep{attias2003planning, toussaint2009robot, botvinick2015reinforcement}, where desirable paths have lower energy or higher probability:
$$
\mathcal{H}_{\theta}(s_T, p) = V_\theta(s_T) - \beta \cdot \log \left( \int_{\tilde{p} \in \mathcal{N}(p)} e^{-\mathcal{E}(\tilde{p})/\tau_c} d\tilde{p} \right)
$$
Here, $V_\theta(s_T)$ is the model's intrinsic value estimate, while the second term acts as a complexity penalty based on the "free energy" over a neighborhood of paths $\mathcal{N}(p)$, discouraging overly surprising or convoluted solutions. Crucially, this feedback is discarded once the task is complete; the model's underlying parameters $\theta$ remain untouched. This makes the reward an ideal tool for \textbf{task-specific, localized objectives} without the risk of corrupting the model's general-purpose policy.

\subsection{A Symbiotic Framework}
Ultimately, policy shaping and deliberative planning are not competing methodologies but two integrated components of a sophisticated decision-making architecture \citep{silver2017mastering}. This mirrors the dual-process theory in human cognition, where a fast, intuitive "System 1" (the policy) is complemented by a slow, deliberate "System 2" (the search) \citep{kahneman2011thinking}. The RL-trained policy provides the foundational intuition, offering high-quality, pre-compiled heuristics that make the search space tractable. Search then provides the focused deliberation needed to refine these intuitions into a precise plan for the current context. This symbiotic relationship can be captured in a single, bi-level optimization objective, where the outer loop learns the policy parameters $\theta$ by anticipating the outcome of the inner-loop search process over a distribution of tasks $\mathcal{I} \in \mathcal{D}$, as exemplified by models like MuZero \citep{schrittwieser2020mastering}. More recent work has explicitly fine-tuned LLMs to improve their synergy with search, training them to function as better policy, value, and reward models within these deliberative frameworks \citep{wan2024alphazerolike, zhang2024restmcts, guo2025deepseek}. This fine-tuning paradigm also extends to other test-time compute methods, such as training models specifically for sequential revision \citep{qu2024recursive}.

The overarching goal is to find policy parameters $\theta^*$ that maximize the true, ground-truth reward $R_{\text{true}}$ of the plans generated by the search algorithm:
$$
\theta^* = \arg\max_\theta \mathbb{E}_{\mathcal{I} \sim \mathcal{D}} \left[ R_{\text{true}} \left( \arg\max_{p \in \mathcal{P}_{\text{plan}}} \left\{ \sum_{t=0}^{T-1} \gamma^t R_{\text{ext}, \mathcal{I}}(s_t, a_t) + \mathcal{H}_{\theta}(s_T, p) \right\} \right) \right]
$$
This formulation reveals the deep connection between the two processes. The outer optimization (learning) seeks to create a model whose internal heuristic, $\mathcal{H}_{\theta}$, is maximally useful for the inner optimization (planning), which in turn must produce plans that score well on the final, external metric $R_{\text{true}}$. In essence, one process builds the artist's foundational skill over a lifetime, while the other guides the brushstrokes for the single masterpiece they are creating now.

\section{Monte Carlo Tree Search (MCTS)}
\label{sec:MCTS}

\begin{figure}[!ht]
    \centering
    \includegraphics[width=0.8\linewidth]{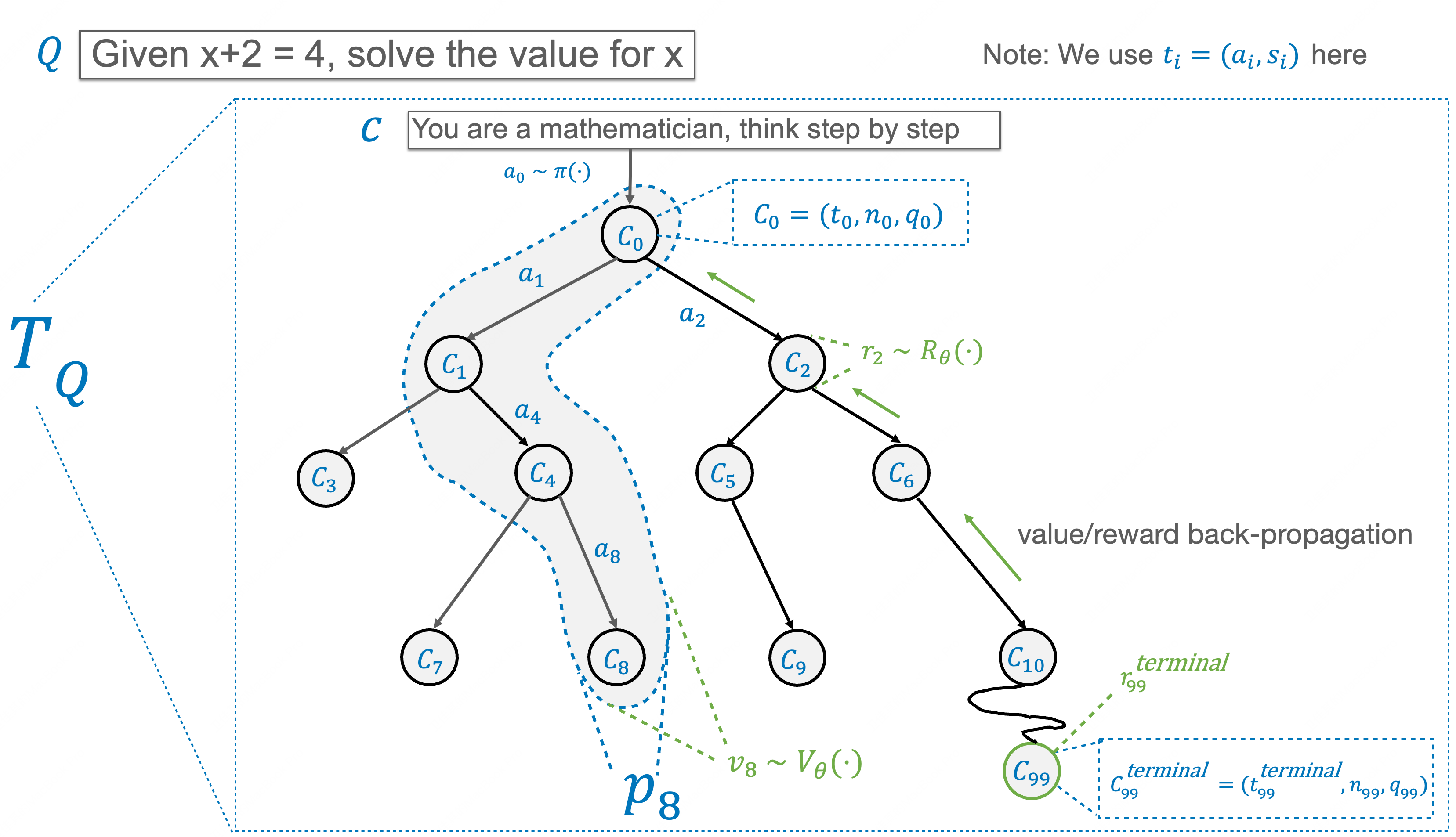}
    \caption{Unified Notations for MCTS-Based Methods in LLM.}
    \label{fig:notations}
    % \vspace{-1.2em}
\end{figure}

\subsection{Unified Notation and Problem Formation}
We adopt the notation conventions introduced in ReST-MCTS$^*$ \citep{zhang2024rest} to formalize MCTS in the context of LLM reasoning in a {\color{red}\textit{unified manner}}. This approach ensures that all the articles surveyed adhere to a {\color{red}\textit{consistent notation system}} (with minor adjustments to accommodate unique designs), allowing for a clear comparison of their methods without the reader having to navigate the discrepancies in notation.

We first introduce the table of notations in Table~\ref{tab:notations} and Figure~\ref{fig:notations}.

\begin{table}[ht]
\centering
\caption{Unified Notations for MCTS-Based Methods in LLM Reasoning}
\begin{adjustbox}{max width=\textwidth}
\begin{tabular}{>{\columncolor{lightblue}\centering}p{0.15\textwidth}>{\columncolor{lightblue}}p{0.8\textwidth}}
\arrayrulecolor{darkblue}
\toprule[1.2pt]
\noalign{\vspace{-3pt}}
\rowcolor{darkblue}
\multicolumn{1}{>{\centering\columncolor{darkblue}}p{0.15\textwidth}}{\textcolor{white}{\textbf{Symbol}}} & 
\multicolumn{1}{>{\centering\columncolor{darkblue}}p{0.8\textwidth}}{\textcolor{white}{\textbf{Definition}}} \\
$Q$ & Input question or problem for which reasoning is being performed \\[0.5em]
$c$ & User prompt or conditioning input used to bias the reasoning traces \\[0.5em]
$a_i$ & Reasoning action at step $i$ generated by the LLM (policy network), where $a_i \in \mathcal{A}$ \\[0.5em]
$s_i$ & Reasoning state at step $i$ resulting from action $a_i$ \\[0.5em]
$p_i$ & Partial reasoning trace up to step $i$, defined as $p_i = [s_1, s_2, \ldots, s_i]$ \\[0.5em]
$r_{s_i}$ & Single-step reward for state $s_i$, measuring its quality independent of previous states \\[0.5em]
$v_i$ & Value of partial solution $p_i$, indicating its potential to reach a correct final answer \\[0.5em]
$T_Q$ & Search tree for problem $Q$, where each node uniquely identifies a reasoning trace \\[0.5em]
$\pi$ & Policy model (LLM) used to generate reasoning steps during tree search \\[0.5em]
$V_\theta$ & Value model that computes partial trace values: $v_i = V_\theta(p_i)$ \\[0.5em]
$R_\theta$ & Reward model that generates single-step rewards: $r_{s_i} = R_\theta(s_i)$ \\[0.5em]
$\mathcal{A}$ & Action space available at state $s_i$, representing all possible next actions \\[1em]
\multicolumn{1}{>{\columncolor{lightblue}\centering}p{0.15\textwidth}}{$C_i$} & 
\multicolumn{1}{>{\columncolor{lightblue}}p{0.8\textwidth}}{\textbf{Search tree node}, represented as $C_i = (t_i, n_i, q_i)$ where:} \\[0.3em]
& $\bullet$ $t_i$: tree node that identifies $C_i$ \\[0.3em]
& $\bullet$ $n_i$: Visit count of node $C_i$, tracking exploration frequency \\[0.3em]
& $\bullet$ $q_i$: Quality value of the partial solution at node $C_i$, indicating its potential to lead to a correct answer \\[-3pt]
\bottomrule[1.2pt]
\end{tabular}
\end{adjustbox}
\label{tab:notations}
\end{table}

With this set of notations defined, a search problem in LLM based reasoning can be generalized as finding the correct solution \textit{or} the optimal reasoning trace $p^\prime = [s^\prime_1, s^\prime_2, \cdots, s^\prime_n] $ for a given problem $Q$.

We categorize approaches for finding correct final solution  (a specific terminal state $s^\prime$)  as goal-driven. Goal-driven methods focus primarily on arriving at the correct final answer for given reasoning problems, paying less attention to the reasoning trace that leads to it. In contrast, approaches that aim to identify good or optimal reasoning steps for a given problem are categorized as step-driven. Step-driven methods not only seek to find the correct solution but also emphasize discovering high-quality intermediate steps that contribute meaningfully to the reasoning process and minimize the reasoning distance.

In the search process, the reasoning LLM acts as a policy network $\pi(\cdot|Q, c)$.  where it generates a sequence of reasoning steps or actions to solve the problem $Q$, under a given instruction prompt $c$. The sequence of generated state-action pairs by $\pi(\cdot|Q, c)$  is denoted as $[s_1, a_1, s_2, a_2, s_3, a_3, \cdots, s_n]$, where $s_1$ is the initial state (often a dummy answer or system prompt) and $s_n$ is the terminal state. The terminal state $s_n$ is reached  when \texttt{[eos]} (i.e. end of sequence) token is produced, which may signify the generation of a final answer (correct or incorrect) or the exhaustion of the step limit (e.g. max context length)~\cite{zhang2024autoregressive+,zhang2025tokenization}.

Note that, unlike most other reinforcement learning (RL) problems, where an action $a_i$ leads to different states $s_{i+1}$ based on a state transition probability, a reasoning action $a_i$ in LLM-based reasoning deterministically leads to a fixed next reasoning state. This deterministic nature is due to the structure of reasoning (with rare exceptions). As a result, we clarify the usage of certain notations, which may differ from those in typical RL formulations:

\begin{itemize}
    \item A reasoning trace, or partial solution, $p_i$, can be expressed in two equivalent forms:
    \[
    p_i = [s_1, a_1, s_2, a_2, s_3, a_3, \ldots, s_i]
    \]
    or 
    \[
    p_i = [s_1, s_2, s_3, \ldots, s_i].
    \]
    The first form treats actions as distinct from states, while the second combines actions and resulting states into $s$. There is no inherent difference between the two representations, as LLM outputs both $s_i$ and $a_i$ into a sentence in each reasoning step during Chain of Thought. Some looks at it separately  (such as RAP) while others take a joint view (such as ReST-MCTS$^*$).

   \item Unlike traditional RL, where the reward is calculated based on the state-action pair, denoted as $R(a, s)$, and depends on the different state transitions resulting from action $a$, the reward of a single LLM reasoning step can be evaluated based on either the action $a_i$ or the resulting state $s_{i+1}$, or even on state action pairs $(s, a)$, due to the deterministic nature of reasoning (each $a$ deterministically determines $s$).

\end{itemize}

For \textit{simplicity}, we typically consider $s_i$ to be a natural language sentence generated as one chain-of-thought (CoT) reasoning step. Consequently, $p_i = [s_1, s_2, s_3, \ldots, s_i]$ represents a CoT trace consisting of $i$ sentences generated in $i$ sequential steps by LLMs.

During reasoning, a given reasoning state $s_i$ can transition to different next reasoning states $s_{i+1}$, deterministically,  depending on the different action $a_i$ that is chosen (from the action space $\mathcal{A}$) by the LLM policy $\pi$, forming a tree structure, denoted as $T_Q$.

Monte Carlo Tree Search (MCTS) optimizes the search for the reasoning trace $[s_1, s_2, \ldots, s_n]$ in $T_Q$ to find correct answers. Each partial solution trace $p_i = [s_1, s_2, \ldots, s_i]$ forms a unique path (or even node) in this tree, associated with its estimated value $v_i$ and visit count $n_i$. The value $v_i$ defines how promising such partial trace is to reach the correct answer. MCTS process is guided by this promising indicator $v_i$. 

Unsurprisingly, the design and computation of $v_i$ become one of the most critical challenges in search algorithm design for LLM reasoning. Our survey places particular emphasis on the methods used to design the value function $V(\cdot)$ in each of the surveyed papers.

All of the search to be discussed here is done in \textit{Answer Space} of problem $Q$, for the discussion of searching in \textit{Prompt Space} of LLM, refer to Section.

\subsection{A 15-Minute Walkthrough of Core Designs with Unified Notation}

\begin{table}[!ht]
\centering
\caption{Comparison of MCTS Node Representations and Evaluations}
\begin{adjustbox}{max width=\textwidth}
\begin{tabular}{@{}lccc@{}}
\toprule
\textbf{Model Name} & \textbf{Tree Node} & \textbf{Node Evaluation} & \textbf{Evaluation Need} \\ \midrule

\textbf{ReST-MCTS*} & 
\({\color{red}p_i} = (s_1, s_2, \ldots, s_i)\) & 
\( {\color{blue}v_i}= V_\text{LLM}({\color{red}p_i})\) & 
\makecell{Current reasoning trace \\ \( p_i = (s_1, s_2, \ldots, s_i\)) } \\ \midrule

\textbf{RAP} & 
\( {\color{red}(a_i, s_i)}\) & 
\makecell{ ${\color{blue}r_i} = R({\color{red}a_i, s_i}) =r_{i,1}^\alpha \cdot r_{i,2}^{1-\alpha} $} & 
\makecell{Current state-action} \\ \midrule

\textbf{LLaMA-Berry} & 
{\color{red}\(s_i^{\text{terminal}}\)} & 
\makecell{\( {\color{blue}r_i} = \alpha R_{\text{local}}({\color{red}s_i^{\text{terminal}}}) + (1-\alpha) R_{\text{global}}({\color{red}s_i^{\text{terminal}}})\)} & 
\makecell{Current and all previously\\ explored  solution nodes in $T_Q$} \\ \midrule
\textbf{MCTSr} & 
\({\color{red}s_i^{\text{terminal}}}\) & 
\makecell{\({\color{blue}r_i} = \frac{1}{2} \left( \min\limits_{\text{$j \in$ evaluating $n$ times}} R^j_\text{LLM}({\color{red}s^\text{terminal}_i}) + \frac{1}{n} \sum_{j=1}^{n} R^j_\text{LLM}({\color{red}s^\text{terminal}_i}) \right)\) \\ re-evaluating \(n\) times to improve robustness.} & 
\makecell{current solution $s^\text{terminal}_i$} \\
\midrule

\textbf{TS-LLM} & 
{\color{red}\( (a_i, s_i)\)} & 
\makecell{
\(  
\begin{cases} 
{\color{blue}v_i} = V_{\text{LLM}}(p_i = (a_1, s_1, \cdots,{\color{red} a_i, s_i})), & \text{if } s_i \neq \text{solution node}, \\ 
{\color{blue}r_i} = R_{\text{ORM}}({\color{red}a_i^\text{terminal}, s_i^{\text{terminal}}}), & \text{if } s_i = \text{solution node}.
\end{cases}
\)} & 
\makecell{All history states \\ $p_i$ = ($s_1$, $s_2$, $\cdots$, $s_i)$ } \\ \midrule

\textbf{ALPHALLM} & 
{\color{red}\( (a_i, s_i)\)} & 
\makecell{
\(  
\begin{cases} 
v_i = V^\text{future}(p_i = (a_1, s_1, \cdots,{\color{red} a_i, s_i}), & \text{if } s_i \neq \text{solution node}, \\ 
{\color{blue}r_i} = R_{\text{PRM}}({\color{red} s_i}), & \text{if } s_i \neq \text{solution node},
\\
{\color{blue}r_i^\text{terminal}} = R_{\text{ORM}}({\color{red} s_i^{\text{terminal}}}), & \text{if } s_i = \text{solution node}.
\end{cases}
\)}
 & 
\makecell{ current state \(s_i\), \\ history $p_i$ of $s_i$} \\ \midrule

\textbf{PG-TD} & 
\( {\color{red}(a_i, s_i)}\) & 
\makecell{
\({\color{blue}r_i} = 
\begin{cases} 
0, & \text{if } s_i \neq \text{solution node}, \\ 
\text{test cases pass rate: TEST($p_i = ({a_1, s_1, \cdots, \color{red}s_i}$) )}, & \text{if } s_i = \text{solution node}.
\end{cases}
\)} & 
\makecell{$p_i = (a_1, s_1, a_2, s_2, \cdots, s_i)$ \\ and test cases provided  } \\
\midrule
\textbf{rStar} & 
${\color{red}(a_i, s_i)}$ & 
${\color{blue}r_i} =
\begin{cases}
0, & \text{if } s_i \neq \text{solution node}, \\
$\text{mutual agreement rate $R_\text{voting}({\color{red}s_i^\text{terminal}})$}$, & \text{if $s_i$ = \text{solution node}}
\end{cases}$ & 
\makecell{Current state-action \\ $(a_i, s_i)$ }\ \\
\midrule
\textbf{RethinkMCTS} & 
$(a_i, s_i)$ & 
${\color{blue}v_i} =
\begin{cases}
\text{test cases pass rate: TEST($p_i = ({a_1, s_1, \cdots, \color{red}s_i}$) ) }, & \text{if } 0 \leq \text{TEST}(p_i) < 1, \\
\alpha \cdot \text{TEST}(p_i) + \beta \cdot V_\text{LLM}(p_i), & \text{if } \text{TEST}(p_i) = 1,
\end{cases}$ & 
\makecell{History  trace $p_i$ and  \\ public test cases } \\ \midrule
\textbf{HiAR-ICL (PRM)} & 
$(a_i, s_i)$ & 
$r_i = R_{\text{PRM}}(s_i)$ & 
\makecell{Current state-action \\ $(a_i, s_i)$ } \\ 
\midrule

\textbf{HiAR-ICL (ORM)} & 
$(a_i, s_i)$ & 
\makecell{
\({\color{blue}r_i} = 
\begin{cases} 
0, & \text{if } s_i \neq \text{solution node}, \\ R_\text{ORM}(s_i^\text{terminal}), & \text{if } s_i = \text{solution node}
\end{cases} \)}& 
\makecell{Current state-action \\ $(a_i, s_i)$} \\ 
\midrule

\textbf{HiAR-ICL (Self-Consistency)} & 
$(a_i, s_i)$ & 
\makecell{
\({\color{blue}r_i} = 
\begin{cases} 
0, & \text{if } s_i \neq \text{solution node}, \\ \text{VOTING}(s_i^\text{terminal}), & \text{if } s_i = \text{solution node}
\end{cases} \)} & 
\makecell{Terminal solution \\ $s_i^{\text{terminal}}$ } \\ 
\midrule

\textbf{Agent-R} &
{\color{red} $s_i$} &
\makecell{Average reward \( {\color{blue}Q(s_i)} \) from rollouts; \\ based on final environmental \\ reward \( {\color{blue}r(\tau)} \in [0, 1] \)} &
\makecell{Final reward \(r(\tau)\) for a \\ complete MCTS rollout} \\ 
\midrule

\textbf{Retro-Search} &
{\color{red}\(s_i\)} &
{\color{blue}\(V(s_i) = \gamma^{N-i}R(a(s_i), a^{*})\)} &
\makecell{Complete trajectory continuation \\ from step \( s_i \) to final answer} \\ 
\midrule

\textbf{MASTER} &
\( {\color{red}s_i} \) &
\makecell{\( ({\color{blue}r_{0,i}}, {\color{blue}c_{0,i}})= V_\text{LLM-self-eval}({\color{red}s_i})\)} &
\makecell{Current agent's full context \\ (Solution + Validation)} \\
\midrule

\textbf{AB-MCTS} &
\({\color{red}t_{out}}\) &
\makecell{ \({\color{blue}p(r | \{r_n\}_{\text{descendants}})}\) \\ (Posterior predictive dist.) } &
\makecell{History of scores from \\ descendant nodes} \\
\midrule

\textbf{SELT} &
\makecell{\(v\) containing state \({\color{red}s}\) \\ which is the reasoning path} &
\( {\color{blue}\Delta}= \text{Score}_{\text{LLM}}({\color{red}s}, s_{\text{represent}})\) &
\makecell{Current answer \({\color{red}s}\) and \\ representative answers \\ from clusters \(s_{\text{represent}}\)} \\
\midrule

\textbf{TRANS-ZERO} &
\({\color{red}y_i}\) &
\({\color{blue}r(y_i)} = \max_{\omega \in \{x_\omega\}}(S(x_\omega, x_{\text{src}}))\) &
\makecell{Original source text \(x_{\text{src}}\) and \\ the set of reconstructions \(\{x_\omega\}\) \\ from multilingual rollouts} \\ 
\midrule

% \textbf{I-MCTS} &
% {\color{red}\(t_i\)} &
% \makecell{\( {\color{blue}q_i} = \alpha_i v_i + (1-\alpha_i) r_i \)} &
% \makecell{Terminal state \\ (full solution code)} \\
% \midrule

\textbf{CMCTS} &
\( {\color{red}s_i}\) &
\makecell{\( {\color{blue}r_i} = Q({\color{red}s_{i-1}, a_i}) + V({\color{red}s_i})\) \\ where Q and V are from PRM} &
\makecell{Current state \(s_i\) and \\ previous state-action \((s_{i-1}, a_i)\)} \\

\bottomrule
\end{tabular}
\end{adjustbox}
\label{tab:mcts_Node}
\end{table}

\begin{table}[!ht]
\centering
\scriptsize
\caption{Comparison of MCTS value $Q$ update and visit count $n$ update}
\label{tab:mcts_evaluations}
\begin{tblr}{
  width = \textwidth,
  colspec = {l c X[1.2, l, m] X[0.8, l, m]},
  row{1} = {font=\bfseries, c}, % 表头加粗居中
  column{1} = {font=\bfseries},   % 模型名称列加粗
  row{2,4,6,8,11,13,15,17,19} = {bg=blockbg}, % 指定行背景色
}
\toprule
Model Name & Tree Node $t_i$ & Update (back-propagate) \(Q_i\) using node value & Update $n_i$ \\ 
\midrule
ReST-MCTS* & \(p_i = (s_1, s_2, \ldots, s_i)\) & $Q^\text{update}_i \leftarrow \frac{\sum\limits_{j \in \text{Children}(p_i)} n_{j} \cdot {\color{blue}v_{j}}}{\sum\limits_{j \in \text{Children}(p_i)} n_{j}}$ & $n_i = n_i + 1 $ \\ 
\midrule
RAP & $(a_i, s_i)$ & $Q^\text{update}_i \leftarrow$ $\max\limits_{\text{each roll out}\,\substack{s_i, a_i, {\color{blue}r_i}, \ldots, s_l, a_l, {\color{blue}r_n}, s^\text{terminal}_{l+1}}} \text{avg}({\color{blue}r_i}, {\color{blue}r_{i+1}}, \ldots, {\color{blue}r_n})$ & $n_i = n_i + 1$ \\ 
\midrule
LLaMA-Berry & $s_i^\text{terminal}$ & \makecell[l]{for $j \in$ Children($s_i^\text{terminal}$) : \\ \qquad $Q^\text{update}_i \leftarrow (1-\gamma)\cdot {\color{blue}Q_i} + \gamma \cdot {\color{blue}r_j}$} & \\ 
\midrule
MCTSr & $s_i^\text{terminal}$ & $Q_i^\text{update} = \frac{1}{2} \left( {\color{blue}Q_i} + \max\limits_{j \in \text{Children}(s_i)} {\color{blue}r_j} \right) $ & \\ 
\midrule
\makecell{TS-LLM (MCTS-$\alpha$) \\ (MCTS-Rollout)} & $(a_i, s_i)$ & \( Q_i^\text{update} \leftarrow
\begin{cases}
Q_i + \gamma {\color{blue}v_n} & \text{if rollout final $s_n$} \neq \text{solution node}, \\ 
Q_i + \gamma {\color{blue}r_n} & \text{if rollout final $s_n$} = \text{solution node}
\end{cases}
\) & \\ 
\midrule
ALPHALLM & $(a_i, s_i)$ & \makecell[l]{$Q_i^\text{update} \leftarrow Q_i + \beta_{\text{v}} \cdot {\color{blue}v_i} + \beta_{\text{PRM}} \cdot {\color{blue}r_i}$ + \\ 
$\beta_{\text{ORM}} \cdot \mathbb{E}_{s^\texttt{terminal}_m \sim \pi_\text{LLM}(s_i)}[{\color{blue}r_m^\text{terminal}}]$ \\ roll-out to terminal node $s_m^\text{terminal}$ n times \\ to estimate expected reward of ORM values.} & \\ 
\midrule
PG-TD & $(a_i, s_i)$ & \makecell[l]{$Q_i^\text{update} \leftarrow \max( Q_i, {\color{blue}r_m})$ \\ where $s_m$ is terminal roll-out state} & \\ 
\midrule
rStar & $(a_i, s_i)$ & \makecell[l]{ $Q_i^\text{update} \leftarrow {\color{blue}Q_i} + {\color{blue}r_m}$ \\ where $s_m$ is terminal roll-out state} & \\
\midrule
RethinkMCTS & $(a_i, s_i)$ & $Q_i^{\text{update}} \leftarrow \max\limits_{\text{j $\in$ Children($(a_i, s_i)$}} (Q_i, {\color{blue}r_j})$ & \\
\midrule
HiAR-ICL (PRM) & $(a_i, s_i)$ & $Q_i^{\text{update}} \leftarrow \alpha \cdot Q_i + (1-\alpha)\cdot \min ( Q_i, r_{i+1} )$ & $n_i \leftarrow n_i + 1$ \\ 
\midrule
HiAR-ICL (ORM) & $(a_i, s_i)$ & \makecell[l]{ $Q_i^\text{update} \leftarrow \alpha \cdot Q_i + (1 - \alpha) \cdot {\color{blue}r_m}$ \\ where $s_m$ is terminal roll-out state} & $n_i \leftarrow n_i + 1$ \\ 
\midrule
HiAR-ICL (Self-Consistency) & $(a_i, s_i)$ & \makecell[l]{ $Q_i^\text{update} \leftarrow \alpha \cdot Q_i + (1 - \alpha) \cdot {\color{blue}r_m}$ \\ where $s_m$ is terminal roll-out state} & $n_i \leftarrow n_i + 1$ \\ 
\midrule
Agent-R & $s_i$ & $Q_i^\text{update} \leftarrow \frac{Q_i \cdot n_i + {\color{blue}r(\tau)}}{n_i + 1}$ & $n_i = n_i + 1$ \\ 
\midrule
Retro-Search & \(s_i\) & \makecell[l]{Greedy trajectory replacement, not value backpropagation. \\ Replaces if \({\color{blue}V(s_\text{new})} > {\color{blue}V(s_\text{old})}\).} & N/A \\ 
\midrule
MASTER & $s_i$ & $Q_i = {\color{blue}c_{0,i}} \cdot {\color{blue}r_{0,i}} + (1-{\color{blue}c_{0,i}}) \cdot \frac{1}{n_i}\sum_{n=1}^{n_i} {\color{blue}r_n}$ & \makecell[l]{Only on backpropagation: \\ $n_i = n_i + 1$} \\
\midrule
AB-MCTS & $t_{out}$ & \makecell[l]{Update posterior parameters. E.g., for Beta dist: \\ $\hat{\alpha} \leftarrow \tilde{\alpha} + \sum {\color{blue}r_n}$, $\hat{\beta} \leftarrow \tilde{\beta} + \sum (1 - {\color{blue}r_n})$} & \makecell[l]{Implicitly tracks number of \\ observations for posterior update} \\
\midrule
SELT & \(v\) containing state \(s\) & \(Q(v) \leftarrow Q(v) + {\color{blue}\Delta}\) & \(N(v) \leftarrow N(v) + 1\) \\
\midrule
TRANS-ZERO & \(y_i\) & $Q_i^\text{update} \leftarrow Q_i + {\color{blue}r_j}$ \text{ (where } \(r_j\) \text{ is from child node)} & $n_i = n_i + 1$ \\
\midrule
CMCTS & $s_i$ & \(Q_i^{\text{update}} \leftarrow \frac{\sum (\sum_{j=i}^{T} {\color{blue}r_j})}{n_i} \) & $n_i = n_i + 1$ \\ 
\bottomrule
\end{tblr}
\end{table}

\begin{table}[!t]
\centering
\scriptsize
\caption{Reward Model Training: Input Generation, Label Construction, and Datasets}
\label{tab:reward_model_training}
\begin{tblr}{
  width = \textwidth,
  colspec = {X[l, m] X[l, m] X[l, m] c c},
  row{1} = {font=\bfseries, c},
  column{1} = {font=\bfseries},
  row{2,4,6,8} = {bg=blockbg},
  % The problematic line "cell{1,1} = {font=\bfseries}," has been removed.
}
\toprule
Model Name &
{Reward Model \\ Input Generation} &
{Reward Label \\ Generation} &
{Reward \\ Model} &
{Reward Model \\ Train Dataset} \\
\midrule
ReST-MCTS* &
{1. SciInstructQuestion, co-training both answers and correct reasoning traces. \\
2. Dataset with only answers; reasoning traces are collected using DFS-based reasoning traces (with Mistral-7B model), generated through breadth-first or depth-first search on verified search trees.} &
{For correct reasoning traces, a \textit{cumulative quality value} \(v_k\) is assigned to each partial trace \(p_k\), computed as \(v_k = k / K\), where \(k\) is the current step, and \(K\) is the total number of steps in the correct reasoning trace. For incorrect traces, \(v_k\) values are penalized or set to 0, reflecting the trace's inability to lead to a correct solution.} &
\(V(p)\) &
{MATH, \\ SciInstructQuestion} \\
\midrule
TS-LLM (Value Model) &
{Rollouts generated using a supervised fine-tuned policy (LLaMA2-7B). \\
Inputs are partial trajectories \(p_i = (s_1, s_2, \dots, s_i)\) from sampled reasoning traces across tasks like GSM8K and Game24.} &
{Target labels are calculated using Temporal Difference (TD-\(\lambda\)) or Monte Carlo (MC) methods. \\
TD-\(\lambda\) uses a weighted sum of \(n\)-step returns and bootstraps with the predicted reward \(V(s_T)\) at the terminal state. MC estimates the full cumulative return directly as the sum of rewards from \(s_i\) to \(s_T\).} &
\(V(p)\) &
{GSM8K, \\ Game24 Rollouts} \\
\midrule
TS-LLM (ORM) &
{Terminal states \(s^{\text{terminal}}\) generated from sampled rollouts of a fine-tuned policy model.} &
{Labels are assigned using a binary reward based on solution correctness and quality. Correct solutions receive \(+1\), while incorrect ones are penalized with \(-1\). \\ Rewards are derived from a task-specific reward function.} &
\(R(s^{\text{terminal}})\) &
{GSM8K, \\ Game24 Rollouts} \\
\midrule
ALPHALLM (Value Model) &
{Reasoning traces are sampled from policy LLM (LLaMA).} &
{Reward \( v_t \) is computed using Temporal Difference (TD) or Monte Carlo (MC) methods by reaching solution nodes and assign scores based on expected correctness of roll-out trace from each node.} &
\( v_{\pi} \) &
{GSM8k Game24 \\ and PrOntoQA} \\
\midrule
ALPHALLM (PRM) &
{reasoning traces are reused from the value function. Inputs are sampled node values \( (s_t, a_t) \).} &
{Immediate rewards \( r^{\text{PRM}}_t \) are assigned using prefix sampling with textual templates for intermediate correctness assessments.} &
PRM &
{GSM8k Game24 \\ and PrOntoQA} \\
\midrule
ALPHALLM (ORM) &
{Terminal states \( s^{\text{terminal}} \) from sampled trajectories are used.} &
{Labels \( r^{\text{ORM}} \) are binary (\(+1\) for correct, \(-1\) for incorrect solutions).} &
ORM &
{GSM8k Game24 \\ and PrOntoQA} \\
\midrule
CMCTS &
{Uses a pre-trained PRM. \\ Input generation is external to the CMCTS framework.} &
{Uses a pre-trained PRM. \\ Label generation is external to the CMCTS framework.} &
{PRM for \\ \(Q(s, a)\) and \(V(s)\)} &
{External to framework \\ (depends on pre-trained \\ PRM, e.g., from [58]).} \\
\bottomrule
\end{tblr}
\end{table}

\begin{center}
\textbf{ReST-MCTS$^*$}
\end{center}

% \subsubsection*{\centering ReST-MCTS$^*$}

%our paper summary focus on its goal and motivation. Novelty and acheivement. 

ReST-MCTS$^*$~\citep{zhang2024rest} adopts a \textbf{\textit{step-driven}} approach, emphasizing the \textbf{\textit{discovery of high-quality reasoning traces and the optimization of intermediate steps}}. Its novelty lies in integrating MCTS with process reward guidance to automatically generate high-quality traces without manual annotation. These traces are then used to iteratively train improved reward and policy (LLM) models. This approach significantly enhances reasoning trace quality, achieving superior performance on datasets such as MATH, GPQA, and CEval compared to baselines that do not leverage MCTS.\\% evaluaotr modeling: Weather same or seperate model (which specific model) used. How its trianed / configured to do evaluation. What it returns, e.g. V or R function, expect to capture what kinda of reward information. Evaluation Frequency.

%Evaluation-Function Design: what new evaluation functions are introduced (stress where is new, compared to standard MCTS calculation), how they affect evaluation score.

%MCTS design, what's something new about selection process, expansion process, score backprob process different from standard MCTS, why such design is adopted.

\textbf{Evaluator-Modeling:}  
A \textbf{\textit{separate}} LLM (Mistral), distinct from the reasoning policy LLM, is used as the value model $V(p_i)$ to \bolditalic{evaluate partial reasoning traces} $p_i = [s_1, s_2, \ldots, s_i]$. The model is fine-tuned on DFS-searched reasoning data with automatically labeled quality scores, capturing the likelihood of $p_i$ leading to a correct solution. During MCTS rollouts, $V(\cdot)$ evaluates each $p_i$ and this value is stored in each node directly. 

\textbf{Evaluation-Function Design:}  
ReST-MCTS$^*$ introduces a weighted reward $w_i$, instead of standard $r_i$, to better reflect the quality and contribution of a single reasoning step (state) $s_i$. The weighted reward is defined as:
\[
w_i = \frac{1 - v_{i-1}}{m_i + 1}(1 - 2r_i),
\]
where $v_{i-1}$ is the value of the previous trace, and $m_i$ is a heuristic measure of the remaining reasoning distance. This approach prioritizes steps closer to the solution, improving exploration of promising paths.

\textbf{MCTS Design:}  
ReST-MCTS$^*$ primarily follows standard MCTS design but introduces a self-critic mechanism during the MCTS expansion phase. Self-critic is also used in deterning termination node in $T_Q$. 

\begin{center}
\textbf{RAP}
\end{center}

% \subsubsection*{\centering RAP}
RAP~\citep{hao2023reasoning} (\textit{Reasoning as Planning with World Models}) adopts a \textbf{\textit{goal-driven}} approach, focusing on efficiently navigating reasoning paths to directly \bolditalic{achieve correct solutions}. This work innovates by re-purposing the \textit{same} LLM as both a reasoning policy (producing action $a_i$) and a \textit{world model} to simulate state transitions (producing $s_i$ and reward returning ($R(a_i)$ model). By combining MCTS with state-action rollouts guided by the world model, RAP demonstrates improved efficiency and accuracy across tasks, achieving strong performance on benchmarks like Blocksworld and logical reasoning datasets while reducing reliance on pretraining or additional reward models.

\textbf{Evaluator-Modeling:}
A \textbf{\textit{repurposed}} LLM  (LLaMA in this work), which is the same model as policy (reasoning) LLM, is used to generate per-step reward. This model generates not only next state $s_t$ but also returns  reward for each newly derived state action pair $(s_t, a_t)$. This world model (same as reward model $R$) capture both action likelihood and task-specific progress when calculating reward $r_i$. Evaluation occurs at each node ($(s_t, a_t)$ in this case) and focuses solely on the current state-action pair, rather than the full trace, ensuring efficient and lightweight evaluation. 

\textbf{Evaluation-Function Design:}  
RAP introduces a novel per-step reward \(r_i\) = $R(s_i, a_i)$ that combines action likelihood and task-specific evaluations to assess the quality of each reasoning step. The reward is calculated using a weighted geometric mean of two components:  
\begin{equation}
    r_t = r_{t,1}^\alpha \cdot r_{t,2}^{1-\alpha}
\end{equation}
where
\(r_{t,1}\) is the log-probability of the action \(a_t\) as predicted by the policy model, reflecting its confidence in the chosen action. \(r_{t,2}\) is a task-specific score evaluating how well the predicted next state \(s_{t+1}\) (generated by the world model) aligns with the task's objectives. This is often derived using heuristics or domain-specific metrics. 

\bolditalic{The value $Q$  of current node is calculated by using the max value of average future reward of each rolled-out future trace starting from this node }:
\begin{equation}
   Q_i = \max_{\substack{s_t, a_t, r_t, \ldots,  s_l, a_l, r_l, s_{l+1}}} \text{avg}(r_t, \ldots, r_l).
\end{equation}
where $s_t, a_t, r_t, \ldots,  s_l, a_l, r_l, s_{l+1}$ is one specific rollout trace starting from current node $(s_t, a_t)$. 
this value $v$ will be used in MCTS selection. 

\textbf{MCTS-design:} RAP uses this $v$ value for selection process in MCTS and sticks with standard MCTS design for expansion and backprogration.

\begin{center}
\textbf{LLaMA-Berry}
\end{center}
% \subsubsection*{\centering LLaMA-Berry}

LLaMA-Berry~\citep{zhang2024llama} is strong \textbf{\textit{goal-driven}} approach, emphasizing the refinement of complete reasoning solutions.  It changes the way of building the reasoning tree $T_Q$ by using only final result  $s^\texttt{terminal}$ as each node, and each child node is a refinement of parent solution. It also innovates the evaluation framework by using\textit{Pairwise Preference Reward Model (PPRM)}, using global win-loss matrices and local adjacency comparisons to rank these solutions. By leveraging critiquing and rewriting during tree expansion, LLaMA-Berry ensures accurate self-refinement of solutions, achieving robust performance across explored solution paths.

\textbf{Evaluator-Modeling: }
A \textbf{\textit{fine-tuned small LLM (Gemma2-2B-Instruct)}} serves as the Pairwise Preference Reward Model (PPRM). PPRM, which is $R_\theta(s^\texttt{terminal})$ in out notation system, predicts the quality ($r_i$) of  single state solutions by comparing directly with other solutions. It is trained using ranked reasoning traces and produces a win-loss matrix \(M(i, j)\), where \(M(i, j)\) encodes the preference between two solutions \(s_i\) and \(s_j\). During evaluation, the PPRM gives a ranked score when each new solution is derived. 

\textbf{Evaluation Function Design:}
LLaMA-Berry introduces an evaluation function that combines a \textbf{local reward} (\(v_{\text{local}}\)) and a \textbf{global reward} (\(v_{\text{global}}\)) of a derived solution $s_i$, defined as:
\[
Q_i = R_{\text{local}}(s_i^{\text{terminal}}) + R_{\text{global}}(s_i^{\text{terminal}})
\]

The local value evaluates the quality of \(s_i^{\texttt{terminal}}\) based on its immediate neighbors in the reasoning tree. Pairwise comparisons are performed with the preceding solution \(s_{i-1}^{\text{terminal}}\) and the subsequent solution \(s_{i+1}^{\text{terminal}}\). Using the Pairwise Preference Reward Model (PPRM), the local value is calculated as:
\[
R_{\text{local}}(s_i^{\text{terminal}}) = \frac{1}{2} \left( \text{PPRM}(s_i^{\text{terminal}}, s_{i-1}^{\text{terminal}}) + \text{PPRM}(s_i^{\text{terminal}}, s_{i+1}^{\text{terminal}}) \right)
\]

The global value assesses \(s_i^{\text{terminal}}\) based on its position within the entire set of explored solutions in $T_Q$. A win-loss matrix \(M(i, j)\) is constructed using pairwise comparisons between all solutions:
\[
M(i, j) = 
\begin{cases} 
1 & \text{if } s_i \text{ is preferred over } s_j, \\
-1 & \text{if } s_j \text{ is preferred over } s_i, \\
0 & \text{if they are equally preferred.}
\end{cases}
\]
The global value is computed using the Enhanced Borda Count (EBC), which aggregates the win-loss scores for each solution:
\[
R_{\text{global}}(s_i^{\text{terminal}}) = \sum_{j} M(i, j)
\]
where the sum is taken over all other solutions \(s_j\) in the reasoning tree.

\textbf{MCTS design:} LLaMA-Berry follows standard MCTS principles with significant enhancements in its \textit{expansion}, \textit{evaluation}, and \textit{backpropagation} processes. During expansion, it employs a critiquing and rewriting mechanism that iteratively refines each solution at the node, improving the quality and accuracy of expanded results. The evaluation phase integrates both global win-loss matrix calculations and local adjacency comparisons to determine the combined quality \(Q_i\) of a solution node. For backpropagation, LLaMA-Berry introduces an updated formula to refine \(Q_i\) by aggregating the values from subsequent child nodes in the tree. This ensures that the backpropagated \(Q_i\) reflects both the cumulative utility of explored paths and the quality of individual solutions. 

\begin{center}
\textbf{MCTSr}
\end{center}
% \subsubsection*{\centering MCTSr}

MCTSr~\citep{zhang2024accessing} is also a strong \textbf{\textit{goal-driven}} approach that emphasizes generating and refining complete reasoning solutions. It structures the reasoning tree \(T_Q\) where each node represents a terminal solution \(s^\texttt{terminal}\), and edges denote iterative refinement attempts. The key innovation in MCTSr lies in its evaluation framework, where rewards for nodes are computed using a combination of minimum and average reward values from multiple resampling attempts. This approach ensures robustness and fairness in evaluating solutions, significantly improving performance in complex reasoning tasks.

\textbf{Evaluator-Modeling:}
MCTSr uses the same LLM (LLaMA) for reasoning and evaluation, where the evaluation model assigns rewards \(R(s^\texttt{terminal})\) to each terminal solutions through a resampling process. The evaluator repeat its rewarding process \(n\) times for each \(s^\texttt{terminal}\), assigning scores between -100 and 100, which are then aggregated to calculate the final node value. This integration avoids the need for a separately trained reward model, instead leveraging the inherent capabilities of the LLM.

\textbf{Evaluation Function Design:}
MCTSr introduces an evaluation function for terminal solutions that combines the minimum reward from resampling and the average reward over all resampled attempts:
\[
Q_i = \frac{1}{2} \left( \min R(s_i^{\texttt{terminal}}) + \frac{1}{n} \sum_{j=1}^{n} R(s_i^{\texttt{terminal}}) \right)
\]
Here, \(R(s_i^{\texttt{terminal}})\) represents the reward assigned to \(s_i^{\texttt{terminal}}\) for a single evaluation, and \(n\) is the number of sampling of rewards attempts. This formula balances the robustness of worst-case performance (\(\min R\)) with the overall quality (\(\text{avg} R\)) of the solution.

\textbf{MCTS Design:}
MCTSr follows the standard MCTS structure with refinements across the selection, evaluation, and backpropagation phases. 
Each child node represents a refined version of its parent solution, generated by iteratively rewriting \(s^\texttt{terminal}\) using the same LLM during expansion phase.
MCTSr employs an updated backpropagation formula that aggregates the values from child nodes, refining the parent node's value \(Q_i\):
\[
Q'(s_i) = \frac{1}{2} \left( Q(s_i) + \max_{j \in \text{Children}(s_i)} Q(s_j) \right)
\]
This formula ensures that parent nodes reflect the best potential of their children while retaining their intrinsic value.
 Dynamic pruning is applied to remove unpromising nodes based on evaluation scores and exploration criteria, improving computation

\begin{center}
\textbf{TS-LLM}
\end{center}
% \subsubsection*{\centering TS-LLM}

TS-LLM~\citep{feng2023alphazero} is both \textbf{\textit{goal-driven}} and Step-driven, leveraging an AlphaZero-inspired MCTS framework to optimize reasoning solutions and better steps to re-train policy LLM. A key innovation in TS-LLM is its dual-model evaluation strategy, where the value model evaluates intermediate reasoning paths based on the entire trajectory \(p_i = (s_1, s_2, \cdots, s_i)\), while the Outcome Reward Model (ORM) scores terminal solutions. This design ensures that the reasoning process balances exploration of promising paths with rigorous evaluation of terminal solutions.

\textbf{Evaluator-Modeling:}
TS-LLM employs two separate models for evaluation:
a) A \textbf{value model}, which predicts \(V(p_i)\), the cumulative reward potential of an intermediate trajectory \(p_i = (s_1, s_2, \cdots, s_i)\);
b) An \textbf{Outcome Reward Model (ORM)}, which assigns a reward \(R_{\text{ORM}}(s_i^{\text{terminal}})\) to terminal nodes based on solution quality.

Both models are trained using supervised learning on collected data from MCTS rollouts. The value model focuses on evaluating reasoning trajectories, while the ORM specializes in scoring terminal solutions. 

\textbf{Evaluation Function Design:}
The evaluation function for a node \(s_i\) is determined by whether it is an intermediate or terminal node:
\[
Q_i = 
\begin{cases} 
V_{\text{value-model}}(p_i = (s_1, s_2, \cdots, s_i)), & \text{if } s_i \neq s_i^{\text{terminal}}, \\ 
R_{\text{ORM}}(s_i^{\text{terminal}}), & \text{if } s_i = s_i^{\text{terminal}}.
\end{cases}
\]
This case-based design ensures that intermediate nodes are evaluated based on their cumulative trajectory, while terminal nodes are directly evaluated for correctness and quality.

\textbf{MCTS Design:}
TS-LLM follows closely standard MCTS design, with a additional pruning mechanism. It dynamically removes unpromising nodes based on updated \(Q_i\) values and visit counts, ensuring efficient resource usage.

\begin{center}
\textbf{\centering ALPHALLM}
\end{center}
% \subsubsection*{\centering ALPHALLM}

ALPHALLM~\citep{tian2024toward} is a strong \textbf{\textit{goal-driven}} framework that integrates an AlphaZero-inspired MCTS with multi-critic evaluation to optimize reasoning paths and terminal solutions. Each node in its tree represents a reasoning step paired with an action, \((s_i, a_i)\), and edges denote transitions between steps. Its core innovation lies in the weighted integration of a value model, process reward model (PRM), and outcome reward model (ORM) to evaluate both intermediate and terminal nodes. ALPHALLM demonstrates superior reasoning capabilities across tasks such as GSM8K and Game24, leveraging its multi-critic evaluation to achieve consistent improvements.

\textbf{Evaluator-Modeling:}  
ALPHALLM employs three distinct evaluation modeling:
The \textbf{value model} predicts \(V^{\text{future}}(p_i)\), capturing the potential reward of intermediate trajectories \(p_i = (s_1, s_2, \ldots, s_i)\). This model is trained using temporal difference (TD-\(\lambda\)) learning and Monte Carlo rollouts to evaluate intermediate states.  
The \textbf{Process Reward Model (PRM)} provides immediate feedback \(R_{\text{PRM}}(s_i)\) for each reasoning step \(s_i\), focusing on local step quality. PRM is fine-tuned on prefixes of reasoning traces using step-level rewards.  
The \textbf{Outcome Reward Model (ORM)} assigns $r = $\(R_{\text{ORM}}(s_i^{\text{terminal}})\) to terminal states based on the correctness and quality of solutions. ORM is trained using solution-specific labels derived from task outcomes.  

This multi-critic design allows ALPHALLM to evaluate reasoning steps both locally and globally, ensuring robust guidance during the search.

\textbf{Evaluation Function Design:}  
The node value \(Q_i\) in ALPHALLM combines contributions from all three critics as :
\[
Q_i = \beta_{\text{v}} \cdot V^{\text{future}}(p_i) + \beta_{\text{PRM}} \cdot R_{\text{PRM}}(s_i) + \beta_{\text{ORM}} \cdot \mathbb{E}_{s^\texttt{terminal} \sim \pi_\text{LLM}(s_i)}[\text{ORM}(s^\text{terminal})]
\]
Where expected values for ORM is calculated by monte carlo sampling to reach terminal (solution) state from current state $s_i$ using policy LLM. 

This formula ensures that the evaluation balances trajectory-level exploration with terminal state quality, facilitating robust reasoning and exploration.

\textbf{MCTS Design:}  
ALPHALLM sticks with standard MCTS process except for evaluation phase as we have discussed above. 

\begin{center}
\textbf{\centering PG-TD}
\end{center}
% \subsubsection*{\centering PG-TD }
PG-TD~\citep{zhang2023planning} adopts a \textbf{\textit{goal-driven}} approach, leveraging a novel integration of Monte Carlo Tree Search (MCTS) and large language models to improve code generation quality. Its primary innovation lies in using test case execution as the evaluation metric during the generation process, rather than relying on deep learning models. By combining the Transformer's beam search probabilities with P-UCB selection for planning, PG-TD achieves significant improvements in code correctness and efficiency compared to standard decoding methods. This framework has shown strong performance across multiple benchmarks, including APPS and CodeContests, particularly in tasks requiring executable and syntactically valid code.

\textbf{Evaluator-Modeling:}
PG-TD does not use a separate deep learning model for evaluation. Instead, it evaluates nodes during MCTS rollouts by executing test cases on the final generated programs. The outcome (pass rate) of these test cases directly determines the reward. This approach simplifies evaluation while aligning it closely with the end goal of generating functional code. Since the evaluation relies purely on test results, no additional value or reward model training is required.

\textbf{Evaluation-Function Design:}
PG-TD evaluates each node in the reasoning tree by running test cases on final complete program represented by the node. The reward for $s^\text{terminal}$ is average pass rate on all test cases.

\textbf{MCTS Design:}
PG-TD introduces several innovations to standard MCTS design. The selection process uses a P-UCB algorithm, which incorporates the probabilities provided by the Transformer's beam search to balance exploration and exploitation effectively. During expansion, the tree grows by selecting the top-k most probable tokens suggested by the Transformer, reducing the likelihood of syntax errors. Additionally, caching mechanisms (tree structure caching and sequence caching) significantly improve efficiency by reusing previously computed paths and evaluations.

\begin{center}
\textbf{\centering rStar}
\end{center}
% \subsubsection*{\centering rStar }
rStar~\citep{qi2024mutual} introduces a novel \textbf{\textit{goal-driven}} framework to enhance the reasoning capabilities of small language models (SLMs) by employing a self-play mutual reasoning paradigm. The core idea is to combine a \textit{generation-discrimination} process, where a target SLM generates reasoning trajectories via MCTS, and another SLM verifies the quality of these trajectories through mutual agreement. This approach is particularly effective in overcoming the limitations of SLMs, such as poor exploration and unreliable self-rewarding. Experiments demonstrate significant performance gains across reasoning tasks, with accuracy improvements on benchmarks like GSM8K, MATH, and StrategyQA, surpassing many fine-tuned models.

\textbf{Evaluator-Modeling:}
rStar employs two SLMs for a collaborative evaluation process. The generator SLM performs MCTS-based trajectory generation, while the discriminator SLM verifies the quality of trajectories. The discriminator applies a mutual reasoning consistency mechanism: it is given partial reasoning traces and asked to complete the remaining steps, validating trajectories based on whether the generator and discriminator agree on the solutions. This unsupervised approach eliminates the need for fine-tuned value models or external supervision.

\textbf{Evaluation-Function Design:}
As SLM does not perform well in partial solution evaluation, rStar applys a AlphaGo-like evaluation framework, where intemediate solutions $p_i$ with $s_i$ gets reward of 0 for simplicity, and reward only assigned if $s^\text{terminal}$ is reached, based on the mutual agreement with discriminator SLM.

\textbf{MCTS Design:}
rStar enhances MCTS with a diverse action space inspired by human reasoning processes. Actions include proposing single steps, generating sub-questions, rephrasing problems, and re-answering sub-questions, enabling broader and deeper exploration of solution trajectories. The P-UCB algorithm balances exploration and exploitation, while mutual consistency during node selection ensures robust trajectory validation.

\begin{center}
\textbf{\centering RethinkMCTS}
\end{center}
% \subsubsection*{\centering RethinkMCTS}

RethinkMCTS~\citep{li2024rethinkmcts} adopts a \textbf{\textit{goal-driven}} approach, enhancing reasoning-to-code performance by leveraging fine-grained feedback and refining erroneous thoughts during the search process. Its novelty lies in combining Monte Carlo Tree Search (MCTS) with a dual evaluation mechanism and introducing a "rethink" operation, which corrects reasoning errors based on execution feedback. This framework significantly improves the quality of search paths and achieves state-of-the-art performance in code generation tasks, with notable gains on benchmarks like APPS and HumanEval.

\textbf{Evaluator-Modeling:}  
RethinkMCTS employs a dual evaluation framework that uses: 1) \textbf{Public Test Cases (denoted as TEST($p_i$))}: The pass rate of public test cases is calculated to assess the correctness of the generated code.
2) \textbf{LLM Self-Evaluation ($V_{\text{LLM}}$)}: When all public test cases are passed, the LLM provides a self-assessment score to further evaluate the likelihood of correctness for private test cases.

Both components are integrated into a unified evaluation system, which ensures more robust assessments for selecting high-quality nodes during tree exploration.

\textbf{Evaluation-Function Design:}  
The evaluation combines scalar and self-assessment rewards to compute the node value $r_i$:
\[
r_i =
\begin{cases}
v_{\text{test}}, & \text{if } 0 \leq v_{\text{test}} < 1, \\
\alpha \cdot v_{\text{test}} + \beta \cdot v_{\text{llm}}, & \text{if } v_{\text{test}} = 1,
\end{cases}
\]
where $\alpha$ and $\beta$ are weighting parameters (e.g., $\alpha = 0.8, \beta = 0.2$). 

The reward ($r_i$) is assigned to each node. (for $p_i$ that's incomplete or incorrect, TEST($p_i$) will always be 0 so there's no need to differentiate the terminal node and intermediate node)

\textbf{MCTS Design:}
RethinkMCTS incorporates several innovations into the MCTS framework. During the selection phase, the P-UCB algorithm is used to balance exploration and exploitation, where verbal feedback stored at nodes influences subsequent thought refinements. In the expansion phase, nodes that fail public test cases incorporate block-level verbal feedback into the prompts, enabling the LLM to propose new thoughts and assign reasonableness scores to each. The rethink operation is employed for leaf nodes that fail public test cases, refining the current thought based on verbal feedback to correct erroneous paths and improve overall search quality. Finally, during backpropagation, node values $Q_i$ are updated using the maximum reward from child nodes, ensuring that the best paths are prioritized for future exploration. Verbal feedback is stored separately and utilized in the next expansion phase but is not directly incorporated into the scalar reward.

\begin{center}
\textbf{\centering HiAR-ICL}
\end{center}
% \subsubsection*{\centering HiAR-ICL }

HiAR-ICL~\citep{wu2024beyond} adopts a versatile and hybrid \textbf{\textit{goal-driven and step-driven}} approach to enhance in-context learning (ICL) by refining reasoning trajectories and leveraging three evaluation strategies: Process Reward Model (PRM), Outcome Reward Model (ORM), and Self-Consistency. This framework introduces hierarchical context construction for iterative refinements and integrates multiple evaluation paradigms to balance intermediate and terminal solution rewards. HiAR-ICL demonstrates significant improvements across reasoning and coding benchmarks, showcasing its adaptability to diverse tasks.

\textbf{Evaluator-Modeling:} HiAR-ICL employs three evaluation mechanisms tailored for different stages of the reasoning process. The Process Reward Model (PRM) evaluates intermediate reasoning steps using a pre-trained language model fine-tuned to assign rewards based on the quality of each step in the trajectory. The Outcome Reward Model (ORM) evaluates terminal solutions using a reward model fine-tuned specifically for outcome-based reasoning tasks. Finally, the Self-Consistency mechanism uses majority voting across multiple reasoning trajectories sampled from the language model to assess terminal solutions. PRM focuses on partial states as inputs and produces scalar intermediate rewards, while ORM and Self-Consistency directly assess terminal states and produce outcome-based rewards.

\textbf{Evaluation-Function Design:} Each evaluation mechanism employs unique reward aggregation strategies. PRM utilizes a Min-based aggregation method where the Process Reward Model evaluates all intermediate steps leading to a state \(s_i\) and calculates the minimum reward across all steps. This emphasizes the weakest link in the trajectory. The reward is defined as:
\[
R_{\text{PRM}}(s_i) = \min_{j=1}^{i} r_{\text{PRM}, j},
\]
where \(r_{\text{PRM}, j}\) represents the reward for the \(j\)-th intermediate step. In contrast, ORM and Self-Consistency directly evaluate terminal solutions using either the Outcome Reward Model or majority voting across sampled traces, respectively, with rewards \(R_{\text{ORM}}(s_m^{\text{terminal}})\) and \(R_{\text{voting}}(s_m^{\text{terminal}})\).

\textbf{MCTS Design:} HiAR-ICL sticks to standard MCTS designs and made changes during backprobgation stages by using a min based value backpropgating approach and a product based reward backpropgating approach.

\begin{center}
\textbf{\centering Agent-R}
\end{center}

The \textbf{Agent-R}~\citep{yuan2025agent} framework is an iterative self-training method designed to improve an agent's ability to recover from errors in interactive environments. Unlike approaches such as RAP that employ an LLM as a learned world model to predict dense, per-step rewards and next states, Agent-R performs post-hoc analysis over complete trajectories to generate corrective training data, turning failure cases into targeted revision examples that refine the underlying policy.

\textbf{Evaluator-Modeling:} Agent-R dispenses with a separately trained reward/evaluator and instead relies on two signals. First, a sparse, terminal \emph{environment reward} $r(\tau)\in[0,1]$ supplies the ground-truth success label for a trajectory $\tau$. Second, the agent's own actor model is repurposed as a critic through ``model-guided critique construction,'' where the policy is prompted to examine a failed trajectory step-by-step, labeling actions as \emph{good}, \emph{bad}, or \emph{uncertain} and identifying the first erroneous step $t'$. This self-critique localizes the source of failure without external experts or a dedicated reward model, in contrast to RAP's dense reward prediction; Agent-R leverages the actor's current competence to diagnose \emph{where} the trajectory went wrong rather than to score \emph{every} step.

\textbf{Evaluation-Function Design:} Full trajectories collected via MCTS are categorized as ``good'' $(\tau^g)$ or ``bad'' $(\tau^b)$ by thresholding their terminal reward, with parameters $\alpha$ and $\beta$ satisfying $r(\tau^b)<\beta<r(\tau^g)\le 1$ and a progressively tightened high-quality bar $\alpha$ (so that $\alpha<r(\tau^g)=r(\tau^r)$ for accepted/revision-worthy traces). The actor-derived transition point $t'$ on a bad trajectory pinpoints the earliest faulty action; Agent-R then splices the correct prefix of $\tau^b$ with the aligned suffix of a good trajectory to form a \emph{revision trajectory} $\tau^r$, e.g., $\tau^r=(\tau^b_{0:t'-1})\circ(\tau^g_{t':m})$. In this way, the evaluation function yields both a coarse trajectory-level judgment and a precise edit location that together produce training pairs emphasizing how to repair failures into successes.

\textbf{MCTS Design:} Agent-R uses Monte Carlo Tree Search not as an inference-time planner but as a data collection engine that systematically explores the action space to yield diverse good/bad trajectories for offline training. The search follows the standard stages (selection, expansion, simulation, backpropagation), with selection guided by the UCT criterion
\[
\mathrm{UCT}(s)=Q(s)+c_{uct}\sqrt{\frac{\log N_p(s)}{N(s)}},
\]
where $Q(s)$ is the average return for state $s$, $N(s)$ its visit count, and $N_p(s)$ the parent's visits; rollouts proceed under a default policy until a terminal state, whose environment reward $r(\tau)$ is then backpropagated to update $Q(s)$ by averaging and to increment $N(s)$ along the traversed path. This usage of MCTS prioritizes breadth and difficulty of experience, furnishing the raw material from which Agent-R constructs revision data that teaches the policy to self-correct.

\begin{center}
\textbf{\centering Retro-Search}
\end{center}

Retro-Search~\cite{lu2025retro} is an MCTS-inspired iterative path revision algorithm rather than a traditional tree search method.

\textbf{Evaluator-Modeling:} The evaluation does not rely on a separately trained reward model. Instead, it uses a ``revision model'' \(\hat{\mathcal{M}}\), which can either be the original reasoning model (in a self-improvement setting) or a weaker student model (in a weak-to-strong setting). This model's role is generative; it produces alternative reasoning trajectories (rollouts) from specific points in an existing path for subsequent evaluation.

\textbf{Evaluation-Function Design:} The quality of a reasoning path is determined by a deterministic value function applied to each step \(s_i\) of the trajectory. The function is defined as \(V(s_i) := \gamma^{N-i}R(a(s_i), a^{*})\), where \(a(s_i)\) is the final answer produced from the trajectory starting at step \(s_i\), \(R\) is a binary function verifying if this answer matches the ground truth \(a^{*}\), \(N\) is the total number of steps in the path, and \(\gamma\) is a decay factor that penalizes longer paths. A new path replaces an old one only if its value is strictly greater, effectively prioritizing shorter, correct solutions.

\textbf{MCTS Design:} The algorithm deviates from standard MCTS by not building an explicit tree or using selection heuristics like UCB. It performs a sequential, greedy revision of a given reasoning trace. The process identifies points where the original model switched its line of thought (e.g., using keywords like ``Alternatively''). At these points, it generates new rollouts by constraining the revision model \(\hat{\mathcal{M}}\) to continue the current thought rather than switching. The resulting trajectory is evaluated, and if it proves more efficient (i.e., has a higher value), it replaces the original path from that point onward. This process repeats for the next thought-switch in the (potentially updated) trajectory. There is no backpropagation of values; decisions are final and greedy.

\begin{center}\textbf{MASTER}\end{center}

\textbf{Evaluator-Modeling.} MASTER~\citep{gan2025master} does not train a separate evaluation model.  Instead, it repurposes the base Large Language Model (LLM) to perform self-evaluation through a structured, multi-step prompting process.  For each generated agent (node), the LLM first executes a \textbf{Validation} step, where it is prompted to verify the key facts within the agent's proposed solution.  Following this, it performs an \textbf{Assessment} step, where it is prompted to generate both a numerical score ($r_0$) indicating progress and a confidence level ($c_0$) for that score. This approach leverages the in-context reasoning capabilities of the LLM itself to serve as the evaluator, avoiding the need for model training and external datasets.  A final \textbf{Evaluation} step is applied only to terminal agents to determine if their solution is correct, which can trigger backpropagation.

\textbf{Evaluation-Function Design.} The evaluation function in MASTER is not a simple reward function but a composite procedure that yields two key values for each agent: an initial reward ($r_0$) and a confidence score ($c_0$). These are extracted from the LLM's textual output during the \textbf{Assessment} phase.  The design is intended to make the reward more reliable by first having the LLM explicitly validate the reasoning steps before assigning a score.  The confidence score ($c_0$) is a crucial component, as it is used to modulate the influence of both the initial reward and the exploration term in the system's modified UCT formula.  For terminal agents that fail the final evaluation, a reward is generated and backpropagated to penalize the preceding reasoning path.

\textbf{MCTS Design.} MASTER introduces a novel adaptation of MCTS tailored for LLMs.  The core modification is the complete \textbf{elimination of the simulation step}, which is traditionally used to estimate long-term rewards. Instead, rewards are derived directly from the LLM's self-assessment at each expansion step. The framework retains the other three MCTS procedures:
\begin{itemize}
     \item \textbf{Selection:} An agent (node) is chosen for expansion based on a modified UCT formula that incorporates the LLM's confidence ($c_0$) to dynamically weigh the initial reward and adjust the exploration term.
     \item \textbf{Expansion:} The selected agent generates a set number of child agents to explore different reasoning paths.
     \item \textbf{Backpropagation:} This step is retained but is only triggered when a terminal agent's final solution fails the evaluation.  The reward from the failed agent is then used to update the Q-values of its ancestors, allowing the system to correct for initially misallocated rewards.
\end{itemize}
This design shifts the computational resources from numerous, costly simulations to a series of refined self-evaluation steps within each node.

\begin{center}\textbf{AB-MCTS}\end{center}

\textbf{Evaluator-Modeling:} AB-MCTS~\citep{inoue2025wider} does not train or model an evaluator. Instead, it presupposes the existence of an external scoring function, $r = R(t_{out})$, which provides direct feedback on a complete, LLM-generated solution candidate $t_{out}$. This function is treated as a black box that returns a score, often normalized to $[0, 1]$, based on task-specific criteria, such as the fraction of passed test cases in a coding challenge. The core method is designed to leverage this external feedback signal for search, rather than modeling the evaluation process itself.

\textbf{Evaluation-Function Design:} The method evaluates actions at a node (either exploring deeper into an existing child's subtree or widening by generating a new child) by modeling the posterior predictive distribution of scores for future nodes. This is implemented in two ways: (1) \textbf{AB-MCTS-M} uses a Bayesian mixed-effects model where each child's subtree is a "group", sharing statistical strength to inform the score distribution of generating a new, unseen child. (2) \textbf{AB-MCTS-A} simplifies this by aggregating all existing children under a `CONT` node and representing new generation with a `GEN` node, modeling the score distribution for each with independent Bayesian models using conjugate priors (e.g., a Beta distribution for scores in $[0, 1]$) for efficient updates. The choice of action is then made via Thompson sampling from these distributions.

\textbf{MCTS Design:} The central innovation is a novel framework that dynamically decides whether to "go wider" (exploration) or "go deeper" (exploitation), enabling adaptive and theoretically unbounded branching. Unlike standard MCTS with a fixed branching factor, AB-MCTS introduces a special `GEN` node at each level of the tree, which represents the action of generating a new child candidate from the current node. The selection policy is not based on UCT but on \textbf{Thompson Sampling}, which naturally balances the choice between selecting an existing child node and selecting the `GEN` node to expand the tree's width based on the Bayesian posterior distributions of expected scores.

\begin{center}\textbf{SELT Introduction}\end{center}

The SELT (Self-Evaluation LLM Tree Search)~\citep{wu2025selt} framework introduces a novel approach to enhance the reasoning capabilities of Large Language Models (LLMs) by integrating a modified Monte Carlo Tree Search (MCTS). Its primary innovation is the elimination of external, pre-trained reward models by leveraging the intrinsic self-evaluation abilities of the LLM itself. By decomposing complex problems into atomic subtasks and employing semantic clustering to guide the evaluation, SELT aims to create a more robust, generalizable, and efficient reasoning process without the need for task-specific fine-tuning.

\textbf{Evaluator-Modeling: }
For its evaluator, SELT repurposes the foundational LLM as an intrinsic, unsupervised \textit{Scorer}, a core design choice that circumvents the dependency on external reward models. This self-evaluation is not performed in a vacuum; its effectiveness is enhanced through a dynamic, reference-based system. At each node in the search tree, SELT performs unsupervised semantic clustering on all previously generated answers to identify distinct, high-quality reasoning paths. From each cluster, a representative answer is selected, and these representatives serve as a contextual benchmark for the LLM Scorer to assess the quality of newly simulated answers.

\textbf{Evaluation-Function Design: }
The evaluation function in SELT produces a reward score, denoted as \(\Delta\), which is generated directly by the LLM Scorer during the simulation phase. Unlike methods that rely on a fixed reward function, SELT's evaluation is dynamic and context-aware. The score \(\Delta\) for a new answer is determined by the LLM's assessment of that answer against the set of representative answers curated through the semantic clustering process. This approach allows the evaluation to be grounded in the diverse and high-quality solutions discovered during the search itself, rather than an abstract or pre-trained notion of correctness. The final score \(\Delta\) is then used in the backpropagation step to update the value of all parent nodes in the traversed path.

\textbf{MCTS Design: }
SELT introduces several significant modifications to the traditional MCTS algorithm. The \textbf{Selection} phase utilizes a custom Upper Confidence Bound for Trees (UCT) formula. The exploitation term, \(S_{LLM\_Exploit}\), is redesigned using Bayesian Averaging to better handle the uncertainty inherent in LLM self-evaluation scores. The exploration term, \(S_{LLM\_Explore}\), is also adjusted to encourage deeper, more focused searches within the reasoning tree. During \textbf{Expansion}, the framework builds out a binary search tree. The \textbf{Simulation} step involves the LLM acting as a `Reasoner` to complete a reasoning path, followed by the clustering and self-evaluation process to generate the reward \(\Delta\). Finally, the \textbf{Backpropagation} phase follows a standard procedure, where the visit count \(N(v)\) and the total reward \(Q(v)\) of each node along the path are updated with the newly calculated score \(\Delta\).

\begin{center}\textbf{TRANS-ZERO}\end{center}

TRANS-ZERO~\citep{zou2025trans} operates as a goal-driven system. The entire search process is optimized to identify the single highest-quality translation for a given source input. Rewards are calculated for complete translation candidates after a comprehensive, multi-step simulation process, rather than being assigned to intermediate steps. The final output of the search is the node (translation candidate) with the highest cumulative utility, reinforcing the framework's focus on achieving a final, high-quality output.

\textbf{Evaluator-Modeling}
TRANS-ZERO does not train a dedicated evaluation model. Instead, it leverages the inherent multilingual capabilities~\citep{zhang2023don} of the base Large Language Model (LLM) for generating translation variations and employs a pre-trained, off-the-shelf text generation metric, BLEURT, to function as the evaluator. This approach bypasses the need for training a separate critic or reward model by defining evaluation as a direct measurement of semantic consistency derived from round-trip translations, making the framework self-contained and reliant only on monolingual data.

\textbf{Evaluation-Function Design}
The evaluation function computes a reward, \(r(y)\), based on the principle of multilingual semantic consistency. For any given translation candidate node, \(y\), the system performs a simulation by rolling out a temporary sub-tree. This involves translating \(y\) through a series of randomly sampled pivot languages and then back to the original source language, generating a set of reconstructions \(\{x_\omega\}\). The final reward is the maximum semantic similarity score, calculated via BLEURT, between these reconstructions and the original source text, effectively measuring how well the meaning is preserved across multiple translation steps.

\textbf{MCTS Design}
The framework introduces a novel Genetic Monte-Carlo Tree Search (G-MCTS), where each node in the tree represents a complete translation candidate. The primary innovation is in the tree expansion phase, which uses two genetic operators to foster diverse exploration. The \textbf{Merge} operator combines the current best-utility node and the best-UCB node as few-shot examples to guide an in-context translation of the original source text. The \textbf{Mutate} operator promotes creative exploration by translating a semantically similar variant of the source text-specifically, a reconstruction generated during a previous simulation-instead of the original input.

\begin{center}
\textbf{CMCTS}
\end{center}

Complementing the constrained action space, CMCTS~\citep{lin2025leveraging} integrates a set of human-like partial order rules during the simulation phase to ensure logical coherence in the reasoning chain. These rules impose constraints on the sequence~\citep{zhang2025pi} of actions, such as mandating an "understand" action at the beginning and a "summary" action at the end. Further rules govern action diversity, the necessity of reflection, and the strategic use of coding actions based on reasoning depth. These rules can be used independently or in combination with the PRM to guide the search, preventing illogical or redundant state transitions. The overall MCTS process follows the standard selection, expansion, simulation, and back-propagation phases, but with these novel constraints and guidance mechanisms integrated to produce higher-quality, long-chain-of-thought reasoning.

\textbf{Evaluator-Modeling}
The CMCTS framework utilizes a pre-trained Process Reward Model (PRM) as its primary evaluator, forgoing the need for training a new model. This PRM, specifically the Qwen2.5-Math-PRM, is designed to assess the quality of intermediate reasoning steps. It functions by taking a given state \(s_t\) and a potential subsequent action \(a_{t+1}\), which are concatenated with a specialized prompt template. The model then processes this combined input to produce logits for "positive" and "negative" outcomes. This mechanism allows the PRM to provide nuanced, context-aware evaluations of reasoning quality without relying on the base Large Language Model (LLM), which is often an unreliable reward signal. The framework uses this external, specialized model to guide the search process toward more rational and effective reasoning paths.

\textbf{Evaluation-Function Design}
The evaluation function in CMCTS is bifurcated into two components: an action-value function \(Q(s_t, a_{t+1})\) and a state-value function \(V(s_t)\), both derived from the PRM. The action-value \(Q(s_t, a_{t+1})\) is calculated by applying a softmax function to the PRM's output logits, yielding the probability of a "positive" assessment for taking action \(a_{t+1}\) in state \(s_t\). Similarly, the state-value \(V(s_t)\) is computed in an action-agnostic manner, representing the intrinsic quality of a given reasoning state \(s_t\). During back-propagation, the reward for a specific transition is defined as the sum \(r_t = Q(s_{t-1}, a_t) + V(s_t)\), which combines the value of the chosen action and the resulting state. This dual-evaluation approach provides a comprehensive signal for updating the cumulative rewards of nodes in the search tree.

\textbf{MCTS Design}
The MCTS design in CMCTS introduces two primary innovations to the standard algorithm: a constrained action space and partial order rules. Unlike traditional methods where the LLM generates subsequent actions, CMCTS samples actions from four predefined, disjoint sets: \(\mathcal{A}^{\text{understand}}\), \(\mathcal{A}^{\text{reflect}}\), \(\mathcal{A}^{\text{code}}\), and \(\mathcal{A}^{\text{summary}}\). This action space constraining, applied during the expansion phase, forces the model to explore diverse and semantically rich reasoning states that are otherwise difficult to sample, such as self-correction and code-based verification. This directly addresses the issue of state-space homogenization common in other MCTS applications with LLMs.

{
\tiny

\begin{tabularx}{\linewidth}{@{} L L L L L L L @{}}
% --- 表格标题与表头 ---
\caption{Comparison of MCTS-based Methods.}
\label{tab:comparison_mcts} \\
\toprule
\textbf{Model Name} & 
\textbf{Policy Model Self-Train} & 
\textbf{Evaluation-Model Training} & 
\textbf{MCTS-eval Algorithm} & 
\textbf{MCTS-eval Model Base} & 
\textbf{Additional MCTS Selection Algorithm} & 
\textbf{Goal-Driven or Reasoning Step-Driven} \\ 
\midrule
\endfirsthead % 这个表头只在第一页显示

\multicolumn{7}{c}%
{{\tablename\ \thetable{} -- continued from previous page}} \\
\toprule
\textbf{Model Name} & 
\textbf{Policy Model Self-Train} & 
\textbf{Evaluation-Model Training} & 
\textbf{MCTS-eval Algorithm} & 
\textbf{MCTS-eval Model Base} & 
\textbf{Additional MCTS Selection Algorithm} & 
\textbf{Goal-Driven or Reasoning Step-Driven} \\ 
\midrule
\endhead % 这个表头会在所有后续页面重复显示

\bottomrule
\endlastfoot % 这个表尾只在表格的最后一页显示

% --- 表格主体内容 ---
\rowcolor{blockbg}
\textbf{ReST-MCTS*} & 
Yes & 
Yes & 
Per-step Process Reward Model (PRM). & 
Separately trained LLM (Mistral); input \(p_i\), returns \(v_i\); using collected reasoning traces. & 
Self-critic mechanism during the selection process. & 
Step-driven \\ 
\midrule

\textbf{RAP} & 
No & 
No & 
Per-step evaluation based on learned policy and dynamics model. & 
Repurposes the same LLM (LLaMA) as a world model; input state-action \((s_t, a_t)\), returns \(r_t\) and \(s_{t+1}\). & 
State exploration guided by predictive heuristics from the dynamics model, focusing on promising states. & 
Goal-driven \\ 
\midrule

\rowcolor{blockbg}
\textbf{LLaMA-Berry} & 
No & 
Yes & 
Pairwise preference comparisons, combining global (win-loss matrix) and local (adjacency) evaluations. & 
Fine-tunes a small LLM (Gemma2-2B-Instruct) as PPRM; evaluates pairwise preferences between reasoning traces. input state $s_i$ and returns score $r_i$ & 
Critiquing and rewriting during expansion to allow more accurate self-refined expanded results. & 
Goal-driven \\ 
\midrule

\textbf{MCTSr} & 
No & 
No & 
Evaluation combines minimum and average rewards sampled reward. & 
Same LLM (e.g., GPT-4) used for reward. Input $s_i$, output $r_i$. & 
Dynamic pruning of unpromising nodes using updated reward backpropagation formula. & 
Goal-driven and Step-driven \\ 
\midrule

\rowcolor{blockbg}
\textbf{TS-LLM} & 
Yes & 
Yes & 
Case-based evaluation depending on node type. & 
Separately trained value model and Outcome Reward Model (ORM) . Input state $s_i$, output $r_i$. & 
AlphaZero-inspired selection process with combined value model and terminal evaluation. & 
Goal-driven \\ 
\midrule

\textbf{ALPHALLM} & 
Yes & 
Yes & 
Evaluation combines weighted contributions from value model, process reward model (PRM), and outcome reward model (ORM). & 
Shared LLM backbone (LLaMA-2-7B) for (multi-critic framework); \(V(p_i)\), PRM, ORM for state-action and terminal evaluation. & 
Combines option-level refinement, and multi-critic evaluation for selecting promising nodes. & 
Goal-driven with multi-level reward integration. \\ 
\midrule

\rowcolor{blockbg}
\textbf{PG-TD} & 
No & 
No & 
Test Case Running. & 
Python/C++ & 
P-UCB selection mechanism with caching to optimize efficiency. & 
Step-driven \\
\midrule

\textbf{rStar} & 
No & 
No & 
Evaluation uses mutual reasoning consistency, where another SLM acts as a discriminator to validate reasoning trajectories. & 
Reuses the same or similar SLM (e.g., Phi3-Mini or LLaMA2-7B); input trajectory, validates partial and full completions. & 
P-UCB selection augmented with diverse reasoning actions inspired by human reasoning, including sub-question generation and rephrasing. & 
Goal-driven \\ 
\midrule

\rowcolor{blockbg}
\textbf{RethinkMCTS} & 
No & 
No & 
Dual evaluation combining public test cases (scalar reward) and LLM self-evaluation (for fully passing nodes). & 
Reuses the same LLM (e.g., GPT-3.5) for self-evaluation; input thought trace $p_i$, returns combined score. & 
P-UCB selection enhanced with verbal feedback during expansion and rethink operations. & 
Goal-driven \\
\midrule

\textbf{HiAR-ICL} & 
No & 
Yes (PRM, ORM), No (Self-Consistency) & 
Process Reward Model evaluates partial reasoning states (PRM); Outcome Reward Model evaluates final reasoning results (ORM); Majority voting over sampled traces (Self-Consistency). & 
Separately trained LLM-based PRM (math-shepherd-mistral-7b-prm) and ORM (Llama3.1-8B -ORM-Mistral-Data; reuses the same LLM for Self-Consistency. & 
P-UCB with hierarchical context construction (PRM, ORM); no additional mechanism for Self-Consistency. & 
Goal-driven \\ 
\midrule

\rowcolor{blockbg}
\textbf{Agent-R} &
Yes &
No &
Final trajectory reward $r(\tau)$ from environment after MCTS rollout. &
External environment's reward function and the policy model itself (Llama-3.1-8B) as a self-critic for revision. &
Standard UCT &
Goal-driven (MCTS exploration) \& Step-driven (self-correction) \\
\midrule

\textbf{Retro-Search} &
Yes &
No &
Length-discounted binary outcome reward. &
Uses the policy LLM (or a weaker one) as a ``revision model'' to generate rollouts. &
Suppresses thought-transition keywords during rollout generation to encourage deeper reasoning paths. &
Goal-driven \\ 
\midrule

\rowcolor{blockbg}
\textbf{MASTER} &
No &
No &
Per-step multi-stage self-evaluation using Validation and Assessment prompts. &
Repurposes the same base LLM (GPT-4); input agent state \(s_i\), returns score \(r_{0,i}\) and confidence \(c_{0,i}\). &
Modified UCT with dynamic exploration weight based on LLM's confidence ($1/c_0$). &
Step-driven \\
\midrule

\textbf{AB-MCTS} &
No &
No &
Bayesian posterior predictive modeling of node scores. &
On-the-fly statistical model (e.g., mixed-effects or conjugate prior models) fitted to observed scores from the tree search. &
Thompson Sampling to decide between "go wider" (new branch) and "go deeper" (refine existing). &
Goal-driven \\
\midrule

\rowcolor{blockbg}
\textbf{SELT} &
No &
No &
Intrinsic self-evaluation. LLM scores a new answer against representative answers from semantic clusters. &
Same foundational LLM (Llama-3.1) used for reasoning. &
Modified UCT with Bayesian Averaging for exploitation and adjusted exploration term. &
Step-driven \\
\midrule

\textbf{TRANS-ZERO} &
Yes &
No &
Multilingual Semantic Consistency via rollouts. &
Pre-trained metric (BLEURT); input sentence pair, returns similarity score. &
Genetic Expansion (Merge/Mutate operators). &
Goal-driven \\
\midrule

\rowcolor{blockbg}
\textbf{CMCTS} &
No &
No &
Per-step Process Reward Model (PRM). &
Uses a pre-trained LLM (Qwen2.5-Math-PRM); input \((s_t, a_{t+1})\) returns Q-value; input \(s_t\) returns V-value. &
Constrained Action Space (during expansion) and Partial Order Rules (during simulation). &
Step-driven \\

\end{tabularx}
}

\begin{figure*}[!b]
    \vspace{-2mm}
    \centering
    \resizebox{0.96\textwidth}{!}{
    \begin{forest}
        forked edges,
        for tree={
            grow=east,
            reversed=true,
            anchor=base west,
            parent anchor=east,
            child anchor=west,
            base=left,
            font=\large,
            rectangle,
            draw=hidden-black,
            rounded corners,
            align=left,
            minimum width=4em,
            edge+={darkgray, line width=1pt},
            s sep=3pt,
            inner xsep=2pt,
            inner ysep=4pt,
            line width=1.1pt,
            ver/.style={rotate=90, child anchor=north, parent anchor=south, anchor=center},
        },
        where level=1{text width=9.5em,font=\normalsize,}{},
        where level=2{text width=11.5em,font=\normalsize,}{},
        where level=3{text width=12em,font=\normalsize,}{},
        % Level 4 styling is now handled by leaf3
        [Taxonomy, ver
            [MCTS for \\ Direct Test-Time \\ Enhancement
                [General Reasoning \\ \& Problem Solving
                    [{ 
                        Discriminator-based Tree Search~\citep{chen2024tree}, 
                        Interpretable Contrastive MCTS~\citep{gao2024interpretable}, \\
                        RoT~\citep{hui2024rot}, 
                        LiteSearch~\citep{wang2024litesearch}, 
                        MindStar~\citep{kang2024mindstar}, \\
                        MARCO-O1~\citep{zhao2024marco}, 
                        Everything of Thoughts~\citep{ding2023everything}, 
                        CoAT~\citep{pan2025coat} 
                    }, leaf3]
                ]
                [Mathematical Reasoning
                    [{ 
                        MCTS Self-Refine~\citep{zhang2024accessing}, 
                        Energy Function MCTS~\citep{xu2023no}, 
                        OVM~\citep{yu2023ovm},  \\
                        LLaMA-Berry~\citep{zhang2025llama}, 
                        Automated Process Supervision~\citep{luo2024improve}, \\
                        Constrained MCTS~\citep{lin2025leveraging},  
                        Markov Chain of Thought~\citep{yang2024markov} \\
                    }, leaf3]
                ]
                [Code Generation \\ \& Software Engineering
                    [{ 
                        RTL Code Gen MCTS~\citep{delorenzo2024make}, 
                        RethinkMCTS~\citep{li2024rethinkmcts},  
                        Verified Multi-step Synthesis~\citep{brandfonbrener2024verified}, \\
                        VerMCTS~\citep{brandfonbrener2024vermcts},   
                        Code World Models MCTS~\citep{dainese2024generating}, 
                        Planning in NL~\citep{wang2024planning},  \\
                        O1-Coder~\citep{zhang2024o1}, 
                        SRA-MCTS~\citep{xu2024sra}, 
                        SWE-Search~\citep{antoniades2024swe}, \\
                        SWE-Debate~\citep{li2025swe},  
                        MCTS-Judge~\citep{wang2025mcts},
                        APRMCTS~\citep{hu2025aprmcts}, 
                        MCTS-Refined CoT~\citep{wang2025mcts}
                    }, leaf3]
                ]
                [LLM Agents \\ \& Interactive Environments
                    [{ 
                        LATS~\citep{zhou2023language}, 
                        SeLa~\citep{chi2024sela}, 
                        BIDA~\citep{yu2025bida},
                        WebPilot~\citep{zhang2025webpilot}, 
                        Prompt-based MCTS~\citep{yu2023prompt}, \\
                        MASTER~\citep{gan2025master}, 
                        SE-Agent~\citep{lin2025se}, 
                        WebSynthesis~\citep{gao2025websynthesis}, 
                        AgentSwift~\citep{li2025agentswift}, 
                        HALO~\citep{hou2025halo} 
                    }, leaf3]
                ]
                [Knowledge-Intensive \\ \& RAG Tasks
                    [{ 
                        KNOT-MCTS~\citep{wu2023knot}, 
                        Search-in-the-Chain~\citep{xu2024search}, \\
                        Contrastive RAG~\citep{gu2025toward}, 
                        RARE~\citep{tran2024rare}, 
                        CoRaG~\citep{wang2024corag}, 
                        RITEK~\citep{huang2024ritek}, \\
                        AirRAG~\citep{feng2025airrag}, 
                        KBQA-O1~\citep{luo2025kbqa}, 
                        MCTS-KBQA~\citep{xiong2025mcts}, 
                        Hierarchical RAG~\citep{dou2025enhancing}, \\ 
                        Explainable MCTS~\citep{kowalski2025towards}, 
                        KCTS~\citep{choi2023kcts},
                        RAG-Star~\citep{jiang2024rag},
                        FREESON~\citep{kim2025freeson} 
                    }, leaf3]
                ]
                [Multimodal Reasoning
                    [{ 
                        Mulberry~\citep{yao2024mulberry}, 
                        Progressive Multimodal Reasoning~\citep{dong2024progressive}, \\
                        MCTS-Automated Structured Thinking~\citep{wu2025boosting}, 
                        Dyfo~\citep{li2025dyfo} 
                    }, leaf3]
                ]
            ]
            [MCTS for \\ Self-Improvement \\ via Data Generation
                [ Self-Improvement \\ Foundational Frameworks
                    [{rStar-Math~\citep{guan2025rstar}, 
                    Alphazero-like tree-search~\citep{feng2023alphazero}, \\
                    Curriculum Preference Learning~\citep{wang2024towards}, 
                    Agent Q~\citep{putta2024agent},\\
                    Iterative Preference Learning~\citep{xie2024monte}, 
                    Imagination, Searching, and Criticizing~\citep{tian2024toward}, \\
                    Mutual Reasoning~\citep{qi2024mutual}, 
                    AlphaMath Almost Zero~\citep{chen2024alphamath}, 
                    CPL~\citep{wang2024cpl}, \\
                    Step-level Value Preference Optimization~\citep{chen2024step}, TreeRPO~\citep{yang2025treerpo}, \\
                    Data Influence-Oriented Tree Search~\citep{shi2025efficient}, 
                    MCTS-Refined CoT~\citep{wang2025mcts},
                    ASTRO~\citep{kim2025astro}, 
                    % Agent-R~\citep{yuan2025agent},
                    }, leaf ]
                ]
                [General Capabilities \\ \& Alignment
                    [{PPL-MCTS~\citep{chaffin2021ppl}, Value-Guided MCTS~\citep{liu2023don}, ARGS~\citep{khanov2024args},  \\ Reflective Tree Search~\citep{yu2024improving}, PromptAgent~\citep{wang2023promptagent}, STAIR~\citep{zhang2025stair}, \\ Dynamic Rewarding~\citep{singla2024dynamic},   Evolutionary Space Search~\citep{li2024optimizing},  APRMCTS~\citep{hu2025aprmcts}}, leaf ]
                ]
                [Scientific \\ \& Specialized Domains
                    [{Self-Play Approach~\citep{guo2024can}, Named Entity Matching~\citep{volkova2024novel},  \\Synthetic Data Generation~\citep{locowic2024synthetic},  
                    Monte Carlo Thought Search~\citep{sprueill2023monte},  \\
                    STRATEGIST~\citep{light2024strategist}, Fast and Slow Thinking~\citep{cheng2025think},  Peptune~\citep{tang2025peptune}, \\
                    Multi-Agent Sampling~\citep{ye2024multi}, Step-level~\citep{ma2025step}, 
                    Process Reward-guided Tree Search~\citep{park2024ensembling}, \\{Think\&Cite}~\citep{li2024think},  
                    SAPIENT~\citep{du2024sapient}, Automatic Heuristic Design~\citep{zheng2025monte}, Trans-Zero~\citep{zou2025trans}, \\
                    Prompt-Based MCTS~\citep{duan2025prompt}, {MedS$^3$}~\citep{jiang2025meds},  MCTSr-Zero~\citep{lu2025mctsr}, K-MSE~\citep{zhuang-etal-2025-boosting} \\   Stepwise Domain Knowledge-Driven Reasoning~\citep{liu2025towards}
                    IRIS~\citep{garikaparthi2025iris}, 
                    ChemAgent~\citep{wu2025chemagent}}, leaf ]
                ]
                [Multimodal Applications
                    [{MCTS-guided Sample Selection~\citep{wang2025sota}, MMC~\citep{liu2025mmc}, MM-PRM~\citep{du2025mm}}, leaf ]
                ]
            ]
            [Advanced Topics and \\ Hybrid Approaches
                [Multi-Agent and \\ Collaborative Search
                    [{Reflective Tree Search~\cite{yu2024improving}, 
                    Multi-agent sampling~\cite{ye2024multi}, \\
                    Process reward-guided tree search~\cite{park2024ensembling}, Webpilot~\cite{zhang2025webpilot}, 
                    MASTER~\cite{gan2025master}, \\
                    Data Influence-Oriented Tree Search~\cite{shi2025efficient}, 
                    Multi-LLM collaborative search~\cite{yang2025multi}, \\
                    KompeteAI~\cite{kulibaba2025kompeteai}, 
                    SWE-Debate~\cite{li2025swe}, 
                    AniMaker~\cite{shi2025animaker}, 
                    HALO~\cite{hou2025halo}}, leaf4]
                ]
                [Reward Model Design \\ and Optimization
                    [{rStar-Math~\cite{guan2025rstar}, 
                    AlphaZero-like tree-search~\cite{feng2023alphazero}, \\
                    Iterative preference learning~\cite{xie2024monte}, 
                    Step-level q-value models~\cite{zhai2025enhancing}, \\
                    Curriculum Preference Learning~\cite{wang2024towards},
                    Interpretable contrastive MCTS~\cite{gao2024interpretable}, \\
                    Energy function guided MCTS~\cite{xu2023no}, 
                    AlphaMath Almost Zero~\cite{chen2024alphamath}, \\
                    CPL~\cite{wang2024cpl}, 
                    OVM~\cite{yu2023ovm}, 
                    Value-Guided MCTS~\cite{liu2023don},\\
                    Step-Level reward model~\cite{ma2023let}, 
                    Step-level value preference optimization~\cite{chen2024step}, \\
                    Automated process supervision~\cite{luo2024improve}, 
                    MCTS-boosted mathematical reasoning~\cite{ma2025step}, \\
                    {Think\&Cite}~\cite{li2024think}, 
                    STAIR~\cite{zhang2025stair}, 
                    MT-RewardTree~\cite{feng2025mt}, \\
                    Hierarchical Multi-Step Reward Models~\cite{wang2025towards}, 
                    ProMed~\cite{ding2025promed}, \\
                    TreeRPO~\cite{yang2025treerpo}, 
                    Unifying RL and Search-Based TTS~\cite{jin2025your},\\
                    Re-ranking Reasoning Context with Tree Search~\cite{yang2025re}
                    }, leaf4]
                ]
                [Search Efficiency \\ and Dynamics
                    [{Tree search for agents~\cite{koh2024tree}, 
                    Discriminator-dependent tree search~\cite{chen2024tree}, \\
                    Information Directed Tree Search~\cite{chandak2024information}, 
                    RoT~\cite{hui2024rot}, 
                    LiteSearch~\cite{wang2024litesearch}, 
                    BoostStep~\cite{zhang2025booststep}, \\
                    Everything of Thoughts~\cite{ding2023everything}, 
                    Dual process of fast and slow thinking~\cite{cheng2025think}, \\
                    T-SCEND~\cite{zhang2025t}, 
                    BFS-Prover~\cite{xin2025bfs}, 
                    MCTS-Judge~\cite{wang2025mcts}, \\
                    Adaptive branching tree search~\cite{inoue2025wider}, 
                    Retro-Search~\cite{lu2025retro}, \\
                    Bilevel MCTS~\cite{asai2025bilevel}, 
                    Test-Time Depth Adaptation~\cite{li2025skip}, \\
                    Abstraction dropping methods~\cite{schmocker2025time}, 
                    SIGMA~\cite{ren2025sigma}, 
                    AgentSwift~\cite{li2025agentswift}, \\
                    FREESON~\cite{kim2025freeson}, 
                    Structural Entropy Guided Agent~\cite{wei2025structural}
                    }, leaf4]
                ]   
            ]
        ]
    \end{forest}
    }
    \caption{A comprehensive taxonomy of MCTS.}
    \label{fig:MCTS-taxonomy}
\end{figure*}
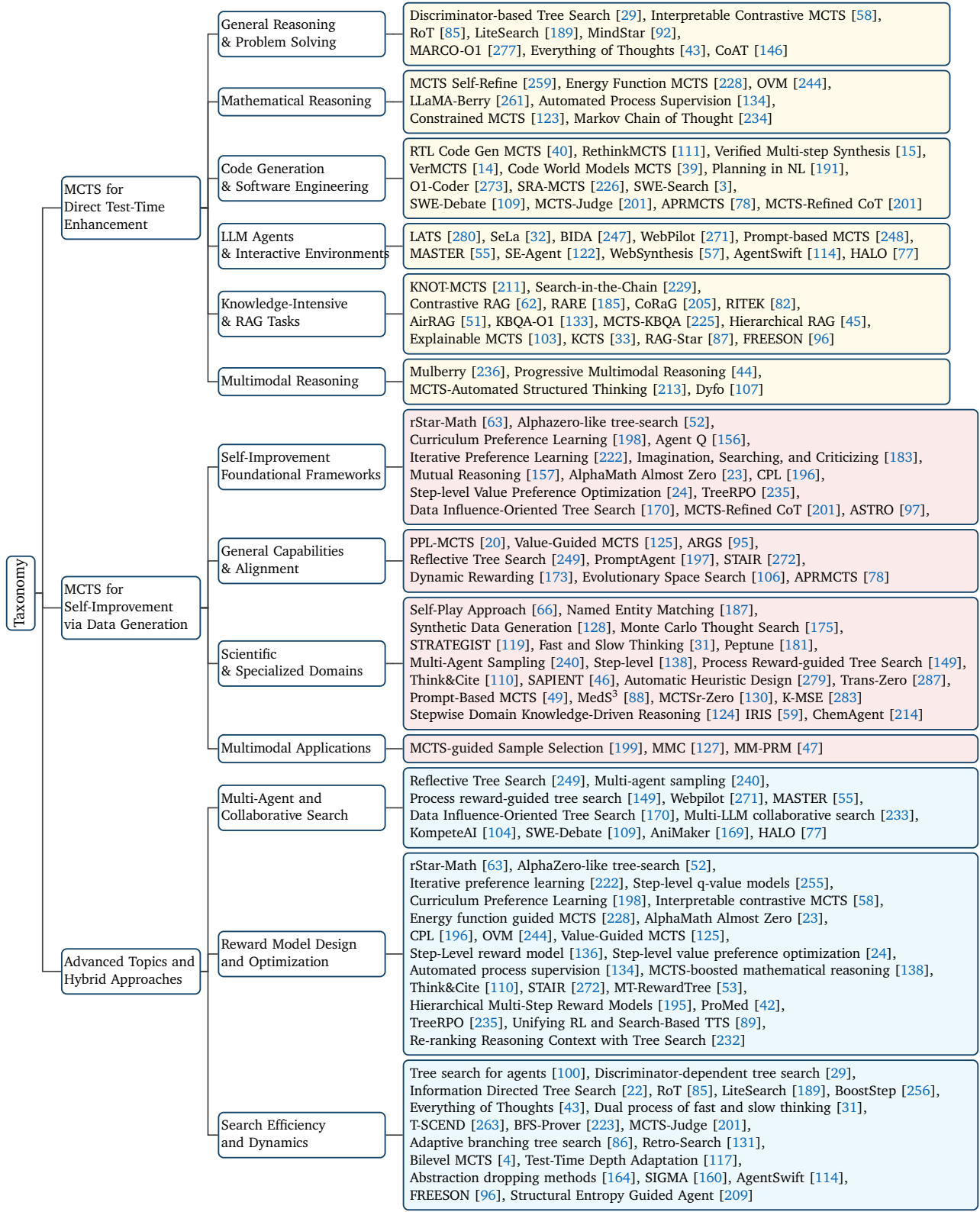

\subsection{Advanced Topics and Hybrid Approaches for MCTS}

As the field of LLM-driven search matures, research is advancing beyond the core MCTS paradigm to explore more sophisticated techniques. These efforts focus on refining the search process through collaborative multi-agent systems, designing more effective reward signals, and enhancing computational efficiency and adaptability.

\subsubsection{Multi-Agent and Collaborative Search}
Rather than relying on a single large language model (LLM) to conduct search, recent approaches employ multiple LLM agents that collaborate, debate, or assume specialized roles to address complex problems more effectively. This paradigm shift—from a monolithic searcher to a coordinated team—enables greater robustness, diversity, and adaptability in problem-solving. Some frameworks leverage MCTS to orchestrate multi-agent interactions, dynamically adjusting agent numbers and communication strategies according to task complexity \cite{gan2025master, ye2024multi, zhang2025webpilot}. Others adopt hierarchical structures comprising specialized agents for high-level planning, role assignment, and low-level execution \cite{hou2025halo, li2023camel}.

A notable strategy is self-play or mutual reasoning, in which one agent generates solutions while another critiques or refines them. This adversarial or cooperative dynamic enhances reasoning quality and filters erroneous paths. In competitive domains such as software issue resolution, debate-based frameworks foster diverse reasoning trajectories and promote more convergent, high-quality outcomes \cite{li2025swe, du2023improving, antoniades2024swe}. For instance, the rStar framework uses one LLM to produce reasoning trajectories via MCTS and another as a discriminator to ensure mutual consistency \cite{qi2024mutual}. Similarly, Reflective MCTS (R-MCTS) integrates debate mechanisms to improve state evaluation reliability during the search process \cite{yu2024improving}.

Collaborative search can also be achieved through ensemble-style designs. The Mixture-of-Search-Agents (MOSA) framework aggregates independent explorations from multiple LLMs and iteratively refines them, mitigating the limitations of any single model \cite{yang2025multi, cao2025multi2, xiong2025quantagent, zhang2025postergen}. This idea extends to process-level ensembling, where MCTS operates over reasoning traces produced by different LLMs to assemble the most coherent and accurate chain of thought \cite{park2024ensembling}. In multimodal contexts, Collective MCTS (CoMCTS) enables models to cooperatively explore reasoning spaces across modalities, thereby enriching datasets with diverse, interpretable reasoning steps \cite{yao2024mulberry}.

\subsubsection{Reward Model Design and Optimization}
The effectiveness of a search algorithm fundamentally depends on the quality of its reward function. Current research emphasizes designing and training reward models that can precisely guide search toward desirable outcomes. A prominent shift has been from coarse, outcome-level rewards to fine-grained, step-level feedback. Process-Supervised Reward Models (PRMs) provide such detailed supervision, enhancing reasoning in domains like mathematics and code generation \cite{ma2023let, lightman2023let}. These models often build upon Reinforcement Learning from Human Feedback (RLHF), which aligns models using preference-based supervision \cite{ouyang2022training, stiennon2020learning}. Since step-level annotation is expensive, recent work employs MCTS to automatically generate large-scale supervision data efficiently \cite{luo2024improve}.

Beyond direct supervision, several self-improving frameworks draw inspiration from AlphaZero. Here, an LLM uses MCTS to generate reasoning trajectories, which are then used to iteratively fine-tune itself or an associated reward model \cite{guan2025rstar, tian2024toward, xu2024sra}. Other studies combine MCTS-guided data generation with preference optimization techniques such as Direct Preference Optimization (DPO) to improve policy alignment \cite{xie2024monte, putta2024agent}. Extensions also explore integrating value models that autonomously provide feedback signals during search, eliminating the need for explicit process annotations \cite{chen2024alphamath}. These iterative loops enable continuous self-improvement by leveraging both successful and failed trajectories \cite{feng2023alphazero, wang2024q, zhai2025enhancing, chen2024step}. Furthermore, value networks learned during policy optimization (e.g., PPO) can be repurposed at inference time to guide MCTS decoding, bridging the training–inference gap \cite{liu2023don}.

Alternative formulations avoid explicit reward modeling. Outcome Value Models (OVMs), for instance, learn value estimation solely from final outcomes yet effectively evaluate partial reasoning paths \cite{yu2023ovm}. Other methods employ external tools as implicit reward sources—logical verifiers for program synthesis \cite{brandfonbrener2024vermcts}, knowledge retrievers for factual grounding \cite{choi2023kcts, wu2023knot, jiang2024rag, cheng2025think}—or repurpose the LLM itself as a world model or zero-shot reward estimator \cite{zhao2023large, hao2023reasoning, dainese2024generating, yu2023prompt}. Analytical studies reveal several counterintuitive findings: the overall accuracy of planning heavily depends on discriminator quality \cite{chen2024tree}, and step-level reward models exhibit superior sensitivity to mathematical logic compared to natural language coherence \cite{ma2025step}.

\subsubsection{Search Efficiency and Dynamics}
The principal challenge in tree-based reasoning lies in its computational expense. Recent work aims to enhance the efficiency, adaptivity, and scalability of MCTS while preserving exploration depth and reasoning quality.

One line of research focuses on structural and algorithmic improvements. Techniques such as LiteSearch employ dynamic node selection and adaptive exploration budgets guided by learned value networks to minimize redundant computation \cite{wang2024litesearch}. Although not strictly MCTS, related paradigms like Tree-of-Thoughts (ToT) have underscored the value of structured reasoning over purely greedy generation \cite{yao2023tree, long2023large}. These ideas have been generalized into Graph-of-Thoughts (GoT), where reasoning paths form graph structures that support merging, feedback loops, and hierarchical composition \cite{besta2024graph}. Other algorithmic innovations, such as bilevel MCTS, achieve amortized $O(1)$ node-selection time, greatly improving efficiency in deep search trees \cite{asai2025bilevel}. Information Directed Tree Search (IDTS) introduces a Bayesian mechanism for quantifying information gain across feedback types, thereby prioritizing more informative expansions \cite{chandak2024information}.

A complementary research direction seeks to make search dynamics adaptive. Frameworks such as Reasoning via Planning (RAP) employ an LLM as a world model to simulate and anticipate outcomes, achieving a more balanced exploration–exploitation trade-off \cite{hao2023reasoning}. Adaptive Branching MCTS (AB-MCTS) adaptively decides whether to expand breadth or depth at each node, generalizing repeated sampling strategies \cite{inoue2025wider}. Some variants embed reflection mechanisms directly into the search process: RethinkMCTS incorporates execution feedback to iteratively correct reasoning errors \cite{li2024rethinkmcts}, while Reflection on Trees (RoT) distills prior search experiences from stronger LLMs into reusable heuristics for weaker ones \cite{hui2024rot}.

Finally, MCTS has been extended to broader and higher-level applications, including natural language planning for diverse code generation \cite{zhang2023planning, wang2024planning}, web-based decision-making \cite{koh2024tree}, conversational policy optimization \cite{li2024planning, du2024sapient, he2024planning}, and automated machine learning pipeline design \cite{chi2024sela}. In some cases, search operates at a meta-level—optimizing prompts \cite{wang2023promptagent} or abstract reasoning schemas \cite{wu2024beyond}. To facilitate standardized benchmarking and comparison across these diverse applications, unified toolkits such as LLM Reasoners have been developed \cite{hao2024llm}.

\subsection{MCTS for Direct Test-Time Enhancement}
\label{sec:inference_enhancement}

This category includes methods that use Monte Carlo Tree Search (MCTS) primarily to improve the quality of the LLM's output for a single, given prompt at inference time, without updating the model's weights. These approaches treat the generation of a solution as a sequential decision-making problem, where the MCTS algorithm explores a tree of possible reasoning steps or text segments to find an optimal path. The core idea is to leverage lookahead planning to overcome the greedy, left-to-right nature of standard autoregressive decoding, thereby enhancing the model's performance on tasks that require strategic thinking, exploration, or backtracking.

\subsubsection{General Reasoning \& Problem Solving}

This research direction seeks to develop domain-agnostic frameworks that augment the general reasoning capabilities of LLMs. Efforts focus on making MCTS-based inference more efficient, interpretable, and robust. A key principle is to represent text generation not as a linear sequence but as a structured search over a tree of possibilities, facilitating exploration, lookahead, and revision—an idea popularized by \textit{Tree-of-Thoughts} (ToT) \cite{yao2023tree, long2023large} and extended by \textit{Graph-of-Thoughts} (GoT) \cite{besta2024graph}.

Subsequent research has advanced both the efficiency and conceptual sophistication of this search paradigm. Some works design lightweight search algorithms and dynamic resource allocation strategies to reduce the computational burden inherent in tree expansion \cite{wang2024litesearch}, for instance by using adaptive branching to decide whether to explore wider or deeper \cite{inoue2025wider}. Others introduce novel MCTS components, such as interpretable reward models based on contrastive decoding, to improve both accuracy and computational efficiency \cite{gao2024interpretable}. A key insight is to repurpose LLMs as world models capable of simulating future states to guide planning, exemplified by the \textit{Reasoning via Planning} (RAP) framework \cite{hao2023reasoning} and extended to virtual web environments \cite{gao2025websynthesis}. Further analyses highlight that the success of MCTS-based reasoning often hinges on the fidelity of the reward model or discriminator used to evaluate intermediate steps \cite{chen2024tree, hao2024llm}, leading to the development of more robust hierarchical reward models \cite{wang2025towards} and frameworks that unify reinforcement learning (RL) and search by using the learned RL reward function as a dynamic process reward model (PRM) \cite{jin2025your}.

Inspired by classical AI paradigms, several frameworks have introduced reinforcement-style or self-play mechanisms. For example, \textit{TS-LLM} employs an AlphaZero-like structure, where a learned value function drives decoding \cite{feng2023alphazero}. The \textit{rStar} framework adopts mutual self-play between a reasoner and a discriminator to refine search quality \cite{qi2024mutual}. Meta-cognitive extensions such as reflection further enhance robustness, allowing models to learn from past trajectories to avoid repeated errors \cite{hui2024rot}. This principle has been formalized in frameworks like \textit{Agent-R}, which uses MCTS to construct training samples that recover correct trajectories from erroneous ones \cite{yuan2025agent}, and \textit{ASTRO}, which teaches models to reason like search algorithms by explicitly reflecting and backtracking \cite{kim2025astro}. Other self-improvement strategies leverage MCTS to retrospectively revise and shorten reasoning paths for more efficient distillation \cite{lu2025retro} or use sibling-guided augmentation to reintegrate valuable insights from non-optimal search branches \cite{ren2025sigma}.

MCTS has become a cornerstone of multi-agent systems, where it orchestrates collaboration and debate. Frameworks like \textit{MASTER} \cite{gan2025master}, \textit{HALO} \cite{hou2025halo}, and \textit{MoSA} \cite{yang2025multi} use MCTS to coordinate agent recruitment, task decomposition, and communication, leveraging the collective expertise of multiple LLMs to solve complex problems. This paradigm extends to specialized domains like software engineering, where agents competitively debate to localize faults \cite{li2025swe}. Efficiency in such systems is enhanced by using data influence scores to guide tree search, prioritizing data synthesis that most effectively improves model training \cite{shi2025efficient}.

The versatility of MCTS is evident in its application across a growing array of domains. In Retrieval-Augmented Generation (RAG), it guides a dynamic interplay between reasoning and information retrieval, activating the model's intrinsic reasoning capabilities \cite{feng2025airrag} and ensuring factual accuracy \cite{hu2025mcts}, even enabling retriever-free models that traverse a corpus directly \cite{kim2025freeson}. In knowledge base question answering (KBQA), MCTS navigates large knowledge graphs to generate logical forms or structured queries \cite{luo2025kbqa, xiong2025mcts, wang2025dynamically}. The paradigm has been extended to multimodal reasoning, enhancing performance on tasks combining vision and language \cite{wu2025boosting, wang2025sota, liu2025mmc, yang2025re} and even for specific applications like multimodal misinformation detection \cite{cui2025t}. Further applications include ensuring safety alignment by guiding introspective reasoning \cite{zhang2025stair}, mitigating hallucinations \cite{duan2025prompt}, and tackling open-ended scientific discovery \cite{garikaparthi2025iris, agarwal2025open} and specialized tasks like heuristic design \cite{zheng2025monte, wang2025planning}, automated machine learning \cite{liang2025mcts}, and text-based game playing \cite{shi2025monte}.

\subsubsection{Mathematical Reasoning}
Mathematics serves as an ideal testbed for MCTS because its problems have unambiguous, verifiable solutions, facilitating the design of precise reward functions. This property enables fine-grained evaluation of both intermediate reasoning steps and final outcomes. Research in this area primarily targets improving reasoning quality and search efficiency. For instance, \textit{MCT Self-Refine} (MCTSr) incorporates a self-correction mechanism within the MCTS loop, enabling the LLM to iteratively refine its reasoning during exploration \cite{zhang2024accessing}. This concept of learning from errors is further explored in frameworks like \textit{LEMMA}, which explicitly constructs training data connecting incorrect solutions to correct ones to improve the model's reflective capabilities \cite{pan2025lemma}.

Efficiency and scalability remain core challenges. To mitigate computational costs associated with long reasoning chains, \textit{Markov Chain of Thought} (MCoT) compresses previous steps into concise state representations \cite{yang2024markov}. Others focus on refining the search process itself. For example, \textit{Constrained MCTS} (CMCTS) restricts the action space to semantically plausible steps, enhancing efficiency and rationality \cite{lin2025leveraging}, while lightweight energy-based path verifiers guide tree exploration without additional fine-tuning \cite{xu2023no}. Retrieval augmentation has also been integrated, with hierarchical systems retrieving abstract templates and analogous solution steps to guide the MCTS process \cite{dou2025enhancing}.

Several studies reduce supervision overhead by training value models solely on final outcomes \cite{yu2023ovm} or by using MCTS to autonomously collect high-quality process supervision data \cite{luo2024improve, lu2024mathcoder2}. Frameworks such as \textit{rStar-Math} \cite{guan2025rstar} and \textit{AlphaMath} \cite{chen2024alphamath} adopt self-evolutionary approaches in which policy and reward models are iteratively refined through MCTS-guided data synthesis without human intervention. In automated theorem proving, MCTS is used to guide proof assistants like Lean \cite{yang2023leandojo} and to solve complex problems by generating and pursuing subgoals within a structured search \cite{zimmer2025bourbaki}. However, some research challenges the necessity of complex search, demonstrating that simpler methods like Best-First Search can achieve competitive performance when properly scaled \cite{xin2025bfs}. The principles of process supervision have also been extended to multimodal mathematical reasoning, enhancing the logical robustness of vision-language models \cite{du2025mm}.

\subsubsection{Code Generation \& Software Engineering}
In software engineering, MCTS is employed to navigate the immense and combinatorial search space of possible code implementations. A key advantage in this domain is the availability of immediate, objective feedback from tools such as compilers, unit tests, and formal verifiers, which provide powerful and interpretable reward signals. Numerous studies leverage this feedback to guide the search toward correct, efficient, and verifiable code. For instance, \textit{RethinkMCTS} explores the reasoning process underlying code generation and uses execution feedback to iteratively refine erroneous reasoning paths \cite{li2024rethinkmcts}. Similarly, \textit{Planning-Guided Transformer Decoding} (PG-TD) integrates MCTS into token-level decoding, using test cases as reward signals to improve correctness \cite{zhang2023planning}. Frameworks like \textit{Adaptive Branching MCTS} have also shown that dynamically deciding whether to explore new code candidates ("wider") or refine existing ones ("deeper") based on feedback can outperform standard MCTS \cite{inoue2025wider}.

MCTS has also been adopted for structured query generation and formal program synthesis. In Text-to-SQL, frameworks like \textit{SQL-o1} \cite{lyu2025sql} and \textit{Alpha-SQL} \cite{li2025alpha} use MCTS for multi-step exploration with self-reward mechanisms. For formal synthesis, \textit{VerMCTS} integrates logical verification at every node in the search tree, ensuring partial correctness and delivering strong guarantees of soundness \cite{brandfonbrener2024verified}. The versatility of MCTS extends beyond program synthesis to hardware design, where it optimizes register-transfer-level (RTL) code for power, performance, and area trade-offs \cite{delorenzo2024make}.

At a larger scale, MCTS supports repository-level software engineering. Multi-agent frameworks such as \textit{SWE-Search} \cite{antoniades2024swe} and \textit{SWE-Debate} \cite{li2025swe} coordinate self-improvement and patch generation through deliberative search and reasoning. MCTS has been applied to automated program repair (\textit{APR-MCTS}) \cite{hu2025aprmcts} and even to code evaluation, where \textit{MCTS-Judge} explores reasoning trajectories to assess code correctness in an LLM-as-a-Judge setting \cite{wang2025mcts}. These approaches are often used not only for inference but also to generate high-quality fine-tuning data. For instance, \textit{SE-Agent} uses a self-evolutionary mechanism to optimize interaction trajectories for solving GitHub issues \cite{lin2025se}, and other work uses a refined MCTS process to generate high-quality Chain-of-Thought data for fine-tuning models on issue resolution \cite{wang2025mcts}. Collectively, these methods demonstrate MCTS’s potential to emulate human-like, deliberative reasoning in software engineering, bridging the gap between symbolic search and data-driven learning.

\subsubsection{LLM Agents \& Interactive Environments}
MCTS naturally complements large language model (LLM) agents operating in interactive environments by enabling strategic planning, exploration of action sequences, and adaptive decision-making based on environmental feedback. The \textit{Language Agent Tree Search} (LATS) framework exemplifies this paradigm, unifying reasoning, acting, and planning within a reflective MCTS structure to enhance deliberative control \cite{zhou2023language}. This approach has proven effective across a wide array of agentic tasks, including web navigation, complex reasoning, and software engineering.

In web navigation and GUI interaction, where agents must execute multi-step tasks in dynamic settings, tree search provides a robust mechanism for managing vast action spaces and uncertain state transitions. Methods such as \textit{WebPilot} employ dual-level optimization strategies—combining high-level planning with MCTS-guided subtask execution—to achieve strong performance \cite{zhang2025webpilot, koh2024tree}. This is further extended by frameworks that learn world models to simulate web environments, allowing MCTS to perform efficient, reversible planning for large-scale trajectory synthesis \cite{gao2025websynthesis}. Similarly, agents can be augmented with hierarchical skill libraries, which MCTS can leverage to prune the action space during online exploration \cite{xie2025mirage}. The paradigm is also effective in classic interactive environments like text-based games, where LLM-guided MCTS with dynamic memory enhances planning and learning \cite{shi2025monte}.

MCTS is also a cornerstone for automated problem-solving in specialized domains. In software engineering, it guides agents in resolving complex GitHub issues and performing automated program repair by globally evaluating explored patches and identifying promising refinement paths \cite{li2025swe, wang2025mcts, hu2025aprmcts}. For tasks requiring formal logic, such as automated theorem proving, MCTS helps navigate large proof spaces \cite{xin2025bfs, zimmer2025bourbaki}. This principle extends to scientific and heuristic discovery, where agents use MCTS to search for optimal AutoML pipelines \cite{chi2024sela, liang2025mcts}, design heuristics for optimization problems \cite{zheng2025monte, wang2025planning}, or even generate novel hypotheses for scientific research \cite{agarwal2025open, garikaparthi2025iris}.

A significant line of research focuses on enhancing the agent's core reasoning and self-improvement capabilities. Iterative self-training frameworks like \textit{Agent-R} use MCTS to construct training samples that teach agents to recover from errors \cite{yuan2025agent}, while others like \textit{AgentQ} combine MCTS with self-critique and offline fine-tuning \cite{putta2024agent}. The search process itself can be improved through introspective reflection and multi-agent debate to refine state evaluation \cite{yu2024improving, gan2025master, yang2025multi}. Furthermore, MCTS is pivotal for generating high-quality training data to instill sophisticated reasoning abilities. Frameworks like \textit{Retro-Search} and \textit{ASTRO} use MCTS-like algorithms to distill optimal reasoning paths from stronger models or teach search-like behaviors, which are then used to fine-tune smaller models \cite{lu2025retro, kim2025astro}. This self-evolution paradigm, where MCTS explores reasoning trajectories that are then used to train policy and reward models, has proven effective in domains ranging from medicine to general problem-solving \cite{jiang2025meds, lin2025se}.

\subsubsection{Retrieval-Augmented Generation (RAG) \& Knowledge-Intensive Tasks}
Ensuring factual accuracy and adherence to constraints such as safety, style, and alignment remains a central challenge for LLMs. MCTS provides a principled mechanism to steer generation toward desired attributes during inference by integrating external knowledge sources and reward models. Early work demonstrated that discriminator-guided MCTS can effectively control textual style and toxicity \cite{chaffin2021ppl}. This has been extended by safety-aware frameworks like \textit{STAIR}, which use a specialized MCTS to find a balance between helpfulness and safety \cite{zhang2025stair}, and by systems like \textit{SEM-CTRL}, which enforce rich syntactic and semantic constraints during decoding \cite{albinhassan2025texttt}.

In knowledge-intensive settings, MCTS helps mitigate hallucination and enhance factual grounding. Frameworks such as \textit{KNOT-MCTS}, \textit{KCTS}, and prompt-based MCTS variants integrate external retrieval or dynamic strategies into the search process to align content with verified sources \cite{wu2023knot, choi2023kcts, duan2025prompt}. \textit{HaluSearch} conceptualizes generation as a deliberative "slow thinking" process, using a self-evaluation reward model to identify reliable reasoning pathways \cite{cheng2025think}. This structured exploration is critical for complex tasks like Knowledge Base Question Answering (KBQA) and Text-to-SQL, where MCTS-driven agents can explore the knowledge base or schema to construct accurate queries \cite{luo2025kbqa, xiong2025mcts, wang2025dynamically, lyu2025sql, li2025alpha}.

The integration of MCTS into Retrieval-Augmented Generation (RAG) represents a significant advancement. Instead of static retrieval, MCTS enables dynamic, interleaved retrieval and reasoning. Systems like \textit{RAG-Star}, \textit{RARE}, and \textit{AirRAG} incorporate retrieval and verification within the search loop, using evidence consistency or strategic planning as a reward signal \cite{jiang2024rag, tran2024rare, feng2025airrag}. Similarly, \textit{Think\&Cite} treats attributed generation as a search problem, rewarding factually supported progress \cite{li2024think}, while \textit{CORAG} optimizes retrieval composition \cite{wang2024corag}. This paradigm transforms retrieval into an adaptive reasoning process, as demonstrated by frameworks like \textit{SearChain}, \textit{MCTS-RAG}, and \textit{R2-LLMs}, which decide when and what to retrieve at each step, incrementally building verifiable solutions \cite{xu2024search, jiang2024rag, hu2025mcts, dou2025enhancing}. This tight coupling between search and knowledge retrieval is further advanced by innovations like \textit{FREESON}, which allows LLMs to autonomously traverse a corpus via MCTS, learning to retrieve without an explicit retriever \cite{kim2025freeson}. These methods significantly augment the deliberative capacity of LLMs, enabling smaller models to perform competitively on complex knowledge-intensive tasks \cite{hu2025mcts, dou2025enhancing}.

\subsubsection{Multimodal Reasoning}
The tree search paradigm has recently been extended to multimodal large language models (MLLMs), enhancing their ability to perform complex reasoning across text, images, and video. By structuring the reasoning process, MCTS decomposes a multimodal problem into a sequence of interpretable steps, where each step can involve grounding textual concepts in visual evidence or actively seeking new information.

A key application is active retrieval, where models dynamically acquire supporting evidence. The \textit{AR-MCTS} framework exemplifies this by integrating retrieval of both textual and visual information within the MCTS loop, ensuring that generated explanations are well-supported by facts from a hybrid-modal corpus \cite{dong2024progressive}. This idea of exploring and re-ranking multimodal contexts is further refined by methods that use tree search to select the most relevant examples for in-context learning, improving response consistency and quality \cite{yang2025re}. Other approaches, such as \textit{DyFo}, use MCTS to simulate human-like visual search, enabling models to dynamically focus on key image regions and reduce hallucinations without additional training \cite{li2025dyfo}.

MCTS is also instrumental in generating high-quality training data for multimodal self-improvement. For instance, MCTS can be repurposed to measure sample difficulty, allowing for the selection of a highly effective, condensed dataset for reinforcement fine-tuning, as demonstrated by \textit{ThinkLite-VL} \cite{wang2025sota}. Similarly, frameworks like \textit{MMC} and \textit{MM-PRM} use MCTS to automatically construct large-scale critique datasets and step-level annotations for training powerful multimodal reward models, which in turn guide the model's reasoning at inference time \cite{liu2025mmc, du2025mm}.

Beyond reasoning, MCTS powers creative and specialized multimodal tasks. In automated storytelling, frameworks like \textit{AniMaker} use a multi-agent system where an MCTS-driven strategy intelligently generates and selects high-quality video clips to form a coherent narrative \cite{shi2025animaker}. MCTS is also used to create comprehensive benchmarks for tasks like video captioning by iteratively generating diverse and detailed descriptive sentences \cite{yu2025evaluating}. Approaches like \textit{AStar} abstract high-level reasoning patterns, or "thought cards," using MCTS and retrieve them at inference time to scaffold solutions for new problems \cite{wu2025boosting}. These methods substantially enhance the robustness and compositional reasoning capabilities of vision-language models \cite{yang2025re}.

\subsubsection{Specialized and Novel Applications}

Beyond general reasoning and language understanding, \textit{MCTS-guided inference} has been increasingly adopted in specialized domains and for novel applications.

\textbf{Game Playing and Strategic Reasoning:} While MCTS is traditionally renowned for its success in board games such as \textit{Go}, its integration with LLMs has opened new frontiers. LLMs can function as \textit{action pruners} and \textit{value function approximators} within an MCTS loop for deterministic games like chess, achieving effective play without additional training \cite{guo2024can}. For games with partial observability, the \textit{STRATEGIST} framework employs a bi-level search process in which an LLM explores high-level textual strategies, while a low-level MCTS refines and executes them \cite{light2024strategist}.

\textbf{Scientific Discovery:} MCTS has also been applied to accelerate scientific reasoning in large combinatorial spaces. Notably, it has been used to discover novel catalysts by balancing multiple chemical property trade-offs \cite{sprueill2023monte} and to guide \textit{de novo} therapeutic peptide generation by jointly optimizing for multiple biological objectives \cite{tang2025peptune}.

\textbf{Prompt and Hyperparameter Optimization:} At a meta-reasoning level, MCTS can be applied not to solve downstream tasks directly, but to optimize the \textit{instructions} or \textit{prompts} that guide an LLM. The \textit{PromptAgent} framework formulates prompt engineering as a strategic planning problem, using MCTS to explore and refine the vast space of possible instructions based on model feedback \cite{wang2023promptagent}. Similar search-based strategies have also been applied for \textit{tuning-free self-alignment}, allowing models to autonomously generate optimal alignment instructions without any gradient updates \cite{singla2024dynamic}.

\textbf{Reasoning over Structured Data:} For tasks that involve reasoning over knowledge graphs, \textit{Relational MCTS} automatically extracts relevant relational paths from textual knowledge graphs, supplying structured evidence to the LLM and enhancing its ability to answer complex, compositional queries \cite{huang2024ritek}.

\subsection{MCTS for Self-Improvement via Data Generation}
\label{sec:self_improvement}

This paradigm leverages MCTS not merely to identify a single optimal solution but to generate high-quality reasoning trajectories. These trajectories serve as synthetic data for fine-tuning large language models (LLMs) or reward models, forming a virtuous cycle of iterative self-improvement.

\subsubsection{Foundational Self-Improvement Frameworks}
Recent works establish the foundations for employing MCTS within self-training loops, drawing inspiration from reinforcement learning frameworks such as AlphaZero and preference optimization. The central idea is to construct a self-evolutionary cycle where both a policy model (the LLM) and a value or reward model are iteratively enhanced. For example, frameworks including rStar-Math, AlphaLLM, TS-LLM, and ReST-MCTS* utilize MCTS to perform extensive rollouts that yield large volumes of verified, step-by-step reasoning data. This data is subsequently used to train both the LLM and an auxiliary process preference model \cite{guan2025rstar, tian2024toward, wan2024alphazerolike, zhang2024restmcts}. In an AlphaZero-like manner, the model learns from its own structured explorations and adapts across diverse tasks and model scales. A learned value function guides the search more effectively than reliance on the pretrained LLM’s intrinsic priors alone \cite{feng2023alphazero}. Some approaches explicitly unify these paradigms, demonstrating that a reward function learned via reinforcement learning can serve as the ideal process reward model (PRM) for guiding search, eliminating the need for labeled process data \cite{jin2025your}. Others refine the learning process with novel policy optimization techniques, such as using segment-level advantage estimation to balance credit assignment and estimation accuracy \cite{yang2025treerpo}.

The data generated from MCTS rollouts is typically converted into preference pairs—comparisons between superior and inferior reasoning steps—and employed with algorithms such as Direct Preference Optimization (DPO) to refine the model’s policy \cite{xie2024monte}. This self-contained process, exemplified by AlphaMath, autonomously generates process supervision and step-level evaluation signals without requiring human or teacher-model annotations \cite{chen2024alphamath}. To improve data quality and diversity, some methods augment MCTS data with hierarchical node compression \cite{wang2025towards} or by reintegrating discarded sibling nodes from the search tree to provide comparative feedback for refining optimal trajectories \cite{ren2025sigma}. Other frameworks like ASTRO and Agent-R focus on learning from both successful and failed trajectories to improve generalization and develop robust recovery mechanisms \cite{kim2025astro, yuan2025agent}. This includes weak-to-strong generalization, where trajectories from a weak model, including failures, are organized into a tree to efficiently elicit a strong model's optimal policy \cite{ye2025weak}. Data generation can be further targeted by using metrics like structural entropy to guide MCTS toward a model's knowledge gaps \cite{wei2025structural} or by using influence scores to select the most impactful data for model training \cite{shi2025efficient}.

The self-improvement principle also underpins distillation, where MCTS is used to generate or refine reasoning paths from a large teacher model to create higher-quality training data for a smaller student model. For example, Retro-Search uses an MCTS-like algorithm to retrospectively revise reasoning paths to discover shorter, more efficient traces \cite{lu2025retro}, while other work constructs tree-based Chain-of-Thought data from scratch to widen the distillation bottleneck \cite{yin2025towards}. Beyond data generation, novel search frameworks enhance the reasoning process itself. For instance, the Chain-of-Associated-Thoughts (CoAT) framework integrates MCTS with a dynamic associative memory to expand the search space \cite{pan2025coat}, while others use self-evaluation to guide tree search without external reward models \cite{wu2025selt} or adaptively decide whether to broaden or deepen the search tree based on external feedback \cite{inoue2025wider}. At a meta-level, frameworks like AgentSwift employ hierarchical MCTS to search for optimal agentic system designs, including workflows and functional components \cite{li2025agentswift}.

\subsubsection{General Capabilities \& Alignment}
MCTS also serves as a powerful mechanism for generating synthetic data that enhances LLM capabilities and improves alignment with human values. In prompt optimization, for example, frameworks such as PromptAgent formulate prompt engineering as a strategic planning task, using MCTS to explore diverse instruction spaces and learn from failures to produce expert-level prompts \cite{wang2023promptagent, yu2025optimizing}. Similarly, Dynamic Rewarding with Prompt Optimization (DRPO) applies MCTS-based search for tuning-free self-alignment, generating optimal alignment instructions at inference time without costly retraining \cite{singla2024dynamic}.

In safety and factuality contexts, MCTS can synthesize fine-grained reasoning data to train models in identifying and mitigating risk. Frameworks such as HaluSearch, KNOT-MCTS, and others employ self-evaluation-guided search to minimize hallucinations by favoring more reliable reasoning pathways \cite{cheng2025think, wu2023knot, duan2025prompt}. The STAIR framework specifically integrates safety-informed MCTS (SI-MCTS) to generate step-level reasoning data that balances helpfulness and safety, which is then used to train a process reward model for improved test-time responses \cite{zhang2025stair}. This principle extends to red-teaming, where MCTS-based fuzzing frameworks like AgentXploit or adaptive stress testing systematically search for prompt perturbations and contextual manipulations to automatically discover and exploit agent vulnerabilities \cite{wang2025agentxploit, chakraborty2025adaptive}. Some approaches use MCTS to enforce syntactic and semantic constraints during decoding \cite{albinhassan2025texttt} or guide search using value models inherited from previous alignment stages like PPO \cite{liu2023don, khanov2024args}. Additionally, multi-agent variants of MCTS coordinate distributed sampling processes to synthesize higher-quality alignment data through collective exploration and collaborative problem-solving \cite{ye2024multi, gan2025master, yang2025multi, hou2025halo}.

\subsubsection{Scientific \& Specialized Domains}
The self-improvement paradigm has been increasingly adopted across a diverse range of specialized domains. In software engineering and code generation, frameworks such as o1-Coder and SRA-MCTS explicitly employ MCTS to generate extensive code samples with intermediate reasoning traces, which are subsequently used to iteratively fine-tune the model’s programming proficiency \cite{zhang2024o1, xu2024sra}. This approach has been extended to automated program repair \cite{hu2025aprmcts} and repository-level issue resolution, where multi-agent debate frameworks or MCTS-refined Chain-of-Thought data are used to generate high-quality patches and fine-tuning data \cite{li2025swe, wang2025mcts, antoniades2024swe}. Other systems use MCTS to improve test-time evaluation of code correctness \cite{wang2025mcts}. In knowledge-intensive tasks, MCTS enhances retrieval-augmented generation (RAG) by planning reasoning actions \cite{feng2025airrag, hu2025mcts} and enables agents to traverse a corpus without a separate retriever model \cite{kim2025freeson}. For structured data, MCTS is used to generate high-quality annotations for knowledge base question answering (KBQA) \cite{luo2025kbqa, xiong2025mcts, wang2025dynamically} and to build experience memory for text-to-SQL tasks \cite{lyu2025sql, li2025alpha, gu2025toward}.

At a meta-level, this paradigm underpins agent-based systems in Automated Machine Learning (AutoML), where frameworks such as SELA, I-MCTS, and KompeteAI utilize MCTS to optimize the discovery and configuration of machine learning pipelines \cite{chi2024sela, liang2025mcts, kulibaba2025kompeteai}. MCTS has also been applied to optimize hyperparameters for fine-tuning models in tasks like named entity matching \cite{volkova2024novel}. Beyond these, MCTS facilitates the generation of high-quality synthetic tabular data \cite{locowic2024synthetic} and the development of domain-specific models through self-evolution, including clinical reasoning systems in medicine \cite{jiang2025meds, ding2025promed}, legal reasoning systems \cite{liu2025towards}, and psychological counseling dialogues \cite{lu2025mctsr}. In chemistry and materials science, search-driven agentic frameworks like ChemAgent have been employed to plan and execute intricate synthesis procedures and integrate specialized tools \cite{bran2023chemcrow, boiko2023emergent, wu2025chemagent}. In strategic settings such as game-playing, MCTS continues to guide the learning of high-level strategies via self-play simulations in traditional \cite{guo2024can, light2024strategist} and text-based games \cite{shi2025monte}. This principle extends to molecular structure elucidation, where K-MSE~\cite{zhuang-etal-2025-boosting} integrates MCTS with a knowledge base to markedly improve chemical reasoning. Other emerging applications include automated scientific discovery \cite{garikaparthi2025iris, wang2025automated, agarwal2025open}, web and GUI agent training \cite{gao2025websynthesis, xie2025mirage}, synthetic data generation for multilingual translation \cite{zou2025trans, feng2025mt}, personalized educational content creation \cite{wu2025personalized}, heuristic discovery for optimization problems \cite{zheng2025monte, zhang2023bridging, wang2025planning}, and automated theorem proving \cite{zimmer2025bourbaki}.

\subsubsection{Multimodal Applications}
The self-improvement paradigm has also expanded into multimodal domains, where MCTS enhances the reasoning capabilities of Vision-Language Models (VLMs). To address the scarcity of fine-grained supervision, MCTS-based pipelines can autonomously generate millions of step-level annotations for training process reward models (PRMs) without human labeling \cite{du2025mm}. For example, MulBerry adopts a Collective MCTS (CoMCTS) strategy that coordinates multiple models to collaboratively search for effective reasoning trajectories, subsequently constructing large-scale multimodal datasets for training advanced Multimodal Large Language Models (MLLMs) \cite{yao2024mulberry}. Other approaches use MCTS to automate structured thinking by retrieving optimal high-level reasoning patterns \cite{wu2025boosting} or re-rank retrieved reasoning context to improve consistency \cite{yang2025re}.

Another approach, AR-MCTS, integrates active retrieval mechanisms into the search process to generate step-wise annotations, which are used to train progressive reward models that verify reasoning chains \cite{dong2024progressive}. This results in a self-improving feedback loop, enabling VLMs to ground their reasoning more effectively in both visual and textual modalities. Recent extensions introduce multimodal actor–critic frameworks, where MCTS directs an actor model to explore diverse reasoning paths, while an annotator model compares successful and failed trajectories to produce critique data that guide self-correction \cite{liu2025mmc}. Data-efficient variants leverage MCTS to quantify the difficulty of visual reasoning tasks by estimating the number of search iterations required for solution discovery, thereby selecting a compact yet highly informative subset of challenging samples for reinforcement fine-tuning \cite{wang2025sota}. The paradigm has also been applied to generate data for vision-language-action (VLA) policies \cite{neary2025improving}, create animated stories \cite{shi2025animaker}, generate fine-grained video captions \cite{yu2025evaluating}, and detect multimodal misinformation \cite{cui2025t}. These principles are even being extended to other generative modalities, such as using MCTS to enhance diffusion models for solving complex reasoning tasks like mazes \cite{zhang2025t} or for steerable scene generation \cite{pfaff2025steerable}.

\section{Informed Search Based Method}
\label{sec:Informed}

To enhance the reasoning capabilities of Large Language Models beyond simple sequential generation, researchers have increasingly turned to informed search algorithms. This paradigm structures problem-solving as a tree traversal, where heuristic guidance helps navigate vast and complex solution spaces efficiently. Early frameworks such as Tree-of-Thoughts (ToT) adapted classical algorithms like Breadth-First Search (BFS) and Depth-First Search (DFS), using the LLM itself to evaluate intermediate 'thoughts' and prioritize promising reasoning paths. Building on this, more recent approaches have implemented A* search, a more sophisticated heuristic method, to further optimize exploration. Methods like LLM-A*, ToolChain*, and Q* exemplify this trend by designing intricate cost and heuristic functions that incorporate memory, self-consistency, and learned value estimates to guide the search for optimal solutions. This section explores these key informed search strategies, detailing how they formalize and direct the LLM's reasoning process.

\subsection{Informed BFS/DFS}
The Tree-of-Thoughts (ToT) framework \citep{10.5555/3666122.3666639} enables Large Language Models (LMs) to systematically explore multiple reasoning paths by formulating problem-solving as a tree search. Each node represents a state $s=[x, z_{1\dots i}]$, where $x$ is the input and $z_{1\dots i}$ is the sequence of generated thoughts. ToT comprises four key components: problem structuring, thought generation, state evaluation, and a search strategy.

In this framework, the problem is first \textbf{decomposed} into intermediate reasoning steps. At each step $i+1$, a generator $G(p_\theta,s,k)$ produces $k$ candidate thoughts from the current state using the LM $p_\theta$. Thought generation can occur via (1) \textbf{sampling} i.i.d.\ thoughts from a Chain-of-Thought (CoT) prompt—effective for broad search spaces—or (2) \textbf{proposing} thoughts sequentially through a “propose prompt” to reduce redundancy in more constrained tasks. An evaluation function $V(p_\theta,S)$ then assesses progress across candidate states $S$, either through a \textbf{value-based} mode, assigning scores to each state, or a \textbf{voting-based} mode, selecting the most promising candidate via LM judgment.

\textbf{Informed search strategies} operationalize these components through two primary algorithms. The \textbf{informed Breadth-First Search (BFS)} algorithm functions like a beam search, maintaining a beam of $b$ states per level to control exponential growth, making it suitable for problems of bounded depth $T$. Conversely, the \textbf{informed Depth-First Search (DFS)} algorithm follows a single reasoning path until its value score drops below a threshold, at which point the path is pruned.

BFS-based approaches have proven particularly versatile. Frameworks such as \textbf{Beam-LLM} \citep{xie2023selfevaluation} and \textbf{PathFinder} \citep{golovneva2023pathfinder} extend ToT’s beam search, respectively using LLM-based value functions and similarity metrics to select top candidates. Beyond text reasoning, this paradigm underlies methods like \textbf{Think-on-Graph} \citep{sun2024thinkongraph,ma2025thinkongraph}, which navigate knowledge graphs through SPARQL-based expansion and LLM-guided pruning. In causal discovery, BFS-style search has been adapted to reduce query complexity from quadratic to linear \citep{jiralerspong2024efficient}, integrate observational data into prompts \citep{susanti2025can}, and incorporate dynamic, LLM-informed scoring for identifying fairness-critical bias paths \citep{zanna2025uncovering}.

Recent work pushes these ideas toward greater autonomy and adaptivity. The \textbf{Autonomous Tree-Search (ATS)} paradigm internalizes BFS-like exploration within the LLM itself using a fixed system prompt, cutting API costs compared to externally controlled search loops \citep{zhang2023autonomous}. Similarly, \textbf{LLM-First Search (LFS)} allows the model to dynamically adjust its exploration depth and width, achieving a more flexible and efficient search process \citep{herr2025llm}. At the architectural level, the \textbf{Coconut} (Chain of Continuous Thought) framework reimagines reasoning in a continuous latent space, where a single “continuous thought” vector implicitly represents multiple parallel reasoning paths, effectively enabling BFS-like exploration in vector space \citep{hao2024training}.

Finally, a related but distinct family of approaches employs \textbf{Best-First Search}, which greedily expands the single most promising node from an open set. \textbf{Best-LLM} \citep{koh2024tree} demonstrates this for web navigation, maintaining a priority queue of states scored by an LLM-based heuristic. In formal reasoning, \textbf{BFS-Prover} \citep{xin2025bfs} combines Best-First Search with expert iteration and Direct Preference Optimization (DPO), achieving state-of-the-art theorem-proving results by normalizing proof lengths and encouraging deeper exploration. Together, these methods highlight how informed search—whether breadth-first, depth-first, or best-first—forms a principled foundation for scalable, LLM-guided reasoning.

\subsection{A*}
To mitigate the computational overhead associated with methods like Monte Carlo Tree Search (MCTS), recent work has explored A*-based search algorithms. These methods guide exploration using a specialized cost function $f(n) = g(n) + h(n)$, which prioritizes nodes that appear to be on the most promising path to a solution. This function balances the cost of the path taken so far, $g(n)$, with an estimated cost to reach the goal, $h(n)$. 

A prominent application of A* in robotics and path-finding is \textbf{LLM-A*} \citep{meng2024llm}, which synergistically combines the precise pathfinding capabilities of classic search algorithms with the global, commonsense reasoning of LLMs. This framework enhances traditional A* search for tasks like maze-solving by incorporating LLM-based heuristics. While the cost-so-far, $g(n)$, remains the standard path distance, the heuristic estimate, $h(n)$, is innovatively formulated. It combines the geometric distance from a candidate node to the goal with the distance to a waypoint suggested by an LLM. This hybrid heuristic grounds the search in physical reality while intelligently biasing it with the LLM's high-level understanding of the problem space. Furthermore, the framework supports a human-in-the-loop approach, where human feedback on intermediate results can refine the planning process, making it transparent and interactive.

Beyond path-finding, A* has been adapted to navigate the vast action spaces of LLM-based agents. \textbf{ToolChain*} \citep{zhuang2023toolchain} addresses the challenge of selecting a correct sequence of tools (e.g., API calls) by formulating the problem as a tree search. In this formulation, each node represents a possible API function call, and the A* algorithm efficiently explores this decision tree. By incorporating a task-specific cost function, ToolChain* prunes high-cost branches corresponding to incorrect or inefficient action sequences, allowing it to identify an optimal plan while effectively balancing exploration and exploitation.

Similarly, A* search can guide the internal reasoning process of an LLM. The \textbf{Q*} framework \citep{wang2024q} casts multi-step reasoning as a heuristic search problem to improve the model's decoding process. Instead of generating text purely auto-regressively, the model uses A* to explore different reasoning steps. Its key innovation is a learned, plug-and-play Q-value model that serves as the heuristic function $h(n)$. This function estimates the expected future reward of a reasoning path, guiding the LLM towards more promising intermediate steps and enhancing performance without requiring costly, task-specific fine-tuning of the base model. The primary innovation in these methods lies in constructing composite heuristics for $g(n)$ and $h(n)$ from diverse, LLM-relevant signals, as summarized in Table~\ref{tab:astar_heuristics}.

\begin{table}[!t]
\centering
\caption{Compact Overview of A* Heuristic Components for LLMs}
\label{tab:astar_heuristics}
% --- 調整間距的指令 ---
\renewcommand{\arraystretch}{0.8} % 增加行高 (預設為 1.0)
\setlength{\tabcolsep}{6pt}       % 增加欄寬間距 (預設為 6pt)
% ---------------------
\begin{tabular}{@{}l c p{7cm}@{}}
\toprule
\textbf{Heuristic} & \textbf{A* Component} & \textbf{Mechanism (and Signal Source)} \\
\midrule
\rowcolor{blockbg}
Process-Based Rewards   & $g(n)$ & Aggregates step-wise rewards from execution feedback (e.g., logits, rule checks). \\
\addlinespace % 這個指令在 booktabs 套件中，會增加美觀的額外空間
Statistical Consistency & $g(n)$ & Favors steps that are frequently proposed across multiple generation samples. \\
\addlinespace
\rowcolor{blockbg}
Memory-Based Comparison & $g(n), h(n)$ & Scores path similarity against a repository of high-quality examples (e.g., using LCS). \\
\addlinespace
Learned Future Value    & $h(n)$ & Estimates the cost-to-goal using a trained proxy model (e.g., a Q-function). \\
\bottomrule
\end{tabular}
\end{table}

\begin{center}
\textbf{ToolChain*}
\end{center}

In ToolChain*, the cost function for a node $n$ is the standard A* formulation, $f(n) = g(n) + h(n)$, where $g(n)$ is the \textbf{cumulative cost} from the start node to $n$, and $h(n)$ is a heuristic estimate of the \textbf{future cost} to the goal. The cumulative cost $g(n)$ is the sum of single-step costs over all ancestors of $n$, denoted $an(n)$. Each single-step cost is derived from two value functions, $g_{t,1}$ and $g_{t,2}$, whose outputs are bounded in $[0,1]$. The cost is formulated as the geometric mean of the complements of these values. The cumulative cost is thus:
\begin{equation}
\label{eq:toolchaingn}
g(n) = \sum_{i\in \{an(n),n\}}(1-g_{t,1}(i))^\alpha \cdot (1-g_{t,2}(i))^{1-\alpha},
\end{equation}
where the hyperparameter $\alpha$ weights the contribution of each value function.

The first value function, $g_{t,1}(n)$, is task-specific and draws from a \textbf{long-term memory} $\mathcal{M}$, which is initialized with seed demonstrations and augmented with successful plans discovered during search. Each memory entry $m_j$ is a plan sequence $(s_{j,0}, a_{j,1}, \dots, a_{j,T_j})$. This function evaluates the current plan $s_n$ by computing its maximum longest common subsequence (LCS) score against all plans in memory: $g_{t,1}(n)=\max_{m_j\in\mathcal{M}}\frac{\text{LCS}(s_n,m_j)}{\min(L(s_n),L(m_j))}$, where $L$ is the sequence length. The second value function, $g_{t,2}(n)$, is based on \textbf{self-consistency frequency}. It measures the frequency with which node $n$ is proposed as the next step across $k$ independently sampled reasoning paths, reflecting its reliability.

The \textbf{future cost} $h(n)$ is formulated analogously to $g(n)$:
\begin{equation}
\label{eq:toolchainhn}
h(n) = \sum_{i\in \{an(n),n\}}(1-h_{t,1}(i))^\beta \cdot (1-h_{t,2}(i))^{1-\beta},
\end{equation}
where $\beta$ is the geometric mean weight. The first heuristic, $h_{t,1}(n)$, leverages the \textbf{long-term memory} $\mathcal{M}$. For an action node $n$, it finds the action $a$ in each memory plan $m_j$ with the highest lexical similarity to $n$. The heuristic is the sum of these actions' relative positions: $h_{t,1}(n)=\sum_{m_j\in\mathcal{M}} \mathbf{1}_{{a\in m_j}}\frac{pos(a,m_j)}{T_j}$. The second heuristic, $h_{t,2}(n)$, is an \textbf{LLM imagination score}. An LLM generates a plausible future plan toward a target node $n_T$, and the heuristic value is the ratio of the current path length to the total imagined path length: $h_{t,2}(n)=\frac{|an(n)|}{|an(n_T)|}$, where $|an(\cdot)|$ is the number of ancestors. A higher score signifies closer proximity to the goal.

\begin{center}
\textbf{Q*}
\end{center}

In Q*, the cost function is $f(n) = g(n) + \lambda h(n)$, where $\lambda$ is a weighting hyperparameter. The accumulated cost $g(n)$ is an aggregation of process-based rewards for the current node and its ancestors: $g(n) = \text{Agg}(\{\mathcal{R}(s) \mid s \in an(n) \cup \{n\}\})$. The reward function $\mathcal{R}$ can be derived from human feedback, ground-truth labels, predefined rules, or LM logit scores. The aggregation function, $\text{Agg}$, can be chosen from $\{\max, \min, \sum, [-1]\}$, where $[-1]$ indicates selecting the reward of the last node.

The heuristic cost $h(n)$ is a Q-function that estimates the expected future reward. As an exhaustive search over subsequent steps is intractable, the heuristic is approximated by taking the maximum Q-value among the top-$k$ actions proposed by the LLM policy $\pi_\theta$: $h(n) = \max_{a_t \in \text{top-k}(\pi_\theta(\cdot|n))} Q(n, a_t)$. A primary challenge is estimating optimal Q-values when the frozen policy $\pi_\theta$ is suboptimal. The authors propose three methods for learning a proxy Q-value model: (1) offline reinforcement learning on curated data, (2) learning from MCTS rollouts, or (3) distillation from a stronger LLM. However, this approach may have limited generalization, and the anticipated computational savings are not guaranteed.

\section{Search In Prompt Space}
\label{sec:search_in_prompt_space}

While most current test-time scaling methods, including all we have covered above, conduct \emph{search in answer space}—directly exploring candidate reasoning paths or intermediate solutions—an emerging line of research shifts the optimization process to the \emph{prompt space}~\cite{zhang2025prompt}. Rather than modifying the generated content, these methods iteratively refine or search over prompts that steer the model's reasoning trajectory. This perspective treats the prompt itself as a control variable in a high-dimensional discrete space, where each candidate prompt induces a distinct reasoning policy over the same base model.

\begin{figure}[t]
    \centering
    \includegraphics[width=1\linewidth]{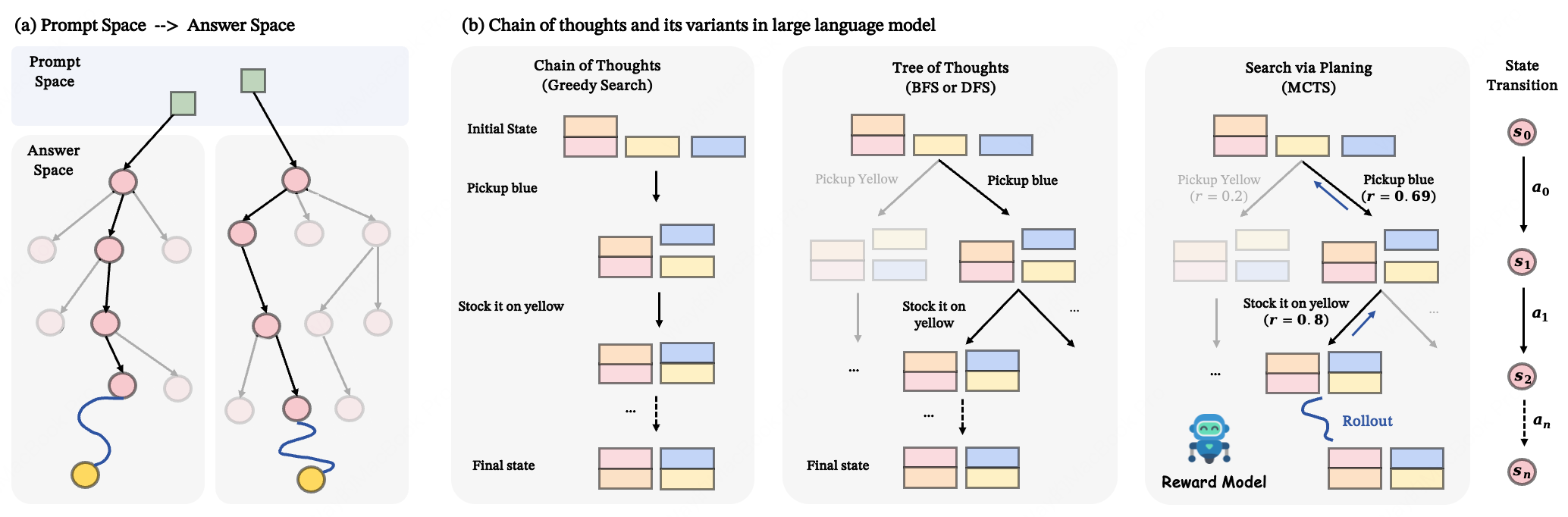}
    \caption{Search in prompt space v.s. search in answer space. Search allows for complicated reasoning beyond single path as seen in CoT.}
    \label{fig:promptSearch}
    % \vspace{-1.2em}
\end{figure}

Recent work such as \textbf{Automatic Prompt Optimization (APO)}~\cite{pryzant-etal-2023-automatic} focus on the search in prompt space~\ref{fig:promptSearch}. Given a task and a validation set, prompt search begins with an initial handcrafted prompt and employs large language models to propose, evaluate, and refine candidate prompts automatically. The framework decomposes the optimization loop into three stages: (i) \emph{prompt generation}, where variants are produced via LLM-driven mutation or rewriting; (ii) \emph{evaluation}, where each prompt’s downstream performance (e.g., accuracy or reward) is measured; and (iii) \emph{selection}, where high-performing candidates are retained or combined to guide the next iteration. This process embodies a meta-level search, optimizing the policy that governs how the model reasons rather than the reasoning steps themselves.

Conceptually, search in prompt space can be interpreted as \emph{meta-inference}—a higher-order optimization that adapts the inference procedure without altering model weights. Compared to answer-space search, which explores reasoning trajectories within a fixed prompting scheme, prompt-space search operates one level higher, learning to generate better reasoning algorithms through prompt adaptation. The optimization objective is therefore external to the reasoning process, focusing on eliciting the most effective \emph{reasoning strategy} rather than the most accurate \emph{solution}.

Such prompt-space optimization methods naturally complement search in answer space. Whereas the latter focuses on inference-time deliberation conditioned on a fixed query formulation, the former performs structural adaptation over the query formulation itself. As prompt optimization techniques mature—ranging from evolutionary strategies to reinforcement learning over discrete prompt tokens—they offer a scalable and model-agnostic route toward improved test-time performance, potentially serving as the outer loop to guide answer-space search policies.

\subsection{Search Complexity in Prompt and Answer Spaces}

Test-time reasoning with large language models can be understood as a search process unfolding across two coupled spaces: the \emph{prompt space}, which defines how reasoning is initiated and structured, and the \emph{answer space}, which represents the trajectories of reasoning or solutions produced under a given prompt. Each space possesses distinct search characteristics and computational bottlenecks, and their interaction fundamentally shapes the efficiency and scalability of inference-time optimization.

\vspace{-1.em}
\paragraph{Prompt Space Search.}
Search in the prompt space operates at a meta level, seeking to identify a high-level control program that guides the model's reasoning behavior (Table~\ref{tab:prompt_search_methods}).  This search exhibits high \emph{semantic branching complexity}: each candidate prompt can induce a qualitatively different reasoning distribution, leading to non-smooth objective landscapes. Because prompt evaluation requires full model inference for each candidate, the computational cost scales linearly with the number of candidates but exponentially with the diversity of the prompt modifications considered. The effective search depth is thus bounded by evaluation cost, prompting the use of heuristic or LLM-driven mutation operators to navigate the combinatorial space efficiently.

\vspace{-1.em}
\paragraph{Answer Space Search.}
Conversely, search in the answer space—exemplified by Monte Carlo Tree Search (MCTS), beam search, and self-refinement—unfolds within the reasoning trajectories conditioned on a fixed prompt. Its branching factor corresponds to the number of possible next-step continuations, while its depth corresponds to the number of reasoning steps or thought expansions. The complexity here arises from the exponential growth of possible paths as reasoning deepens, making it crucial to incorporate value estimates or pruning strategies to maintain tractability. In this sense, answer-space search optimizes the inference path given a static reasoning policy defined by the prompt.

\vspace{-1.em}
\paragraph{Coupled Search Dynamics.}
Prompt and answer spaces interact hierarchically: the prompt determines the policy that shapes exploration in the answer space, while the observed outcomes in the answer space provide feedback signals for prompt adaptation. This coupling forms a bi-level optimization problem:
\begin{equation}
    \min_{p \in \mathcal{P}} \; \mathbb{E}_{x \sim \mathcal{D}} \big[ \mathcal{L}(f_{\theta}(x; p)) \big],
\end{equation}
where $p \in \mathcal{P}$ represents a prompt sampled from the prompt space and $f_{\theta}(x; p)$ denotes the model’s reasoning trajectory in answer space under parameters $\theta$. The outer loop optimizes over prompts, while the inner loop executes inference-time search within each induced reasoning policy. In practice, these two searches are not independent: better prompts shape smoother answer landscapes, reducing search depth and variance, while richer answer-space exploration yields more reliable reward signals for prompt optimization.

\begin{table}[!t]
\centering
\small
\caption{Comparison of representative methods for search in \textbf{prompt space}. Each approach optimizes prompts as discrete or continuous control variables, differing in search paradigm, reward design, and interpretability.}
\label{tab:prompt_search_methods}

% Adjust column types: l for short text, X for wrapped long text
\begin{tabularx}{\linewidth}{
>{\raggedright\arraybackslash}p{3cm}  % Method
>{\raggedright\arraybackslash}p{2.cm}  % Paradigm
>{\raggedright\arraybackslash}p{3cm}    % Search Type
>{\raggedright\arraybackslash}p{2.5cm}  % Reward Signal
X                                         % Key Characteristics
}
\toprule
\textbf{Method} & \textbf{Paradigm} & \textbf{Search Type} & \textbf{Reward Signal} & \textbf{Key Characteristics} \\ 
\midrule
\rowcolor{blockbg}
\textbf{GPS}~\cite{xu-etal-2022-gps} & Evolutionary & Discrete (Population-based) & Task accuracy & Genetic algorithm with mutation and crossover over natural-language prompts; gradient-free and data-efficient. \\
\textbf{EvoPrompt}~\cite{guo2025evopromptconnectingllmsevolutionary} & Evolutionary & Discrete (Hybrid) & Task accuracy & Integrates LLM-driven semantic mutation with classical evolutionary operators for robust prompt evolution. \\
\rowcolor{blockbg}
\textbf{PromptBreeder}~\cite{fernando2023promptbreederselfreferentialselfimprovementprompt} & Meta-Evolutionary & Discrete & Task accuracy & Co-evolves both task prompts and mutation prompts; self-referential optimization of mutation strategy. \\
\textbf{ProTeGi}~\cite{pryzant-etal-2023-automatic} & Gradient-inspired & Discrete (Textual Gradients) & Loss reduction & Constructs natural-language “gradients” to iteratively improve prompts via beam search and bandit selection. \\
\rowcolor{blockbg}
\textbf{GrIPS}~\cite{prasad2023gripsgradientfreeeditbasedinstruction} & Local Search & Discrete (Edit-based) & Task accuracy & Applies insertion/deletion/substitution edits to seed prompts; efficient for API-limited settings. \\
\textbf{APE}~\cite{zhou2023largelanguagemodelshumanlevel} & LLM-as-Search-Engine & Discrete & Task score (e.g., truthfulness) & Two-model setup where one LLM generates and another evaluates prompts; self-evaluative optimization loop. \\
\rowcolor{blockbg}
\textbf{PromptAgent}~\cite{wang2023promptagent} & Planning / Tree Search & Discrete (Sequential) & Composite reward & Models prompt editing as sequential decision-making with MCTS-based exploration and rollout evaluation. \\
\textbf{MPA}~\cite{wu-etal-2025-monte} & Adversarial MCTS & Discrete (Tree-based) & Attack success rate & Uses MCTS to discover adversarial suffixes for jailbreak attacks; highlights vulnerability exploration. \\
\rowcolor{blockbg}
\textbf{BPO}~\cite{cheng-etal-2024-black} & Preference-based & Discrete (Black-box) & Human preference & Aligns LLM outputs with human preferences via black-box search; alternative to PPO/DPO fine-tuning. \\
\textbf{HPME}~\cite{wen2023hardpromptseasygradientbased} & Gradient-based & Soft / Hybrid & Task accuracy & Learns discrete textual prompts via gradient relaxations; bridges hard and soft prompts. \\
\rowcolor{blockbg}
\textbf{InstructZero}~\cite{chen2023instructzeroefficientinstructionoptimization} & Latent / Bayesian & Soft (Latent) & Task performance & Optimizes latent soft prompts via Bayesian optimization for black-box models; transferable instruction generator. \\
\textbf{Adversarial Soft Search}~\cite{zou2023universaltransferableadversarialattacks} & Gradient-guided & Soft (Continuous) & Unsafe output likelihood & Learns continuous adversarial suffix embeddings transferable across aligned models. \\
\bottomrule
\end{tabularx}
\end{table}

\subsection{Discrete (Text-Based) Prompt Search}

Discrete prompt search methods aim to optimize natural language prompts directly in the token space without gradient updates or parameter tuning. Unlike soft prompt tuning, which learns continuous embeddings, these methods treat prompts as symbolic sequences and explore their combinatorial structure through heuristic or metaheuristic search. This makes them model-agnostic, accessible via APIs, and highly suitable for closed-source large language models (LLMs). The general framework can be summarized as: (i) generate new candidate prompts through mutation, rewriting, or planning; (ii) evaluate their performance using a task-specific metric or feedback signal; and (iii) retain or recombine the top-performing candidates. Below, we review representative methods and categorize them by their underlying search paradigm.

\vspace{-1.em}
\paragraph{Evolutionary and Population-Based Search.}
\textbf{GPS (Genetic Prompt Search)}~\cite{xu-etal-2022-gps} and \textbf{EvoPrompt}~\cite{guo2025evopromptconnectingllmsevolutionary} instantiate evolutionary algorithms for prompt optimization. GPS introduces a genetic algorithm that evolves prompts via mutation and crossover operators guided by task accuracy on a small validation set. EvoPrompt extends this idea by integrating an LLM into the evolutionary loop—LLMs are used to perform semantically coherent mutations and recombinations while classical evolutionary operators (selection, mutation, crossover) drive convergence. Both methods are gradient-free and exhibit strong data efficiency, outperforming manual and tuning-based baselines. Their \emph{reward function} is typically the task-specific performance score (e.g., accuracy or BLEU) evaluated on a held-out set. These approaches demonstrate that population diversity and linguistic mutation can jointly enhance prompt robustness and transferability.

\textbf{Promptbreeder}~\cite{fernando2023promptbreederselfreferentialselfimprovementprompt} extends evolutionary search into a self-referential paradigm, where the system co-evolves both task prompts and mutation prompts. The LLM learns to iteratively refine its own mutation strategy, forming a meta-evolutionary loop. This introduces a novel self-improvement dynamic where the mutation operator itself adapts to the optimization landscape, leading to improved generalization and domain-specific expertise.

\vspace{-1.em}
\paragraph{Gradient-Inspired and Beam Search Methods.}
\textbf{ProTeGi (Prompt Optimization with Textual Gradients)}~\cite{pryzant-etal-2023-automatic} proposes a hybrid search procedure inspired by numerical gradient descent. It constructs natural-language “gradients” that describe directions for prompt improvement, which are then propagated back into the text of the prompt. Beam search and multi-armed bandit selection are used to explore the most promising gradient edits efficiently. This approach combines human-readable textual updates with algorithmic rigor, bridging the gap between discrete prompt editing and continuous optimization. The reward signal corresponds to a task-specific loss reduction across mini-batches, aligning the search direction with performance improvement.

\vspace{-1.em}
\paragraph{Edit-Based Local Search.}
\textbf{GrIPS (Gradient-free, Edit-based Instruction Search)}~\cite{prasad2023gripsgradientfreeeditbasedinstruction} performs local edits (insertions, deletions, and substitutions) to improve task instructions. It searches within a constrained neighborhood around a seed instruction, using validation performance as the fitness measure. This method is efficient for API-constrained models, requiring minimal evaluations. GrIPS highlights that even small, syntactically simple edits can yield substantial semantic gains, demonstrating that local search in prompt space can approximate global optimization when guided by strong reward signals.

\vspace{-1.em}
\paragraph{LLM-as-Search-Engine Paradigm.}
\textbf{APE (Automatic Prompt Engineer)}~\cite{zhou2023largelanguagemodelshumanlevel} reframes prompt search as a two-model optimization loop. One LLM (the “engineer”) proposes candidate prompts, while another LLM (the “executor”) evaluates their zero-shot performance. The search proceeds by ranking and selecting candidates based on a scoring function that may incorporate truthfulness, informativeness, or accuracy. This black-box, self-evaluative process effectively treats the LLM as both generator and judge. The evaluation metric functions as a reward, and the system converges toward human-level prompt engineering performance without explicit supervision.

\vspace{-1.em}
\paragraph{Planning and Tree Search.}
\textbf{PromptAgent}~\cite{wang2023promptagent} and \textbf{MPA (Monte Carlo Tree Search-Based Prompt Autogeneration)}~\cite{wu-etal-2025-monte} apply strategic planning and tree-based exploration to the prompt optimization problem. PromptAgent models prompt design as a sequential decision process, where each editing or expansion step forms a node in a search tree. Monte Carlo Tree Search (MCTS) guides exploration using simulated rollouts and reward backpropagation from model feedback. The reward is defined as a composite of task success, coherence, and informativeness, enabling the discovery of expert-level prompts that rival human-crafted ones. MPA adapts this MCTS framework to adversarial prompt generation, where nodes represent attack suffixes and rewards correspond to successful jailbreak attempts. Both highlight the flexibility of search-based reasoning applied to prompt optimization.

\vspace{-1.em}
\paragraph{Preference-Guided Optimization.}
\textbf{Black-Box Prompt Optimization (BPO)}~\cite{cheng-etal-2024-black} optimizes prompts to align LLM outputs with human preferences, rather than fixed task metrics. The search process treats prompt editing as a preference-based black-box optimization problem, where feedback from human annotators or reward models serves as the fitness signal. This enables alignment without model retraining and achieves results comparable to reinforcement learning–based methods such as PPO and DPO. The approach thus reframes alignment as prompt-level optimization in the discrete space, demonstrating that search-based prompt tuning can substitute for costly fine-tuning.

\vspace{-1.em}
\paragraph{Comparative Analysis.}
Across these methods, the \emph{search complexity} depends on whether the algorithm explores global (population-based or tree search) or local (edit-based or gradient-inspired) neighborhoods. Evolutionary and MCTS-based methods are computationally intensive but robust to local minima, while edit-based and bandit-guided methods offer efficiency at the cost of search coverage. The \emph{reward design} also varies: task-level accuracy for supervised settings (GPS, GrIPS, ProTeGi), preference alignment for human-feedback–driven setups (BPO), and adversarial success for safety probing (MPA). Together, these discrete prompt search methods reveal a growing consensus that textual optimization—driven by structured search and LLM-in-the-loop evaluation—provides a scalable, interpretable, and model-agnostic pathway to enhance reasoning performance and alignment in large language models.

\subsection{Soft (Latent) Prompt Search}

While discrete prompt search operates directly in the textual space, \emph{soft prompt search} explores a continuous or latent representation of prompts. These methods treat prompts as trainable embeddings in the model’s input layer or in a lower-dimensional latent space, enabling differentiable optimization and gradient-based adaptation. The advantage lies in their efficiency and fine-grained controllability: small movements in the latent space can correspond to semantically rich changes in model behavior. However, the resulting prompts are often uninterpretable and may require additional mechanisms to map latent representations back to natural language.

\vspace{-1.em}
\paragraph{Gradient-Based Optimization and Discrete Projection.}
\textbf{Hard Prompts Made Easy (HPME)}~\cite{wen2023hardpromptseasygradientbased} introduces a hybrid optimization framework that bridges soft and hard prompts. The method applies gradient-based discrete optimization to directly learn textual prompts that are robust across modalities, such as text-to-image and text-to-text tasks. By estimating gradients through continuous relaxations of the discrete token space, the algorithm discovers effective hard prompts automatically, without requiring manual engineering. The optimized prompts outperform human-crafted ones in both generative and classification settings, showing that latent search with gradient feedback can serve as a bridge between interpretable text prompts and continuous embeddings. HPME also establishes that latent search can yield interpretable, transferable prompts through differentiable decoding mechanisms.

\vspace{-1.em}
\paragraph{Latent Instruction Optimization for Black-Box Models.}
\textbf{InstructZero}~\cite{chen2023instructzeroefficientinstructionoptimization} extends latent prompt optimization to black-box LLMs where backpropagation is unavailable. The approach trains a \emph{soft prompt} on an open-source model to generate explicit natural-language instructions for a target black-box model (e.g., GPT or ChatGPT). Each iteration converts the learned soft prompt into text using the open-source LLM, evaluates its zero-shot performance on the target model, and updates the latent representation using \emph{Bayesian optimization}. This process decouples gradient estimation from model access, effectively turning prompt optimization into a surrogate learning problem. The reward signal is the performance score (accuracy or preference alignment) on the target model’s outputs, allowing indirect gradient-free updates in latent space. This framework demonstrates that latent prompts can act as transferable “instruction generators” across models.

\vspace{-1.em}
\paragraph{Gradient-Guided Adversarial Prompt Search.}
\textbf{Universal and Transferable Adversarial Attacks on Aligned Language Models}~\cite{zou2023universaltransferableadversarialattacks} explores soft prompt search from a safety and robustness perspective. The authors use a combination of greedy and gradient-based search to optimize continuous adversarial suffix embeddings that induce objectionable behavior in aligned models. These suffixes, though discovered in latent embedding space, are decoded back into natural language to serve as universal adversarial prompts. The optimization objective maximizes the probability of eliciting affirmative (unsafe) responses, and the learned suffixes demonstrate strong transferability across model families, including both open- and closed-source LLMs (e.g., Vicuna, ChatGPT, Bard, and Claude). This highlights that continuous search techniques can capture global vulnerabilities shared across models, emphasizing the dual use of latent prompt search for alignment and red-teaming studies.

\section{Challenges and Future of Tree-Search Methods}
\label{sec:Challenges}

\noindent \textbf{Search Efficiency and Intelligence}. 
Tree search algorithms, despite their power, often require significantly greater computational resources than greedy decoding. As noted by \cite{wang2024litesearch}, resource demands can exceed 10 times that of greedy approaches, and other studies show that tree search can be 10--20 times slower than methods like iterative refinement, particularly if the evaluation model lacks high discrimination accuracy \citep{chen-etal-2024-tree}. This high computational and memory overhead presents a substantial barrier to practical deployment. Algorithms like MCTSr and LLaMA-Berry, which generate multiple solutions sequentially at each node, exacerbate these resource demands. To mitigate these limitations, future research could prioritize improving the efficiency of tree search algorithms. This includes investigating trade-offs between policy and reward models, incorporating dynamic control mechanisms, employing effective pruning techniques to optimize tree expansion, and developing more efficient memory solutions like tree-structured Key-Value (KV) caches to reduce I/O overhead in transformer models \citep{yao2025deft}.

\noindent \textbf{Overthinking Issues in Simple Queries.}
Task complexity is closely related to the length of reasoning chains, highlighting the need for extended cognitive processing in more difficult problems~\citep{qin2024o1,huang2025o1}. However, \cite{chen2024not} and \cite{zeng2024scaling} observe that tree-search models often overanalyze simple questions, dedicating excessive computational resources to tasks that have clear answers. For instance, a query like "3-2=?" does not require complex reasoning, yet these models may engage in unnecessary computations. This not only consumes valuable resources but can also degrade performance. Indeed, empirical studies show that in some scenarios, complex search frameworks like ToT and RAP perform even worse than simpler methods like Chain-of-Thought (CoT) or self-consistency \citep{parashar2025inference}. For tasks where the base LLM can already produce reasonable answers, simple sequential revision may be a more efficient and effective alternative to complex search \citep{snell2025scaling}. Future work should focus on methods for dynamically allocating computational resources, enabling models to quickly recognize and handle straightforward queries while reserving deliberate search for genuinely complex problems.

\noindent \textbf{Self-play Between Policy Models and Reward Models.} 
Certain tree-search algorithms encounter challenges due to limited parallelism, which constrains their search speed, especially in resource-intensive settings.
As detailed in Section~\ref{sec:MCTS}, various tree-search techniques can generate traces that are then employed to iteratively refine reward and policy models, such as ReST-MCTS and rStar-Math. This self-play paradigm is crucial for internalizing the reasoning system into the policy model, thereby endowing LLMs with sophisticated reasoning abilities~\citep{xiang2025towards}. By internalizing tree-search reasoning into LLMs, the search process can be structured within a CoT framework, facilitating sequential reasoning. This not only enhances reasoning efficiency but also mitigates parallelism limitations, thereby improving scalability. Future research should investigate strategies to optimize this self-play paradigm further, facilitating more efficient problem-solving.

\noindent \textbf{Reward Modeling and Reward Model Training.} 
Section~\ref{sec:MCTS} examines various MCTS-based evaluation strategies. A central element of the search strategies is the reward or evaluation model, which provides essential supervision to guide search processes effectively~\citep{lightman2023let,setlur2024rewarding,xiang2025towards}. 
Reward models are broadly categorized into two types: the Outcome Reward Model (ORM) and the Process Reward Model (PRM). Unlike outcome rewards, which deliver feedback only at the task's conclusion, process rewards provide signals at both intermediate steps and the final outcome, enabling finer-grained and more frequent supervision. Nevertheless, learning process rewards present significant challenges. For example, \cite{uesato2022solving,lightman2023let} relies on human annotators for process supervision, a costly and inherently unscalable method. While automated methods for constructing process rewards have been proposed~\citep{wang2024math,luo2024improve,wang2024multi}, they are predominantly designed for specialized areas such as mathematics and programming. These approaches struggle to generalize to broader domains, such as scientific reasoning and complex problem-solving, where human evaluation remains essential. Overcoming these limitations necessitates the development of more efficient methods to generate high-quality fine-grained rewards and scalable techniques to advance reward model capabilities, which remain open and pressing research challenges.

\noindent \textbf{Reward Model Quality and Its Effect on Search.}
The performance and efficiency of test-time search depend critically on the quality of the reward model \citep{setlur2024rewarding,xiang2025towards}. An imperfect reward model can give rise to inverse inference scaling, where expanding the search space negatively impacts performance due to a distribution shift between the reward and policy models \citep{gao2023scaling,zeng2024scalingsearchlearningroadmap}. Furthermore, the value functions produced by LLMs often act as heuristics based on world knowledge rather than being rigorously defined by the Bellman equation, which is the foundation of traditional reinforcement learning \citep{sutton2018reinforcement,yao2023tree}. This deviation means that fundamental properties, such as the optimality guarantees in A* search, can be compromised if the LLM-based heuristic is inadmissible \citep{bellman1966dynamic}. These findings underscore the critical need to bridge the performance gap between oracle and learned reward models. Understanding how scaling laws for process supervision models influence their effectiveness in large-scale search tasks remains a pivotal challenge \citep{xiang2025towards}.

\noindent \textbf{Applicability to Irreversible Environments}.
A significant limitation of many current search frameworks is their reliance on the ability to undo actions. Processes like backtracking in DFS \citep{yao2023tree}, selecting an alternative node in Best-First Search \citep{koh2024tree}, or simulating rollouts in MCTS \citep{zhou2024language} all presuppose that the agent can return to a previous state to explore a different path. This restricts their applicability primarily to virtual or simulated environments (e.g., solving puzzles, web navigation) where state transitions are fully reversible and carry no real-world consequences. However, in dynamic, interactive environments, many actions are irreversible—such as sending an email, executing a financial trade, or administering a medication. A crucial direction for future research is to devise new frameworks capable of planning and acting effectively in environments with irreversible actions, which will be essential for deploying these agents in real-world scenarios.

\noindent \textbf{Balancing Test-Time Search with Foundational Model Capabilities}.
Finally, the field must consider the broader trade-off between enhancing LLMs through intensive test-time computation versus improving their innate, built-in reasoning capabilities. While complex search algorithms can significantly boost performance on difficult tasks, an alternative and complementary research direction focuses on training LLMs to internalize tree-like reasoning directly. Optimization frameworks such as Chain of Preference Optimization \citep{zhang2024chain} aim to distill the deliberative process of search into the model's parameters, reducing the need for expensive real-time computation. Future work should continue this parallel investigation: enhancing the power and efficiency of test-time search while simultaneously advancing training methods that endow LLMs with more autonomous, multi-path thinking capabilities from the outset.

\section{Conclusion}
\label{sec:Conclusion}

This survey has charted the evolution of tree search-based algorithms as a pivotal strategy for scaling the reasoning capabilities of Large Language Models at inference time.  By establishing a unified framework, we have systematically compared a diverse landscape of methods-from foundational search techniques to the more sophisticated Monte Carlo Tree Search-highlighting their distinct approaches to node representation, reward formulation, and algorithmic adaptation.  Our analysis underscores that while these methods successfully overcome the single-path limitations of approaches like Chain-of-Thought, their practical application is constrained by two primary challenges: substantial computational overhead and the critical bottleneck of designing high-quality reward models.  As imperfect reward signals can lead to "inverse inference scaling", where more search degrades performance, the future of the field hinges on developing more intelligent, resource-efficient search dynamics and, most importantly, creating scalable techniques for generating high-fidelity process rewards. Addressing these open problems is essential to unlocking the full potential of tree search for advancing general-purpose AI reasoning.

\clearpage
\bibliographystyle{refstyle}
\bibliography{ref}

\end{document}